\pdfoutput=1
\documentclass[11pt]{article}

\usepackage[]{ACL2023}

\usepackage{times}
\usepackage{amssymb}
\usepackage{amsmath}
\usepackage{latexsym}
\usepackage{graphicx,xcolor}  % グラフィックス関連
\usepackage{multirow}
\usepackage{booktabs}
\usepackage{color, colortbl}

\definecolor{Gray}{gray}{0.92}

\usepackage[T1]{fontenc}
\usepackage[utf8]{inputenc}

\usepackage{microtype}

\usepackage{inconsolata}

\newcommand{\M}{\textsc{NoisyICL}}

\title{\M: A Little Noise in Model Parameters Calibrates\\ In-context Learning}

\author{Yufeng Zhao${}^{1}$, Yoshihiro Sakai${}^{1}$, Naoya Inoue${}^{1,2}$\\
${}^{1}$Japan Advanced Institute of Science and Technology, ${}^{2}$RIKEN\\
\texttt{\{yfzhao, y.sakai, naoya-i\}@jaist.ac.jp}}

\begin{document}
\maketitle
\begin{abstract}
\textbf{I}n-\textbf{C}ontext \textbf{L}earning (ICL) is suffering from unsatisfactory performance and under-calibration due to high prior bias and unfaithful confidence. Some previous works fine-tuned language models for better ICL performance with enormous datasets and computing costs. In this paper, we propose \M, simply perturbing the model parameters by random noises to strive for better performance and calibration. Our experiments on two models and 12 downstream datasets show that \M{} can help ICL produce more accurate predictions. Our further analysis indicates that \M{} enables the model to provide more fair predictions, and also with more faithful confidence. Therefore, we believe that \M\ is an effective calibration of ICL. 
Our experimental code is uploaded to Github\footnote{\url{github.com/HHkz/NoisyICL}}.
%\footnote{Not available during anonymous review.}.

\end{abstract}

\section{Introduction}
\label{1}

\textbf{L}anguage \textbf{M}odels (LMs) have demonstrated the ability of \textbf{I}n-\textbf{C}ontext \textbf{L}earning (ICL), where LMs learn tasks from few-shot input-label demonstrations in the form of natural language, without explicit parameter updates~\cite{dong2022survey}.

Nevertheless, ICL still underperforms the end-to-end models \cite{mosbach2023few}. A recent study shows that vanilla LMs are biased towards the knowledge acquired during pre-training, which is deemed harmful to ICL \cite{fei-etal-2023-mitigating}. There has been some effort in fine-tuning or calibrating LMs towards ICL tasks \cite{min2022metaicl, zhao2021calibrate, wei2021finetuned, fei-etal-2023-mitigating, lu2022fantastically}, which focus on reducing the gap between the knowledge gained from pre-training and ICL tasks, producing significant improvements in the ICL performance. However, fine-tuning these enormous LMs on additional data incurs a considerably high computational cost.

\begin{figure}[t]
    \centering
    \centerline{\includegraphics[width=\linewidth]{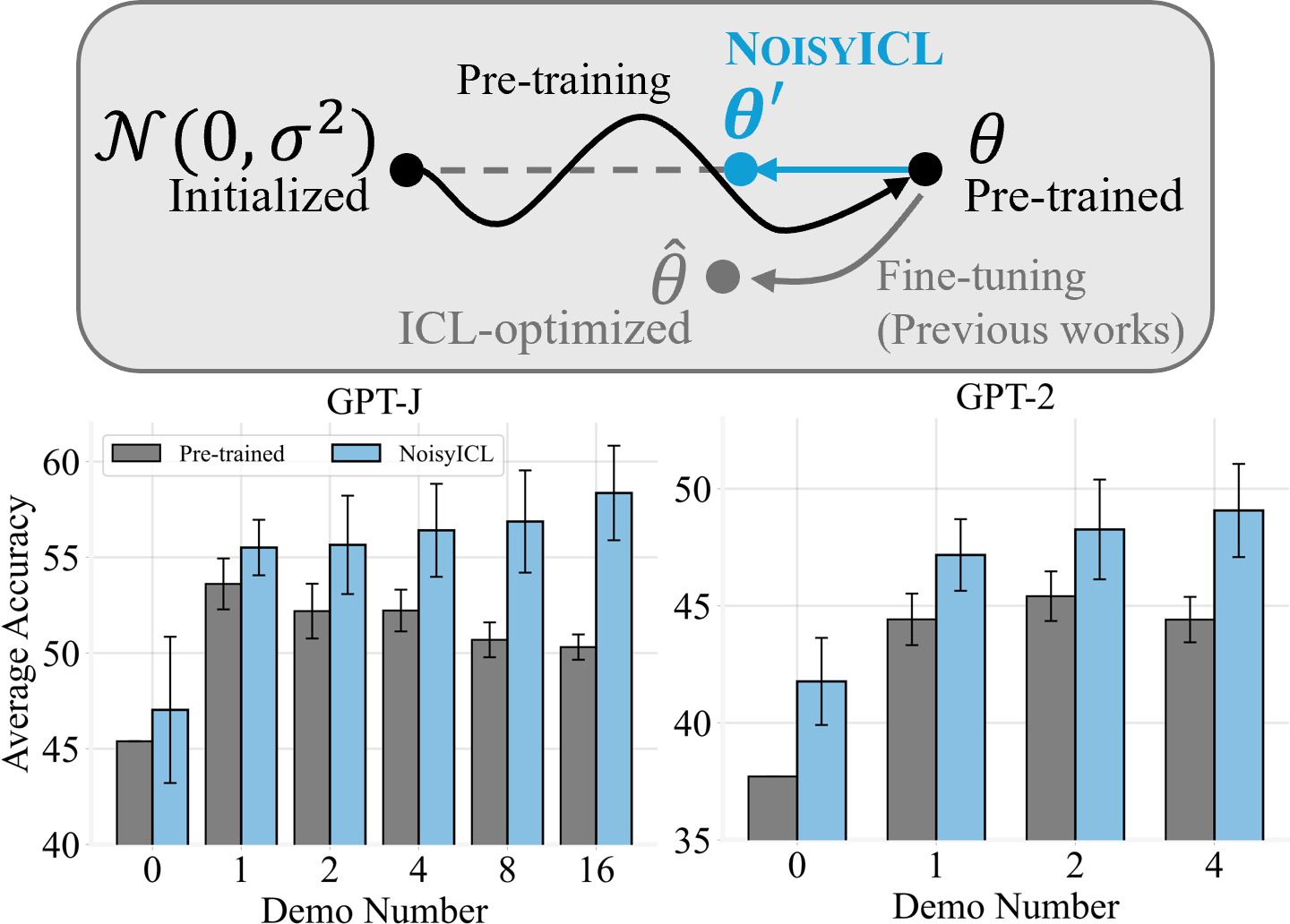}}
    \caption{\textbf{Upper}: A sketch of \M{}: Unlike previous works which fine-tuned LMs towards ICL tasks, we perturb LMs by random noise sampled from the normal distribution $\mathcal{N}(0,\sigma^2)$ with intensity $\lambda$, then perform ICL. \textbf{Lower}: The average accuracy of ICL with and without \M{} w.r.t. the number of demos.}
    \label{fig:0}
\end{figure}

% \begin{figure}[t]
%     \centering
%     \centerline{\includegraphics[width=\linewidth]{Fig/f_1.jpg}}
%     \caption{A sketch of \textbf{A). In-context Learning}: Language models solve downstream tasks by predicting the next token of a given prompt with few-shot demos. \textbf{B). \M{}}: Language models are first perturbated by random noise sampled from the normal distribution $N(0,\sigma^2)$ with intensity $\lambda$, then perform ICL.}
%     \label{fig:0}
% \end{figure}

\begin{table*}[t]
\caption{Accuracy and Macro-F1 results ($mean_{std}$, $k=4$). A better result is in {\color[HTML]{00ba55}green}. $\mathbf{\lambda}$: The optimal intensity of noise searched on the validation set, \textbf{Acc.}: Accuracy, \textbf{MF1}: Macro-F1, $\mathbf{ECE_1}$: the Expected Calibrated Error described in \S\ref{3.3}; \textbf{w/o}: Not using \M , \textbf{w/}: Using \M ; Datasets abbreviation described in Appendix. \ref{Appendix:Datasets}; \textbf{Val.}: the results on the validation sets, \textbf{Test}: the results on the testing sets.} 
\label{tab:1}
\centering
\resizebox{2.05\columnwidth}{!}{

\begin{tabular}{@{}cccc|cccccccccccc|c@{}}
\toprule
\multicolumn{4}{c|}{Dataset} & PS & HS & SE'14R & SE'14L & RTE & MRPC & Ethos & FP & SST2 & TEE & TES & TEH & Mean \\ \midrule
 & \multicolumn{3}{c|}{$\lambda$} & 0.2 & 0.2 & 0.1 & 0.1 & 0.012 & 0.2 & 0.004 & 0.08 & 0.004 & 0.01 & 0.08 & 0.002 & — \\ \cmidrule(l){2-17} 
 &  &  & w/o & ${37.46}_{1.36}$ & ${71.97}_{1.19}$ & ${36.48}_{0.84}$ & ${34.82}_{0.21}$ & ${49.88}_{1.03}$ & ${43.18}_{1.48}$ & ${53.44}_{0.56}$ & ${62.36}_{0.17}$ & ${81.48}_{0.97}$ & ${45.68}_{0.48}$ & ${49.32}_{1.07}$ & ${50.04}_{1.68}$ & ${51.34}$ \\
 &  & \multirow{-2}{*}{Val.} & w/ & {\color[HTML]{00ba55} ${59.82}_{5.82}$} & {\color[HTML]{00ba55} ${82.95}_{1.36}$} & {\color[HTML]{00ba55} ${51.54}_{7.44}$} & {\color[HTML]{00ba55} ${45.14}_{4.06}$} & {\color[HTML]{00ba55} ${51.00}_{1.35}$} & {\color[HTML]{00ba55} ${59.82}_{6.15}$} & {\color[HTML]{00ba55} ${56.00}_{0.89}$} & {\color[HTML]{00ba55} ${62.42}_{0.43}$} & {\color[HTML]{00ba55} ${82.52}_{1.03}$} & {\color[HTML]{00ba55} ${46.35}_{0.53}$} & {\color[HTML]{00ba55} ${50.90}_{1.32}$} & {\color[HTML]{00ba55} ${50.35}_{1.06}$} & {\color[HTML]{00ba55} ${61.23}$} \\ \cmidrule(l){3-17} 
 &  &  & w/o & ${33.09}_{1.76}$ & ${81.30}_{0.79}$ & ${33.67}_{1.31}$ & ${34.26}_{1.33}$ & ${46.52}_{1.11}$ & ${50.97}_{1.43}$ & {\color[HTML]{00ba55} ${57.74}_{1.38}$} & ${61.87}_{0.16}$ & {\color[HTML]{00ba55} ${79.15}_{1.17}$} & ${44.72}_{0.62}$ & {\color[HTML]{00ba55} ${50.20}_{0.85}$} & {\color[HTML]{00ba55} ${53.16}_{1.07}$} & ${52.22}$ \\
 & \multirow{-4}{*}{\begin{tabular}[c]{@{}c@{}}Acc.\\ (\%)\end{tabular}} & \multirow{-2}{*}{Test} & w/ & \cellcolor[HTML]{EFEFEF}{\color[HTML]{00ba55} ${55.96}_{3.43}$} & \cellcolor[HTML]{EFEFEF}{\color[HTML]{00ba55} ${87.16}_{5.29}$} & \cellcolor[HTML]{EFEFEF}{\color[HTML]{00ba55} ${46.79}_{4.47}$} & \cellcolor[HTML]{EFEFEF}{\color[HTML]{00ba55} ${41.77}_{3.80}$} & \cellcolor[HTML]{EFEFEF}{\color[HTML]{00ba55} ${47.64}_{1.27}$} & \cellcolor[HTML]{EFEFEF}{\color[HTML]{00ba55} ${53.43}_{5.23}$} & \cellcolor[HTML]{EFEFEF}${57.07}_{1.53}$ & \cellcolor[HTML]{EFEFEF}{\color[HTML]{00ba55} ${61.90}_{0.30}$} & \cellcolor[HTML]{EFEFEF}${78.82}_{0.75}$ & \cellcolor[HTML]{EFEFEF}{\color[HTML]{00ba55} ${45.17}_{0.57}$} & \cellcolor[HTML]{EFEFEF}${48.11}_{1.38}$ & \cellcolor[HTML]{EFEFEF}${53.12}_{1.10}$ & \cellcolor[HTML]{EFEFEF}{\color[HTML]{00ba55} ${56.41}$} \\ \cmidrule(l){2-17} 
 &  &  & w/o & {\color[HTML]{00ba55} ${24.11}_{0.87}$} & {\color[HTML]{00ba55} ${26.75}_{1.38}$} & {\color[HTML]{000000} ${32.53}_{1.46}$} & ${28.62}_{0.68}$ & ${47.37}_{1.14}$ & ${43.18}_{1.48}$ & ${53.24}_{0.69}$ & ${25.91}_{0.26}$ & ${81.24}_{1.04}$ & ${25.65}_{0.54}$ & ${32.87}_{1.81}$ & ${49.01}_{1.81}$ & ${39.20}$ \\
 &  & \multirow{-2}{*}{Val.} & w/ & ${20.80}_{1.13}$ & ${24.22}_{0.63}$ & {\color[HTML]{00ba55} ${46.88}_{6.70}$} & {\color[HTML]{00ba55} ${43.25}_{4.12}$} & {\color[HTML]{00ba55} ${48.61}_{1.38}$} & {\color[HTML]{00ba55} ${48.38}_{2.69}$} & {\color[HTML]{00ba55} ${55.89}_{0.86}$} & {\color[HTML]{00ba55} ${27.81}_{0.74}$} & {\color[HTML]{00ba55} ${82.33}_{1.06}$} & {\color[HTML]{00ba55} ${26.31}_{0.80}$} & {\color[HTML]{00ba55} ${38.19}_{5.33}$} & {\color[HTML]{00ba55} ${49.37}_{0.92}$} & {\color[HTML]{00ba55} ${42.67}$} \\ \cmidrule(l){3-17} 
 &  &  & w/o & ${21.86}_{0.86}$ & {\color[HTML]{00ba55} ${26.61}_{1.61}$} & ${30.71}_{1.47}$ & ${33.17}_{1.52}$ & ${44.94}_{1.30}$ & {\color[HTML]{00ba55} ${49.56}_{1.48}$} & {\color[HTML]{00ba55} ${57.24}_{1.45}$} & ${26.59}_{0.52}$ & {\color[HTML]{00ba55} ${77.44}_{1.36}$} & ${23.57}_{0.86}$ & {\color[HTML]{00ba55} ${34.23}_{1.25}$} & {\color[HTML]{00ba55} ${53.16}_{1.07}$} & ${39.92}$ \\
 & \multirow{-4}{*}{\begin{tabular}[c]{@{}c@{}}MF1\\ (\%)\end{tabular}} & \multirow{-2}{*}{Test} & w/ & \cellcolor[HTML]{EFEFEF}{\color[HTML]{00ba55} ${22.87}_{1.57}$} & \cellcolor[HTML]{EFEFEF}${24.22}_{0.89}$ & \cellcolor[HTML]{EFEFEF}{\color[HTML]{00ba55} ${43.88}_{3.90}$} & \cellcolor[HTML]{EFEFEF}{\color[HTML]{00ba55} ${42.17}_{4.22}$} & \cellcolor[HTML]{EFEFEF}{\color[HTML]{00ba55} ${45.93}_{1.31}$} & \cellcolor[HTML]{EFEFEF}${46.51}_{1.56}$ & \cellcolor[HTML]{EFEFEF}${56.46}_{1.54}$ & \cellcolor[HTML]{EFEFEF}{\color[HTML]{00ba55} ${27.24}_{1.22}$} & \cellcolor[HTML]{EFEFEF}${77.11}_{0.84}$ & \cellcolor[HTML]{EFEFEF}{\color[HTML]{00ba55} ${23.78}_{0.84}$} & \cellcolor[HTML]{EFEFEF}${32.68}_{2.17}$ & \cellcolor[HTML]{EFEFEF}${53.11}_{1.10}$ & \cellcolor[HTML]{EFEFEF}{\color[HTML]{00ba55} ${41.33}$} \\ \cmidrule(l){2-17} 
 &  &  & w/o & ${28.70}_{1.78}$ & ${14.39}_{1.13}$ & ${31.20}_{0.69}$ & ${38.56}_{0.36}$ & ${29.35}_{0.95}$ & ${27.79}_{1.50}$ & ${23.20}_{0.62}$ & ${23.90}_{0.17}$ & {\color[HTML]{00ba55} ${11.45}_{0.93}$} & ${42.79}_{0.36}$ & ${17.57}_{0.81}$ & ${28.45}_{1.23}$ & ${26.44}$ \\
 &  & \multirow{-2}{*}{Val.} & w/ & {\color[HTML]{00ba55} ${12.77}_{2.94}$} & {\color[HTML]{00ba55} ${9.87}_{2.56}$} & {\color[HTML]{00ba55} ${13.96}_{7.26}$} & {\color[HTML]{00ba55} ${21.34}_{6.75}$} & {\color[HTML]{00ba55} ${28.11}_{0.95}$} & {\color[HTML]{00ba55} ${8.79}_{3.41}$} & {\color[HTML]{00ba55} ${20.33}_{0.99}$} & {\color[HTML]{00ba55} ${21.43}_{3.83}$} & ${12.86}_{1.06}$ & {\color[HTML]{00ba55} ${42.03}_{0.88}$} & {\color[HTML]{00ba55} ${10.28}_{5.37}$} & {\color[HTML]{00ba55} ${28.18}_{1.37}$} & {\color[HTML]{00ba55} ${19.16}$} \\ \cmidrule(l){3-17} 
 &  &  & w/o & ${30.33}_{1.42}$ & ${9.41}_{0.78}$ & ${34.45}_{1.39}$ & ${36.78}_{1.56}$ & ${32.42}_{1.10}$ & ${28.29}_{1.47}$ & {\color[HTML]{00ba55} ${18.97}_{1.45}$} & ${23.70}_{0.49}$ & ${10.75}_{1.02}$ & ${44.09}_{0.79}$ & ${16.56}_{0.56}$ & {\color[HTML]{00ba55} ${24.50}_{0.99}$} & ${25.85}$ \\
\multirow{-13}{*}{GPT-J} & \multirow{-4}{*}{\begin{tabular}[c]{@{}c@{}}$ECE_1$\\ (\%, $\downarrow$)\end{tabular}} & \multirow{-2}{*}{Test} & w/ & \cellcolor[HTML]{EFEFEF}{\color[HTML]{00ba55} ${9.65}_{2.83}$} & \cellcolor[HTML]{EFEFEF}{\color[HTML]{00ba55} ${5.97}_{3.45}$} & \cellcolor[HTML]{EFEFEF}{\color[HTML]{00ba55} ${17.53}_{5.87}$} & \cellcolor[HTML]{EFEFEF}{\color[HTML]{00ba55} ${23.85}_{5.43}$} & \cellcolor[HTML]{EFEFEF}{\color[HTML]{00ba55} ${31.49}_{1.45}$} & \cellcolor[HTML]{EFEFEF}{\color[HTML]{00ba55} ${20.24}_{6.56}$} & \cellcolor[HTML]{EFEFEF}${19.28}_{1.91}$ & \cellcolor[HTML]{EFEFEF}{\color[HTML]{00ba55} ${22.77}_{3.17}$} & \cellcolor[HTML]{EFEFEF}{\color[HTML]{00ba55} ${10.62}_{0.84}$} & \cellcolor[HTML]{EFEFEF}{\color[HTML]{00ba55} ${43.81}_{0.76}$} & \cellcolor[HTML]{EFEFEF}{\color[HTML]{00ba55} ${14.43}_{2.40}$} & \cellcolor[HTML]{EFEFEF}${24.67}_{0.92}$ & \cellcolor[HTML]{EFEFEF}{\color[HTML]{00ba55} ${20.36}$} \\ \midrule
 & \multicolumn{3}{c|}{$\lambda$} & 0.014 & 0.1 & 0.08 & 0.06 & 0.002 & 0.08 & 0.2 & 0.04 & 0.014 & 0.04 & 0.04 & 0.2 & --- \\ \cmidrule(l){2-17} 
 &  &  & w/o & ${45.44}_{1.50}$ & ${39.26}_{0.95}$ & ${44.65}_{0.51}$ & ${42.83}_{0.93}$ & ${51.41}_{0.76}$ & {\color[HTML]{00ba55} ${69.06}_{0.19}$} & ${43.20}_{0.41}$ & ${53.03}_{1.35}$ & ${59.63}_{0.28}$ & ${29.67}_{0.77}$ & ${38.55}_{1.43}$ & ${41.82}_{0.51}$ & ${46.54}$ \\
 &  & \multirow{-2}{*}{Val.} & w/ & {\color[HTML]{00ba55} ${49.17}_{1.43}$} & {\color[HTML]{00ba55} ${64.06}_{3.41}$} & {\color[HTML]{00ba55} ${46.05}_{0.66}$} & {\color[HTML]{00ba55} ${46.99}_{1.77}$} & {\color[HTML]{00ba55} ${52.11}_{1.50}$} & ${59.65}_{3.26}$ & {\color[HTML]{00ba55} ${51.64}_{1.67}$} & {\color[HTML]{00ba55} ${58.26}_{2.02}$} & {\color[HTML]{00ba55} ${61.50}_{1.23}$} & {\color[HTML]{00ba55} ${31.17}_{0.94}$} & {\color[HTML]{00ba55} ${42.44}_{1.86}$} & {\color[HTML]{00ba55} ${51.48}_{4.38}$} & {\color[HTML]{00ba55} ${51.21}$} \\ \cmidrule(l){3-17} 
 &  &  & w/o & ${39.62}_{1.33}$ & ${43.62}_{1.17}$ & ${40.03}_{1.26}$ & {\color[HTML]{00ba55} ${39.41}_{1.29}$} & ${50.89}_{1.04}$ & {\color[HTML]{00ba55} ${63.12}_{0.38}$} & ${45.93}_{0.51}$ & ${52.27}_{0.81}$ & ${59.47}_{1.41}$ & ${27.43}_{0.59}$ & ${29.48}_{1.29}$ & ${41.66}_{0.57}$ & ${44.41}$ \\
 & \multirow{-4}{*}{\begin{tabular}[c]{@{}c@{}}Acc.\\ (\%)\end{tabular}} & \multirow{-2}{*}{Test} & w/ & \cellcolor[HTML]{EFEFEF}{\color[HTML]{00ba55} ${40.95}_{1.42}$} & \cellcolor[HTML]{EFEFEF}{\color[HTML]{00ba55} ${67.25}_{4.00}$} & \cellcolor[HTML]{EFEFEF}{\color[HTML]{00ba55} ${45.28}_{3.50}$} & \cellcolor[HTML]{EFEFEF}${36.90}_{0.18}$ & \cellcolor[HTML]{EFEFEF}{\color[HTML]{00ba55} ${51.01}_{0.91}$} & \cellcolor[HTML]{EFEFEF}${61.20}_{0.68}$ & \cellcolor[HTML]{EFEFEF}{\color[HTML]{00ba55} ${50.90}_{2.63}$} & \cellcolor[HTML]{EFEFEF}{\color[HTML]{00ba55} ${56.19}_{1.18}$} & \cellcolor[HTML]{EFEFEF}{\color[HTML]{00ba55} ${61.54}_{1.51}$} & \cellcolor[HTML]{EFEFEF}{\color[HTML]{00ba55} ${33.38}_{1.16}$} & \cellcolor[HTML]{EFEFEF}{\color[HTML]{00ba55} ${34.01}_{4.15}$} & \cellcolor[HTML]{EFEFEF}{\color[HTML]{00ba55} ${50.28}_{2.44}$} & \cellcolor[HTML]{EFEFEF}{\color[HTML]{00ba55} ${49.07}$} \\ \cmidrule(l){2-17} 
 &  &  & w/o & ${25.62}_{1.81}$ & ${17.88}_{0.37}$ & {\color[HTML]{00ba55} ${37.69}_{0.08}$} & {\color[HTML]{00ba55} ${37.71}_{0.88}$} & ${50.82}_{0.67}$ & ${41.45}_{0.49}$ & ${34.36}_{0.41}$ & ${35.02}_{2.00}$ & ${57.74}_{0.42}$ & ${20.33}_{0.49}$ & ${33.74}_{1.86}$ & ${30.98}_{0.50}$ & ${35.28}$ \\
 &  & \multirow{-2}{*}{Val.} & w & {\color[HTML]{00ba55} ${26.05}_{1.59}$} & {\color[HTML]{00ba55} ${24.20}_{0.85}$} & ${32.40}_{0.25}$ & ${30.78}_{1.69}$ & {\color[HTML]{00ba55} ${51.57}_{1.59}$} & {\color[HTML]{00ba55} ${48.90}_{1.11}$} & {\color[HTML]{00ba55} ${49.38}_{0.17}$} & {\color[HTML]{00ba55} ${35.35}_{1.00}$} & {\color[HTML]{00ba55} ${61.28}_{1.18}$} & {\color[HTML]{00ba55} ${22.14}_{0.91}$} & {\color[HTML]{00ba55} ${34.46}_{1.13}$} & {\color[HTML]{00ba55} ${46.20}_{3.91}$} & {\color[HTML]{00ba55} ${38.56}$} \\ \cmidrule(l){3-17} 
 &  &  & w/o & {\color[HTML]{00ba55} ${24.70}_{1.17}$} & ${18.24}_{0.52}$ & {\color[HTML]{00ba55} ${35.78}_{1.49}$} & {\color[HTML]{00ba55} ${38.79}_{1.25}$} & ${49.37}_{1.31}$ & ${39.90}_{0.94}$ & ${35.56}_{0.75}$ & ${35.78}_{1.11}$ & ${59.46}_{1.42}$ & ${18.89}_{0.62}$ & ${29.36}_{1.29}$ & ${34.43}_{0.79}$ & ${35.02}$ \\
 & \multirow{-4}{*}{\begin{tabular}[c]{@{}c@{}}MF1\\ (\%)\end{tabular}} & \multirow{-2}{*}{Test} & w/ & \cellcolor[HTML]{EFEFEF}${24.49}_{1.20}$ & \cellcolor[HTML]{EFEFEF}{\color[HTML]{00ba55} ${23.74}_{1.10}$} & \cellcolor[HTML]{EFEFEF}${34.03}_{1.20}$ & \cellcolor[HTML]{EFEFEF}${27.41}_{1.97}$ & \cellcolor[HTML]{EFEFEF}{\color[HTML]{00ba55} ${49.56}_{0.99}$} & \cellcolor[HTML]{EFEFEF}{\color[HTML]{00ba55} ${41.46}_{1.26}$} & \cellcolor[HTML]{EFEFEF}{\color[HTML]{00ba55} ${47.25}_{3.97}$} & \cellcolor[HTML]{EFEFEF}{\color[HTML]{00ba55} ${36.67}_{2.14}$} & \cellcolor[HTML]{EFEFEF}{\color[HTML]{00ba55} ${61.15}_{1.50}$} & \cellcolor[HTML]{EFEFEF}{\color[HTML]{00ba55} ${24.06}_{1.39}$} & \cellcolor[HTML]{EFEFEF}{\color[HTML]{00ba55} ${31.48}_{3.32}$} & \cellcolor[HTML]{EFEFEF}{\color[HTML]{00ba55} ${49.05}_{1.35}$} & \cellcolor[HTML]{EFEFEF}{\color[HTML]{00ba55} ${37.52}$} \\ \cmidrule(l){2-17} 
 &  &  & w/o & ${5.94}_{0.81}$ & ${36.18}_{0.96}$ & {\color[HTML]{00ba55} ${14.79}_{0.92}$} & {\color[HTML]{00ba55} ${15.51}_{1.25}$} & ${30.35}_{1.12}$ & {\color[HTML]{00ba55} ${19.39}_{0.40}$} & ${47.50}_{0.37}$ & ${8.33}_{1.12}$ & {\color[HTML]{00ba55} ${2.10}_{0.88}$} & {\color[HTML]{00ba55} ${30.85}_{1.06}$} & ${14.17}_{1.04}$ & ${51.18}_{0.62}$ & ${23.02}$ \\
 &  & \multirow{-2}{*}{Val.} & w/ & {\color[HTML]{00ba55} ${5.74}_{1.40}$} & {\color[HTML]{00ba55} ${11.64}_{2.21}$} & ${16.36}_{4.41}$ & ${17.55}_{1.50}$ & {\color[HTML]{00ba55} ${29.51}_{1.54}$} & ${21.23}_{1.51}$ & {\color[HTML]{00ba55} ${24.01}_{0.97}$} & {\color[HTML]{00ba55} ${8.20}_{1.78}$} & ${2.99}_{1.10}$ & ${30.92}_{1.99}$ & {\color[HTML]{00ba55} ${13.52}_{1.18}$} & {\color[HTML]{00ba55} ${23.93}_{5.48}$} & {\color[HTML]{00ba55} ${17.13}$} \\ \cmidrule(l){3-17} 
 &  &  & w/o & ${9.29}_{1.37}$ & ${28.14}_{1.21}$ & {\color[HTML]{00ba55} ${17.90}_{1.04}$} & {\color[HTML]{00ba55} ${15.04}_{1.29}$} & ${31.42}_{1.13}$ & {\color[HTML]{00ba55} ${23.02}_{0.54}$} & ${45.21}_{0.78}$ & ${8.00}_{0.92}$ & {\color[HTML]{00ba55} ${1.89}_{0.55}$} & ${32.76}_{0.80}$ & ${22.30}_{1.33}$ & ${48.69}_{0.51}$ & ${23.64}$ \\
\multirow{-13}{*}{GPT-2} & \multirow{-4}{*}{\begin{tabular}[c]{@{}c@{}}$ECE_1$\\ (\%, $\downarrow$)\end{tabular}} & \multirow{-2}{*}{Test} & w/ & \cellcolor[HTML]{EFEFEF}{\color[HTML]{00ba55} ${8.16}_{1.12}$} & \cellcolor[HTML]{EFEFEF}{\color[HTML]{00ba55} ${9.01}_{2.57}$} & \cellcolor[HTML]{EFEFEF}${18.69}_{1.81}$ & \cellcolor[HTML]{EFEFEF}${25.35}_{2.98}$ & \cellcolor[HTML]{EFEFEF}{\color[HTML]{00ba55} ${31.07}_{1.22}$} & \cellcolor[HTML]{EFEFEF}${26.06}_{1.64}$ & \cellcolor[HTML]{EFEFEF}{\color[HTML]{00ba55} ${25.52}_{4.12}$} & \cellcolor[HTML]{EFEFEF}{\color[HTML]{00ba55} ${6.88}_{1.71}$} & \cellcolor[HTML]{EFEFEF}${3.12}_{0.80}$ & \cellcolor[HTML]{EFEFEF}{\color[HTML]{00ba55} ${29.73}_{2.28}$} & \cellcolor[HTML]{EFEFEF}{\color[HTML]{00ba55} ${19.15}_{4.69}$} & \cellcolor[HTML]{EFEFEF}{\color[HTML]{00ba55} ${21.63}_{2.11}$} & \cellcolor[HTML]{EFEFEF}{\color[HTML]{00ba55} ${18.70}$} \\ \bottomrule
\end{tabular}}
\end{table*}

% We believe that adding noise to model parameters, which is effective in reducing the pre-trained overfitting of the model in the pre-training and fine-tuning paradigm \cite{wu2022noisytune, zhang2020revisiting}, fits the model from the pre-training to ICL.   

Inspired by \newcite{wu2022noisytune}, which shows the benefits of noise for fine-tuning end-to-end LMs, we hypothesize that introducing noise to the model parameters of pre-trained LMs can fit LMs towards ICL with a lower computational cost. In this paper, we propose \M{}, which simply adds noise to model parameters and then performs ICL on the noised models, as shown in Fig.~\ref{fig:0}, to investigate the benefit of noise for ICL. 

As shown in Fig.~\ref{fig:0} and Table~\ref{tab:1}, our experiments on two models and 12 classification datasets show that adding appropriate noise into model parameters improves the performance of ICL by around $10\%$ with obviously lower computational cost. 

To investigate the reason for such performance improvement, our further experiments hypothesize that \M{} acts as a model calibration. In detail, we find that: \textbf{1.} \M\ neutralizes prediction bias among general tokens and labels, and \textbf{2.} \M\ leads the model to produce more faithful confidence. 
%Firstly, we introduce the calibration investigated in this paper following these two aspects:
%Following such concepts of calibration, we conduct further analysis and point out that

\paragraph{Our contribution can be summarized as:}

\begin{itemize}
    \item We propose \M, which simply adds noise into LMs and then executes ICL (\S\ref{sec:2}). Our experiment shows that \M\ obtains a better ICL performance (\S\ref{3.2}).
    \item We show that adding noise effectively calibrates LMs to reduce prediction bias and unfaithful confidence in ICL (\S\ref{3.3}).
\end{itemize}

\section{\M}
\label{sec:2}

Here we introduce the basic form of ICL and our perturbation method named \M.

\paragraph{In-context Learning.} Suppose a supervised classification dataset $\mathcal{D} = \{(x_i, y_i)\}_{i=1}^n$, where $x_i$ is an input, and $y_i\in \mathbb{U}$ is the corresponding label in a label space $\mathbb{U}$. For each query $x_q$ to be predicted, we sample a demo sequence $G = \{(x_{a_j}, y_{a_j})\}_{j=1}^k$ from $\mathcal{D}$, where $k$ is the number of demos, and construct a prompt input in natural language form $s = f(G, x_q)$ with a pattern $f$. Then, we input $s$ into the LM $P_\theta(\cdot)$ with parameters $\theta$ and get an output token distribution $P_\theta(\cdot|s)$. We choose the label token $l$ with the maximum probability \textbf{among the label space} as the prediction $\hat{y_q}$, that is:
\begin{equation}
    \hat{y_q} = \mathop{\mathrm{argmax}}\limits_{l\in \mathbb{U}}P_\theta(l|s).
\end{equation}
Notice that we only construct prompts to drive the model to predict answers generatively, without any parameter updates. % Such a paradigm is In-context Learning.

\paragraph{\M .} For each parameter matrix $\theta_i$ in the LM $P_\theta(\cdot)$ used for ICL, we simply do an interpolation between $\theta_i$ and a noise matrix sampled from an isotropic normal distribution $\mathcal{N}(0, \sigma^2)$ with intensity $\lambda$, that is:
\begin{equation}
    \theta_i' = (1-\lambda)\theta_i + \lambda \mathcal{N}(0, \sigma^2),
\end{equation}
\noindent where $\lambda$ and $\sigma$ are model or task-specific hyperparameters. Then we perform the aforementioned ICL with the interpolated LM $P_{\theta'}(\cdot)$. % We name this \M.

\begin{figure}[t]
    \centering
    \centerline{\includegraphics[width=\linewidth]{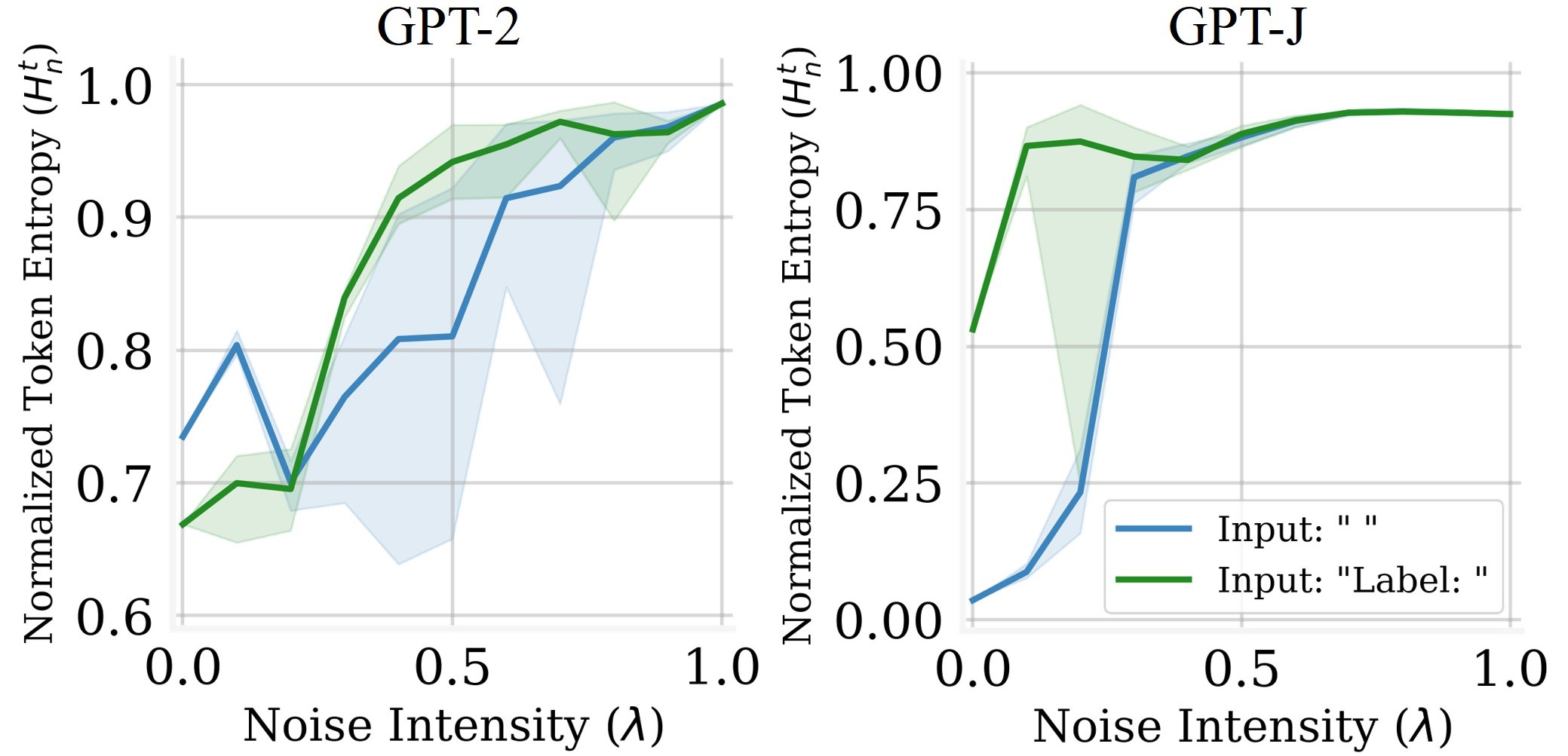}}
    \caption{The correlation between the normalized token entropy $H_n^t$ and the noise intensity $\lambda$ with empty inputs. When the noise gets stronger, the $H_n^t$ becomes higher, which indicates a fairer output.}
    \label{fig:nte}
\end{figure}

\section{Experiments and Results}
\label{sec:3}

We investigate the effectiveness and calibration abilities of \M. We find that \M\ can improve ICL performance by an average of $10\%$ (\S\ref{3.2}) and effectively calibrate the model's prediction bias and unfaithful confidence (\S\ref{3.3}).

\subsection{Settings}

% Here we introduce the datasets, models, and other details of our experiments. 

\paragraph{Data.} We use 12 commonly used classification datasets in our experiments. Since some of the datasets do not provide valid splitting, we randomly split these datasets into validation sets and test sets (see Appendix~\ref{Appendix:Datasets} for further details).

\paragraph{Models.} We use GPT-2 (parameters: 137M) \cite{radford2019language} and GPT-J (parameters: 6053M) \cite{gpt-j}. The model checkpoints are loaded from huggingface\footnote{\url{huggingface.co/gpt2}, and \url{huggingface.co/EleutherAI/gpt-j-6b}}.

\paragraph{Hyperparameters.} We fix $\sigma$, the standard deviation of the noise distribution, to $0.02$, which is the same as the initialization distribution of both models. We believe the noise intensity $\lambda$ should vary w.r.t. datasets and models. Therefore, we determine the most suitable $\lambda$ by a simple search method for each dataset and model on the validation set (details in Appendix~\ref{Appendix:lambda}). The selected intensities are shown in Table~\ref{tab:1}, which are concentrated in $(0,0.3]$. Moreover, we confirm that the $\lambda$ keeps relatively stable w.r.t. $k$ given a dataset and model (Appendix~\ref{Appendix:lambda}).

\paragraph{Other details.} We default to use four demos and a simple prompt template as shown in Appendix~\ref{Appendix:Prompt}. Every labeled data is treated as the test query once, and for each query, we do 2 tries. Each experiment is repeated 10 times with different noise matrices.

\subsection{\M\ Improves ICL Performance}
\label{3.2}

We show accuracy and macro-F1 on the 12 downstream datasets with and without \M{} on the optimal $\lambda$ in Table~\ref{tab:1}, and averaged results in Fig.~\ref{fig:0} (see Appendix~\ref{Appendix:diffk} for results of other numbers of demos ($k$)).

The results show that \M\ produces an average performance improvement of around $10\%$. This phenomenon preliminarily confirms our hypothesis: \M{} fits the pre-trained LMs towards ICL.

However, such gains vary depending on the dataset and model. In some combinations of datasets and models, significant performance improvement cannot be observed. We speculate the reason is that \M{} does not provide new knowledge from new training examples, and these datasets are too hard for the model intrinsically no matter whether \M{} is used.

\subsection{\M\ Is A Calibration}
\label{3.3}

\begin{figure}[t]
    \centering
    \centerline{\includegraphics[width=\linewidth]{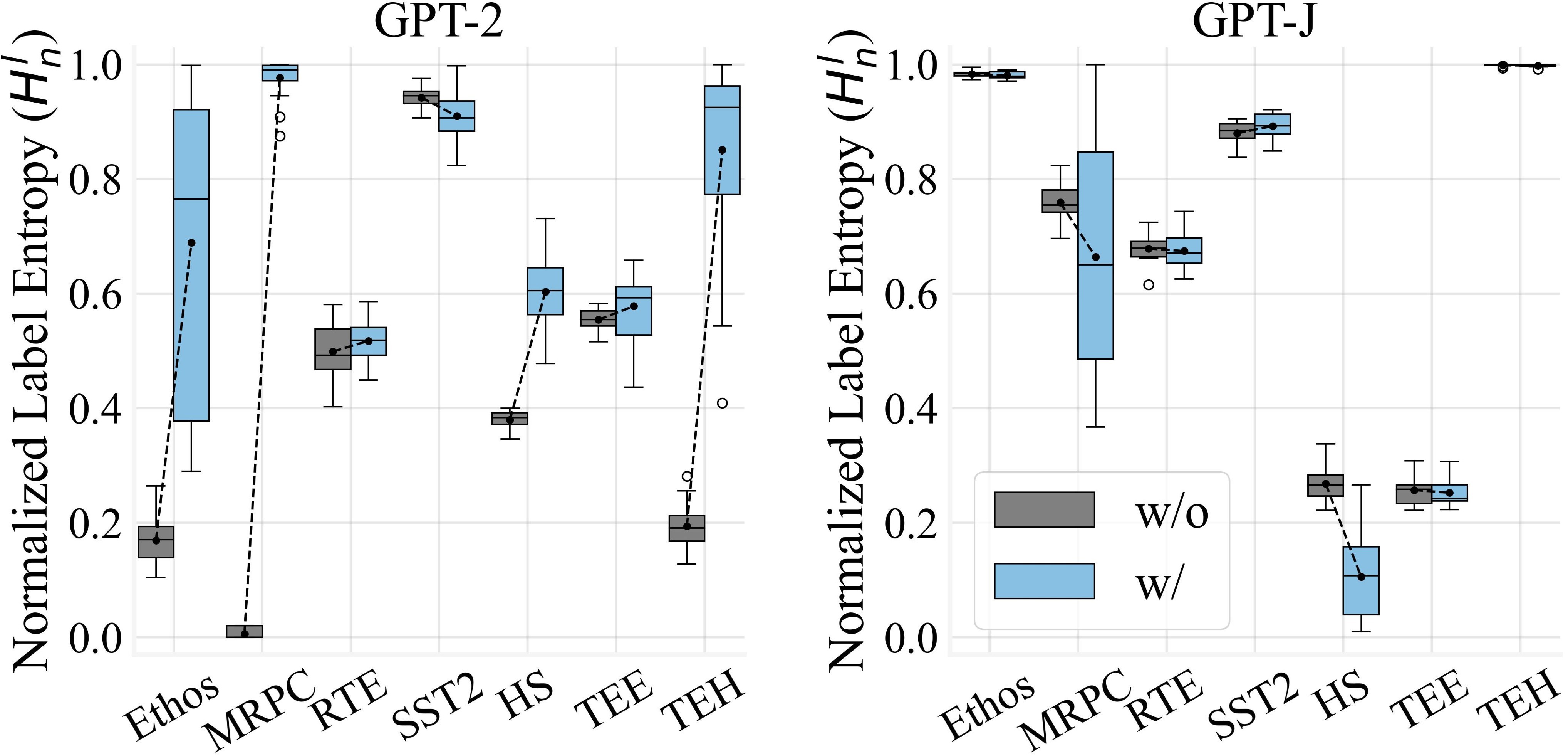}}
    \caption{The normalized label entropy $H_n^l$ on both models and 7 datasets with and without appropriate-noised \M. In most cases, the $H_n^l$ with \M{} (\textbf{w/}) is greater than without \M{} (\textbf{w/o}).}
    \label{fig:nle}
\end{figure}

This section finds that \M\ conducts the following aspects of calibrations:

\begin{itemize}
    \item\textbf{Lower prediction bias.} When no valid query is given, the predicted labels should have balanced frequencies. However, predictions conducted by under-calibrated LMs usually have significant bias, which is harmful to ICL \cite{fei-etal-2023-mitigating, zhao2021calibrate, wei2023symbol, lu2022fantastically}. Eliminating these biases can be seen as an aspect of calibration.
    \item\textbf{More faithful confidence.} In classification tasks, the outputted probability of the predicted label is called confidence. Suitable confidence should faithfully reflect the accuracy of outputs, that is, a prediction with greater accuracy should be assigned with greater confidence \cite{corbiere2019addressing}. It has been proven that faithful confidence improves the stability of the model \cite{guo2017calibration, grabinski2022robust}. Making the confidence more faithful is also an aspect of calibration \cite{guo2017calibration, tian2023just}. 
\end{itemize}

Specifically, we find that \M\ induces the model to predict labels with more fair and faithful confidence. 

% \paragraph{\M\ reduces prediction bias.} We use two metrics to quantify the prediction bias. See Appendix~\ref{Appendix:entropy_calcu} for further details.

% \textbf{(1) Normalized token entropy} $H^t_n$: the entropy of the predicted probability distribution among the whole vocabulary given an empty input (we use ``~'', and ``Label: ''), for detecting the intrinsic token bias \cite{fei-etal-2023-mitigating}.  
% To calculate $H_n^t$, we use two different $x_0$ and various noise intensities.

% The results are shown in Fig.~\ref{fig:nte}. A clear positive correlation between the noise intensity and $H_n^t$ can be observed, which means the model is giving a fairer output when more noise is given.

% \textbf{(2) Normalized label entropy} $H^l_n$: the entropy of the predicted probability distribution among the label space given some demos and an empty query, for detecting the overall bias in the ICL.

% Notice that it is natural for the model to predict a neutral label when no query is given, so the prediction on a dataset with a "neutral" label can be quite benignly biased in this experiment. Therefore, we ignore these datasets. The results are shown in Fig.~\ref{fig:nle}, indicating that in most cases, \M{} calibrates the overall bias, especially on GPT-2.

\paragraph{\M\ reduces prediction bias.} We investigate the following two types of prediction bias. (See Appendix~\ref{Appendix:entropy_calcu} for further calculation details.)

\textbf{(1) Intrinsic token bias.} LMs are pre-trained with natural corpus, where different tokens have various frequencies. LMs learn such frequencies and act as biases among different tokens in prediction, which are deemed harmful to ICL ~\cite{fei-etal-2023-mitigating}. We believe that such token bias can be reduced by \M{}. 

To quantify this, we calculate normalized token entropy $H^t_n$: the entropy of the predicted probability distribution among the whole vocabulary given an empty input (we use ``~'', and ``Label: ''), w.r.t. various noise intensities.

The results are shown in Fig.~\ref{fig:nte}. A clear positive correlation between the noise intensity and $H_n^t$ can be observed, meaning the model gives a fairer output when more noise is given.

\textbf{(2) Overall label bias.} In classification tasks, when no query is given, equal probabilities should be assigned to labels fairly, which benefits ICL ~\cite{lu2022fantastically}. We believe that \M{} promotes such fairness. 

To quantify this, we use normalized label entropy $H^l_n$: the entropy of the predicted probability distribution among the label space given some demos and an empty query. Notice that it is natural for a model to predict a ``neutral'' label when no query is given, so label bias on a dataset with a ``neutral'' label cannot be fairly measured with $H^l_n$. Therefore, we ignore these datasets with ``neutral'' labels (details in Appendix~\ref{Appendix:Datasets}). 

The results are shown in Fig.~\ref{fig:nle}, indicating that in most cases, \M{} calibrates the overall bias, especially on GPT-2.

\paragraph{\M\ promotes faithful confidence.} The \textbf{E}xpected \textbf{C}alibration \textbf{E}rror ($ECE_p$) \cite{naeini2015obtaining} is a widely-used indicator for the faithfulness of confidence in classification tasks:
\begin{equation}
    ECE_p = \mathbb{E}(\vert \mathop{\mathrm{max}}(\hat{z}) - \mathbb{E}_{y=\mathop{\mathrm{argmax}}\limits_{i}\hat{z_i}}(1) \vert^p)^{\frac{1}{p}},
\end{equation}
\noindent where $\hat{z}$ is the predicted probability vector on the label space by a classification model, the final prediction $(\mathop{\mathrm{argmax}}_{i}\hat{z_i})$ can be obtained with a confidence $(\mathop{\mathrm{max}}\hat{z})$, and the ground-truth label is $y$.

\begin{figure}[t]
    \centering
    \centerline{\includegraphics[width=\linewidth]{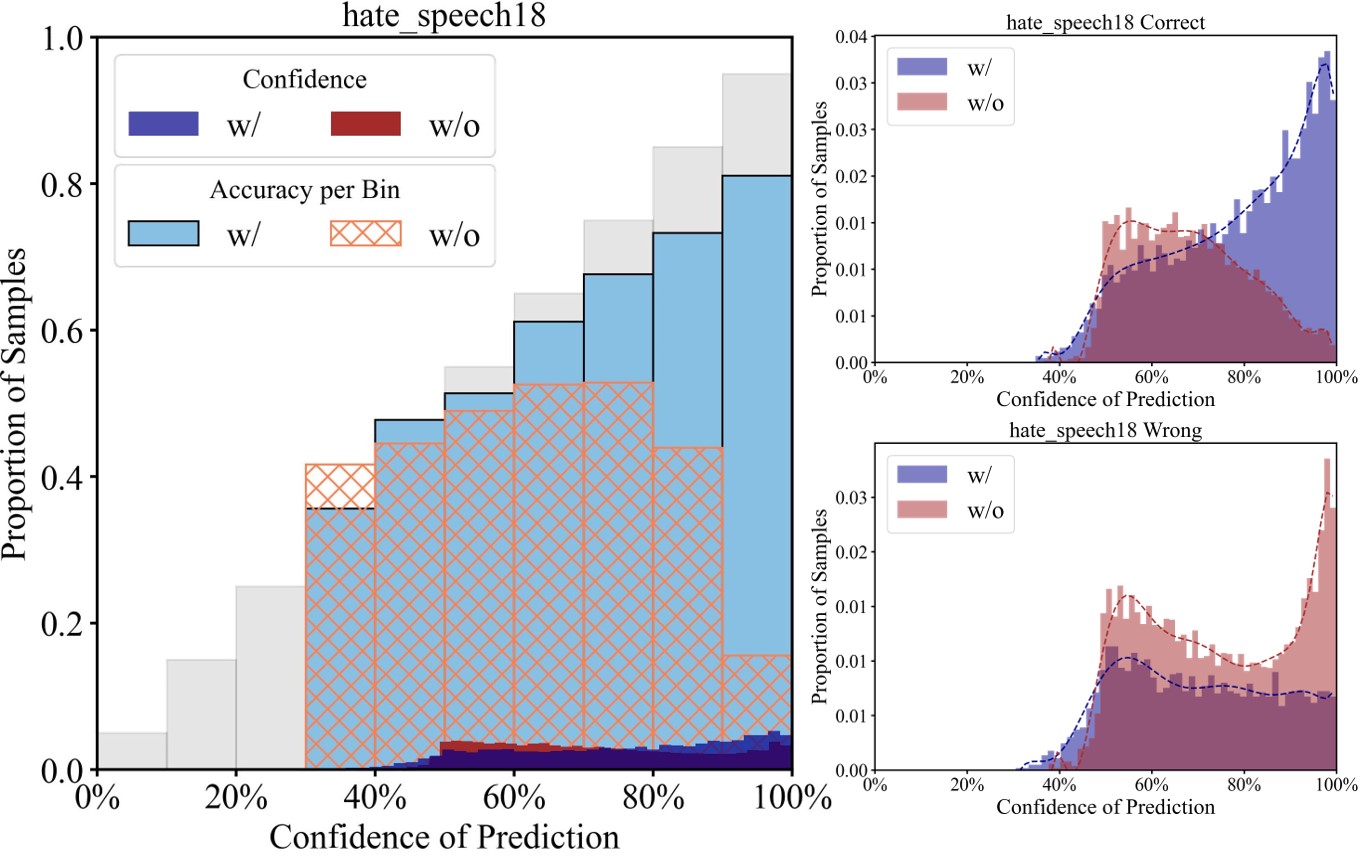}}
    \caption{\textbf{Left}: Reliability diagrams (sparse bars) and global confidence distribution (dense bars) of GPT-2 on hate\_speech18 with (\textbf{w/}, $ECE_1 = 9.01\%$) and without (\textbf{w/o}, $ECE_1 = 28.14\%$) \M. The predictions are divided into bins according to confidence, and we visualize the accuracy of each bin as a histogram. The grey bars are ideal, which is closer to the one with \M{}. \textbf{Right upper}: Confidence distribution on correct predictions. Relatively right-shifted with \M{}. \textbf{Right lower}: Confidence distribution on wrong predictions. Relatively left-shifted with \M{}. ($k=4$)}
    \label{fig:confexam}
\end{figure}

Let $p = 1$, we use the $ECE_1$ to investigate the faithfulness of the ICL output. The details of the calculation are shown in Appendix \ref{Appendix:ece1}. A lower $ECE_1$ means more faithful confidence, that is, the confidence becomes a more accurate prediction of accuracy \cite{corbiere2019addressing}. Fig.~\ref{fig:confexam} shows a case study of GPT-2 on the hate\_speech18 dataset, indicating that the prediction is more faithful with \M{} (full visualized data are in Appendix~\ref{Appendix:confd}). We show the results with and without appropriate-noised \M\ for $ECE_1$ on the 12 datasets in Table~\ref{tab:1}. 

In most cases, the $ECE_1$ is reduced by around $25\%$ by \M{}, which means the confidence is more faithful with \M. This suggests that \M\ can drive the model to produce more faithful confidence, that is, less over-confidence in wrong predictions, and less under-confidence in correct predictions, which suggests that \M\ solves the confidence calibration.

Based on the above two investigations, we believe that \M{} is an effective calibration for the ICL scenario. This can be a reasonable explanation for the observed performance improvement in \S\ref{3.2}.

\section{Related Works}

\paragraph{Fine-tuning and calibrating LMs to ICL.} There have been some efforts in fine-tuning and calibrating LMs for better ICL performance. \citet{min2022metaicl} proposed MetaICL, fine-tuning the LM on ICL-style examples sampled from a huge set of downstream tasks. \citet{zhao2021calibrate} suggested that ICL is suffering from high variance due to the under-calibration of LMs. Therefore, they proposed a contextual calibration, calibrating the model output to the uniform distribution when no queries are given. And \citet{wei2021finetuned} proposed FLAN, an instruction fine-tuning methodology to improve both the few-shot and zero-shot ICL performance of LMs. All of these methods need parameter updating to LMs on huge datasets by gradient, which is quite expensive with large models. Moreover, \citet{lu2022fantastically} tried to propose a method to confirm the best demo order for less output bias and less over-confidence.

\paragraph{Adding noise to pre-trained models for fine-tuning.} In the previous pre-training and fine-tuning paradigm, some scholars pointed out that adding noise to model parameters is beneficial to fine-tuning. \citet{zhang2020revisiting} re-initialized some layers of BERT, that is, making some layers full-noised, and found that it is beneficial to few-shot fine-tuning. \citet{wu2022noisytune} directly added noise to pre-trained Transformers, and got results with slight improvements.

\section{Conclusion}

In this paper, we propose \M, which simply adds random noise to the parameters of LMs to fit them from pre-trained knowledge to ICL. We show that \M\ can improve the ICL performance and also calibrate the model for fairer outputs and more faithful confidence.

\section{Limitations}

As mentioned before, unlike the fine-tuning on additional ICL-style datasets \cite{min2022metaicl, zhao2021calibrate, wei2021finetuned}, \M\ does not provide new knowledge for the model, so the calibrated model can not discover tasks that are not potentially included in the pre-training data \cite{gu2023pre}. Moreover, a naive search for the best noise intensity is not efficient and satisfactory. An effective detection of $\lambda$ should be developed. Currently, we can only confirm that for one dataset and model, the $\lambda$ keeps relatively stable, especially when a large $k$ is given (Appendix. \ref{Appendix:lambda}).

\paragraph{Future Works.} Besides fixing the limits, future works can focus on where and how the noise should be introduced. In Transformer-based models, different layers have different abilities \cite{wang-etal-2023-label, jawahar2019does, kobayashi-etal-2023-transformer}. So, treating these layers differently may be an effective improvement of \M. An isotropic normal distribution is too simple, noise sampling methods should be discussed.

Moreover, adding noise to model parameters can be an erasing of pre-training \cite{ilharco2022editing}, so, the search for $\lambda$ is the search for the best checkpoint of pre-training. With these checkpoints, we can determine \cite{han2023understanding, han2022orca} which data is disadvantageous to ICL, and what knowledge is essential to ICL, to better reveal the essence of ICL.

\section*{Acknowledgements}

This work was supported by JSPS KAKENHI Grant Number 19K20332.

\noindent The authors would like to thank Mr. Daichi Haraguchi at JAIST, Mr. Ji’an Liu, Mr. Kai Kang at Beijing Institute of Technology, and MSc. Tongyuan Wei at the University of Cambridge, for their proofreading and constructive criticism.

\bibliography{custom}

\appendix

\begin{table*}[t] 
    \centering
    \caption{Datasets and Abbreviations used in this paper. \textbf{Abbr.}: Abbreviation, \textbf{Data\#}: The total number of data, \textbf{Label\#}: The number of classes, \textbf{Neutral?}: Does the dataset have a neutral label?, \textbf{Major. $\sim$ Minor.(\%)}: The proportion of majority labels $\sim$ the proportion of minority labels.}
    \label{tab:dataset}
    \resizebox{2\columnwidth}{!}{
    \begin{tabular}{lccccc}
    \toprule
      Dataset & Abbr. & Data\# & Label\# & Neutral? & Major. $\sim$  Minor.(\%)\\
    \midrule
      \multicolumn{5}{l}{\textit{single-sentence classification:}} \\
      \quad poem\_sentiment \cite{sheng2020investigating} & PS & 1101 & 4 & \checkmark & $62.2 \sim 5.5$ \\
      \quad hate\_speech18 \cite{gibert2018hate} & HS & 10944 & 4 & $\times$ & $86.9 \sim 1.5$ \\
      \quad Ethos* (binary) \cite{mollas2020ethos} & --- & 980 & 2 & $\times$ & $56.6 \sim 43.4$ \\
      \quad financial\_phrasebank (all agree) \cite{Malo2014GoodDO} & FP & 2264 & 3 & \checkmark & $61.4 \sim 13.4$ \\
      \quad GLUE-SST2 \cite{wang2019glue, SST2} & SST2 & 68221 & 2 & $\times$ & $55.8 \sim 44.2$ \\
      \quad tweet\_eval\_emotion \cite{mohammad2018semeval} & TEE & 5052 & 4 & $\times$ & $43 \sim 9$ \\
      \quad tweet\_eval\_sentiment \cite{rosenthal2017semeval} & TES & 59899 & 3 & \checkmark & $45.3 \sim 15.5$ \\
      \quad tweet\_eval\_hate \cite{basile-etal-2019-semeval} & TEH & 12970 & 2 & $\times$ & $58 \sim 42$ \\
      \multicolumn{3}{l}{\textit{aspect-based sentiment classification:}} \\
      \quad SemEval 2014-Task 4 Restaurants \cite{pontikietal2014semeval} & SE'14R & 4722 & 3 & \checkmark & $61.2 \sim 17.6$ \\
      \quad SemEval 2014-Task 4 Laptops \cite{pontikietal2014semeval} & SE'14L & 2951 & 3 & \checkmark & $45.3 \sim 15.5$ \\
      \multicolumn{3}{l}{\textit{double-sentence classification:}} \\
      \quad GLUE-RTE \cite{wang2019glue, RTE1} & RTE & 2767 & 2 & $\times$ & $50.2 \sim 49.8$ \\ \multicolumn{3}{l}{\quad \quad \quad \cite{RTE2, RTE3, RTE5}}\\
      \quad GLUE-MRPC \cite{wang2019glue, mrpc} & MRPC & 4076 & 2 & $\times$ & $67.4 \sim 32.6$\\
    \bottomrule
    \end{tabular}}
\end{table*}

\section{Datasets}
\label{Appendix:Datasets}

The datasets used in this paper are shown in Table~\ref{tab:dataset}.

\noindent *To construct inputs of appropriate length, we remove data with a length exceeding 500 from the Ethos, and the number of the remaining data is 980.

\paragraph{Dataset Splitting.} For each dataset, we first shuffle it with random seed 42. Then, we choose the 512 data at the tail as the testing data, and the 512 data at the head (if the total number of data is less than 1024, we choose the rest of the data.) as the validation data.

\section{Prompt Patterns}
\label{Appendix:Prompt}

In this paper, we use a minimum prompt template. For each task category, we design a template as shown below.

For single-sentence classification datasets ${(x, y)}$, we use:

\begin{verbatim}
   Input: <x>, Label: <y> \n
   ...
   Input: <x>, Label: 
\end{verbatim}

For aspect-based sentiment classification datasets ${((x, a), y)}$, we use:

\begin{verbatim}
   Input: <x>, Aspect: <a>, Label: <y> \n
   ...
   Input: <x>, Aspect: <a>, Label:
\end{verbatim}

For double-sentence classification datasets ${((x_1, x_2), y)}$. we use:

\begin{verbatim}
   Input: <x1>, Text 2: <x2>, Label: <y> \n
   ...
   Input: <x1>, Text 2: <x2>, Label:
\end{verbatim}

\section{Complete Results: Accuracy, Macro-F1, and $ECE_1$ With Various Demo Numbers ($k$)} 
\label{Appendix:diffk}

We use various numbers of demos in the experiments shown in the \S\ref{3.2}.

\noindent The results of zero-shot ($k=0$) are shown in Table~\ref{tab:K0}.

\noindent The results of 1-shot ($k=1$) are shown in Table~\ref{tab:K1}.

\noindent The results of 2-shot ($k=2$) are shown in Table~\ref{tab:K2}.

\noindent The results of 8-shot ($k=8$) are shown in Table~\ref{tab:K8}.

\noindent The results of 16-shot ($k=16$) are shown in Table~\ref{tab:K16}.

Notice that in Table~\ref{tab:K8} and Table~\ref{tab:K16}, some experiments are not feasible due to the length of sequences.

In these results, \M{} still outperforms the baseline, which proves that \M{} is generally applicable with various $k$.
 
\begin{table*}[t]
\caption{Accuracy and Macro-F1 results ($\%$, $mean_{std}$, $k=0$). A better result is in {\color[HTML]{00ba55}green}. Notation is the same with the Tabel~\ref{tab:1}.}
\label{tab:K0}
\centering
\resizebox{2\columnwidth}{!}{
\begin{tabular}{@{}cccc|cccccccccccc|c@{}}
\toprule
\multicolumn{4}{c|}{{\color[HTML]{000000} Dataset}} & {\color[HTML]{000000} PS} & {\color[HTML]{000000} HS} & {\color[HTML]{000000} SE'14R} & {\color[HTML]{000000} SE'14L} & {\color[HTML]{000000} RTE} & {\color[HTML]{000000} MRPC} & {\color[HTML]{000000} Ethos} & {\color[HTML]{000000} FP} & {\color[HTML]{000000} SST2} & {\color[HTML]{000000} TEE} & {\color[HTML]{000000} TES} & {\color[HTML]{000000} TEH} & {\color[HTML]{000000} Mean} \\ \midrule
{\color[HTML]{000000} } & \multicolumn{3}{c|}{{\color[HTML]{000000} $\lambda$}} & {\color[HTML]{000000} 0.1} & {\color[HTML]{000000} 0.1} & {\color[HTML]{000000} 0.1} & {\color[HTML]{000000} 0.1} & {\color[HTML]{000000} 0.06} & {\color[HTML]{000000} 0.02} & {\color[HTML]{000000} 0.004} & {\color[HTML]{000000} 0.02} & {\color[HTML]{000000} 0.1} & {\color[HTML]{000000} 0.2} & {\color[HTML]{000000} 0.008} & {\color[HTML]{000000} 0.3} & {\color[HTML]{000000} ---} \\ \cmidrule(l){2-17} 
{\color[HTML]{000000} } & {\color[HTML]{000000} } & {\color[HTML]{000000} } & {\color[HTML]{000000} w/o} & {\color[HTML]{000000} ${16.60}_{0.00}$} & {\color[HTML]{000000} ${37.30}_{0.00}$} & {\color[HTML]{000000} ${33.98}_{0.00}$} & {\color[HTML]{000000} ${38.87}_{0.00}$} & {\color[HTML]{000000} ${52.54}_{0.00}$} & {\color[HTML]{000000} ${65.82}_{0.00}$} & {\color[HTML]{000000} ${51.07}_{0.00}$} & {\color[HTML]{000000} ${61.13}_{0.00}$} & {\color[HTML]{000000} ${57.42}_{0.00}$} & {\color[HTML]{00ba55} ${39.84}_{0.00}$} & {\color[HTML]{000000} ${49.22}_{0.00}$} & {\color[HTML]{000000} ${43.36}_{0.00}$} & {\color[HTML]{000000} $45.60$} \\
{\color[HTML]{000000} } & {\color[HTML]{000000} } & \multirow{-2}{*}{{\color[HTML]{000000} Val.}} & {\color[HTML]{000000} w/} & {\color[HTML]{00ba55} ${21.88}_{6.75}$} & {\color[HTML]{00ba55} ${50.70}_{4.33}$} & {\color[HTML]{00ba55} ${40.74}_{10.27}$} & {\color[HTML]{00ba55} ${45.94}_{2.72}$} & {\color[HTML]{00ba55} ${53.40}_{0.55}$} & {\color[HTML]{00ba55} ${67.77}_{0.55}$} & {\color[HTML]{00ba55} ${51.07}_{0.19}$} & {\color[HTML]{00ba55} ${61.37}_{0.60}$} & {\color[HTML]{00ba55} ${60.98}_{3.87}$} & {\color[HTML]{000000} ${39.57}_{3.50}$} & {\color[HTML]{00ba55} ${49.53}_{0.20}$} & {\color[HTML]{00ba55} ${54.14}_{5.91}$} & {\color[HTML]{00ba55} $49.76$} \\ \cmidrule(l){3-17} 
{\color[HTML]{000000} } & {\color[HTML]{000000} } & {\color[HTML]{000000} } & {\color[HTML]{000000} w/o} & {\color[HTML]{000000} ${17.38}_{0.00}$} & {\color[HTML]{000000} ${29.88}_{0.00}$} & {\color[HTML]{000000} ${28.91}_{0.00}$} & {\color[HTML]{000000} ${39.45}_{0.00}$} & {\color[HTML]{00ba55} ${49.80}_{0.00}$} & {\color[HTML]{000000} ${59.18}_{0.00}$} & {\color[HTML]{000000} ${54.10}_{0.00}$} & {\color[HTML]{00ba55} ${61.91}_{0.00}$} & {\color[HTML]{00ba55} ${66.41}_{0.00}$} & {\color[HTML]{00ba55} ${41.60}_{0.00}$} & {\color[HTML]{00ba55} ${49.22}_{0.00}$} & {\color[HTML]{000000} ${46.88}_{0.00}$} & {\color[HTML]{000000} $45.39$} \\
{\color[HTML]{000000} } & \multirow{-4}{*}{{\color[HTML]{000000} \begin{tabular}[c]{@{}c@{}}Acc.\\ (\%)\end{tabular}}} & \multirow{-2}{*}{{\color[HTML]{000000} Test}} & w/ & \cellcolor[HTML]{EFEFEF}{\color[HTML]{00ba55} ${22.97}_{7.52}$} & \cellcolor[HTML]{EFEFEF}{\color[HTML]{00ba55} ${38.30}_{10.89}$} & \cellcolor[HTML]{EFEFEF}{\color[HTML]{00ba55} ${31.25}_{5.79}$} & \cellcolor[HTML]{EFEFEF}{\color[HTML]{00ba55} ${40.84}_{3.85}$} & \cellcolor[HTML]{EFEFEF}{\color[HTML]{000000} ${49.77}_{1.46}$} & \cellcolor[HTML]{EFEFEF}{\color[HTML]{00ba55} ${59.38}_{0.79}$} & \cellcolor[HTML]{EFEFEF}{\color[HTML]{00ba55} ${54.36}_{0.15}$} & \cellcolor[HTML]{EFEFEF}{\color[HTML]{000000} ${61.37}_{0.65}$} & \cellcolor[HTML]{EFEFEF}{\color[HTML]{000000} ${65.96}_{2.27}$} & \cellcolor[HTML]{EFEFEF}{\color[HTML]{000000} ${41.15}_{3.79}$} & \cellcolor[HTML]{EFEFEF}{\color[HTML]{343434} ${49.14}_{0.26}$} & \cellcolor[HTML]{EFEFEF}{\color[HTML]{00ba55} ${49.86}_{8.37}$} & \cellcolor[HTML]{EFEFEF}{\color[HTML]{00ba55} $47.03$} \\ \cmidrule(l){2-17} 
{\color[HTML]{000000} } & {\color[HTML]{000000} } & {\color[HTML]{000000} } & {\color[HTML]{000000} w/o} & {\color[HTML]{000000} ${11.32}_{0.00}$} & {\color[HTML]{000000} ${20.81}_{0.00}$} & {\color[HTML]{000000} ${35.19}_{0.00}$} & {\color[HTML]{000000} ${36.56}_{0.00}$} & {\color[HTML]{000000} ${46.42}_{0.00}$} & {\color[HTML]{000000} ${48.13}_{0.00}$} & {\color[HTML]{000000} ${48.58}_{0.00}$} & {\color[HTML]{000000} ${36.13}_{0.00}$} & {\color[HTML]{000000} ${56.98}_{0.00}$} & {\color[HTML]{000000} ${15.99}_{0.00}$} & {\color[HTML]{000000} ${29.81}_{0.00}$} & {\color[HTML]{000000} ${33.40}_{0.00}$} & {\color[HTML]{000000} $34.94$} \\
{\color[HTML]{000000} } & {\color[HTML]{000000} } & \multirow{-2}{*}{{\color[HTML]{000000} Val.}} & {\color[HTML]{000000} w/} & {\color[HTML]{00ba55} ${14.87}_{2.10}$} & {\color[HTML]{00ba55} ${25.82}_{2.65}$} & {\color[HTML]{00ba55} ${39.90}_{8.62}$} & {\color[HTML]{00ba55} ${41.36}_{3.98}$} & {\color[HTML]{00ba55} ${50.10}_{1.31}$} & {\color[HTML]{00ba55} ${49.08}_{0.54}$} & {\color[HTML]{00ba55} ${48.58}_{0.25}$} & {\color[HTML]{00ba55} ${37.35}_{0.74}$} & {\color[HTML]{00ba55} ${59.80}_{4.50}$} & {\color[HTML]{00ba55} ${25.30}_{2.60}$} & {\color[HTML]{00ba55} ${30.36}_{0.60}$} & {\color[HTML]{00ba55} ${39.66}_{6.23}$} & {\color[HTML]{00ba55} $38.62$} \\ \cmidrule(l){3-17} 
{\color[HTML]{000000} } & {\color[HTML]{000000} } & {\color[HTML]{000000} } & {\color[HTML]{000000} w/o} & {\color[HTML]{000000} ${12.70}_{0.00}$} & {\color[HTML]{000000} ${16.02}_{0.00}$} & {\color[HTML]{000000} ${27.01}_{0.00}$} & {\color[HTML]{000000} ${39.38}_{0.00}$} & {\color[HTML]{000000} ${45.56}_{0.00}$} & {\color[HTML]{00ba55} ${40.48}_{0.00}$} & {\color[HTML]{000000} ${50.99}_{0.00}$} & {\color[HTML]{000000} ${34.91}_{0.00}$} & {\color[HTML]{00ba55} ${64.45}_{0.00}$} & {\color[HTML]{000000} ${16.26}_{0.00}$} & {\color[HTML]{00ba55} ${29.81}_{0.00}$} & {\color[HTML]{00ba55} ${42.58}_{0.00}$} & {\color[HTML]{000000} $35.85$} \\
{\color[HTML]{000000} } & \multirow{-4}{*}{{\color[HTML]{000000} \begin{tabular}[c]{@{}c@{}}MF1\\ (\%)\end{tabular}}} & \multirow{-2}{*}{{\color[HTML]{000000} Test}} & w/ & \cellcolor[HTML]{EFEFEF}{\color[HTML]{00ba55} ${16.93}_{4.12}$} & \cellcolor[HTML]{EFEFEF}{\color[HTML]{00ba55} ${17.92}_{3.40}$} & \cellcolor[HTML]{EFEFEF}{\color[HTML]{00ba55} ${30.20}_{7.55}$} & \cellcolor[HTML]{EFEFEF}{\color[HTML]{00ba55} ${40.50}_{4.43}$} & \cellcolor[HTML]{EFEFEF}{\color[HTML]{00ba55} ${49.06}_{2.03}$} & \cellcolor[HTML]{EFEFEF}{\color[HTML]{000000} ${39.96}_{0.59}$} & \cellcolor[HTML]{EFEFEF}{\color[HTML]{00ba55} ${51.20}_{0.12}$} & \cellcolor[HTML]{EFEFEF}{\color[HTML]{00ba55} ${35.56}_{1.38}$} & \cellcolor[HTML]{EFEFEF}{\color[HTML]{000000} ${61.74}_{5.39}$} & \cellcolor[HTML]{EFEFEF}{\color[HTML]{00ba55} ${22.54}_{5.12}$} & \cellcolor[HTML]{EFEFEF}{\color[HTML]{000000} ${29.69}_{0.59}$} & \cellcolor[HTML]{EFEFEF}{\color[HTML]{000000} ${39.70}_{8.87}$} & \cellcolor[HTML]{EFEFEF}{\color[HTML]{00ba55} $36.25$} \\ \cmidrule(l){2-17} 
{\color[HTML]{000000} } & {\color[HTML]{000000} } & {\color[HTML]{000000} } & {\color[HTML]{000000} w/o} & {\color[HTML]{000000} ${36.56}_{0.00}$} & {\color[HTML]{000000} ${22.03}_{0.00}$} & {\color[HTML]{000000} ${28.11}_{0.00}$} & {\color[HTML]{000000} ${17.80}_{0.00}$} & {\color[HTML]{000000} ${29.83}_{0.00}$} & {\color[HTML]{00ba55} ${7.24}_{0.00}$} & {\color[HTML]{00ba55} ${28.08}_{0.00}$} & {\color[HTML]{000000} ${13.45}_{0.00}$} & {\color[HTML]{000000} ${5.45}_{0.00}$} & {\color[HTML]{000000} ${48.09}_{0.00}$} & {\color[HTML]{00ba55} ${22.10}_{0.00}$} & {\color[HTML]{000000} ${46.21}_{0.00}$} & {\color[HTML]{000000} $25.41$} \\
{\color[HTML]{000000} } & {\color[HTML]{000000} } & \multirow{-2}{*}{{\color[HTML]{000000} Val.}} & {\color[HTML]{000000} w/} & {\color[HTML]{00ba55} ${22.52}_{6.40}$} & {\color[HTML]{00ba55} ${16.20}_{3.75}$} & {\color[HTML]{00ba55} ${16.77}_{9.48}$} & {\color[HTML]{00ba55} ${8.39}_{3.82}$} & {\color[HTML]{00ba55} ${23.90}_{1.51}$} & {\color[HTML]{000000} ${9.20}_{0.54}$} & {\color[HTML]{000000} ${28.12}_{0.24}$} & {\color[HTML]{00ba55} ${12.75}_{0.37}$} & {\color[HTML]{00ba55} ${4.22}_{2.43}$} & {\color[HTML]{00ba55} ${18.66}_{2.88}$} & {\color[HTML]{000000} ${22.53}_{0.19}$} & {\color[HTML]{00ba55} ${21.99}_{12.15}$} & {\color[HTML]{00ba55} $17.10$} \\ \cmidrule(l){3-17} 
{\color[HTML]{000000} } & {\color[HTML]{000000} } & {\color[HTML]{000000} } & {\color[HTML]{000000} w/o} & {\color[HTML]{000000} ${35.33}_{0.00}$} & {\color[HTML]{000000} ${26.92}_{0.00}$} & {\color[HTML]{000000} ${33.41}_{0.00}$} & {\color[HTML]{000000} ${16.92}_{0.00}$} & {\color[HTML]{000000} ${31.46}_{0.00}$} & {\color[HTML]{000000} ${13.80}_{0.00}$} & {\color[HTML]{000000} ${24.56}_{0.00}$} & {\color[HTML]{000000} ${15.09}_{0.00}$} & {\color[HTML]{00ba55} ${3.49}_{0.00}$} & {\color[HTML]{000000} ${46.49}_{0.00}$} & {\color[HTML]{00ba55} ${22.10}_{0.00}$} & {\color[HTML]{000000} ${36.32}_{0.00}$} & {\color[HTML]{000000} $25.49$} \\
\multirow{-13}{*}{{\color[HTML]{000000} GPT-J}} & \multirow{-4}{*}{{\color[HTML]{000000} \begin{tabular}[c]{@{}c@{}}$ECE_1$\\ (\%, $\downarrow$)\end{tabular}}} & \multirow{-2}{*}{{\color[HTML]{000000} Test}} & w/ & \cellcolor[HTML]{EFEFEF}{\color[HTML]{00ba55} ${24.86}_{6.34}$} & \cellcolor[HTML]{EFEFEF}{\color[HTML]{00ba55} ${23.67}_{6.86}$} & \cellcolor[HTML]{EFEFEF}{\color[HTML]{00ba55} ${26.19}_{10.00}$} & \cellcolor[HTML]{EFEFEF}{\color[HTML]{00ba55} ${12.16}_{5.22}$} & \cellcolor[HTML]{EFEFEF}{\color[HTML]{00ba55} ${24.43}_{2.49}$} & \cellcolor[HTML]{EFEFEF}{\color[HTML]{00ba55} ${13.65}_{0.60}$} & \cellcolor[HTML]{EFEFEF}{\color[HTML]{00ba55} ${24.49}_{0.13}$} & \cellcolor[HTML]{EFEFEF}{\color[HTML]{00ba55} ${13.34}_{0.90}$} & \cellcolor[HTML]{EFEFEF}{\color[HTML]{000000} ${5.91}_{2.43}$} & \cellcolor[HTML]{EFEFEF}{\color[HTML]{00ba55} ${25.21}_{14.36}$} & \cellcolor[HTML]{EFEFEF}{\color[HTML]{000000} ${22.37}_{0.36}$} & \cellcolor[HTML]{EFEFEF}{\color[HTML]{00ba55} ${24.95}_{16.80}$} & \cellcolor[HTML]{EFEFEF}{\color[HTML]{00ba55} $20.10$} \\ \midrule
{\color[HTML]{000000} } & \multicolumn{3}{c|}{{\color[HTML]{000000} $\lambda$}} & {\color[HTML]{000000} 0.04} & {\color[HTML]{000000} 0.002} & {\color[HTML]{000000} 0.1} & {\color[HTML]{000000} 0.1} & {\color[HTML]{000000} 0.012} & {\color[HTML]{000000} 0.06} & {\color[HTML]{000000} 0.1} & {\color[HTML]{000000} 0.06} & {\color[HTML]{000000} 0.02} & {\color[HTML]{000000} 0.02} & {\color[HTML]{000000} 0.002} & {\color[HTML]{000000} 0.1} & {\color[HTML]{000000} ---} \\ \cmidrule(l){2-17} 
{\color[HTML]{000000} } & {\color[HTML]{000000} } & {\color[HTML]{000000} } & {\color[HTML]{000000} w/o} & {\color[HTML]{000000} ${4.30}_{0.00}$} & {\color[HTML]{00ba55} ${15.43}_{0.00}$} & {\color[HTML]{000000} ${23.44}_{0.00}$} & {\color[HTML]{000000} ${39.84}_{0.00}$} & {\color[HTML]{000000} ${49.41}_{0.00}$} & {\color[HTML]{00ba55} ${69.53}_{0.00}$} & {\color[HTML]{000000} ${44.23}_{0.00}$} & {\color[HTML]{000000} ${43.55}_{0.00}$} & {\color[HTML]{000000} ${60.55}_{0.00}$} & {\color[HTML]{000000} ${32.03}_{0.00}$} & {\color[HTML]{00ba55} ${46.68}_{0.00}$} & {\color[HTML]{000000} ${41.99}_{0.00}$} & {\color[HTML]{000000} $39.25$} \\
{\color[HTML]{000000} } & {\color[HTML]{000000} } & \multirow{-2}{*}{{\color[HTML]{000000} Val.}} & {\color[HTML]{000000} w/} & {\color[HTML]{00ba55} ${5.00}_{0.50}$} & {\color[HTML]{000000} ${15.23}_{0.00}$} & {\color[HTML]{00ba55} ${40.90}_{15.99}$} & {\color[HTML]{00ba55} ${40.35}_{5.80}$} & {\color[HTML]{00ba55} ${50.82}_{0.42}$} & {\color[HTML]{000000} ${63.52}_{2.09}$} & {\color[HTML]{00ba55} ${55.98}_{1.93}$} & {\color[HTML]{00ba55} ${58.13}_{2.45}$} & {\color[HTML]{00ba55} ${63.87}_{0.96}$} & {\color[HTML]{00ba55} ${38.71}_{0.57}$} & {\color[HTML]{000000} ${46.41}_{0.16}$} & {\color[HTML]{00ba55} ${49.38}_{3.75}$} & {\color[HTML]{00ba55} $44.02$} \\ \cmidrule(l){3-17} 
{\color[HTML]{000000} } & {\color[HTML]{000000} } & {\color[HTML]{000000} } & {\color[HTML]{000000} w/o} & {\color[HTML]{000000} ${5.86}_{0.00}$} & {\color[HTML]{00ba55} ${10.74}_{0.00}$} & {\color[HTML]{000000} ${22.27}_{0.00}$} & {\color[HTML]{000000} ${35.35}_{0.00}$} & {\color[HTML]{00ba55} ${48.63}_{0.00}$} & {\color[HTML]{00ba55} ${63.48}_{0.00}$} & {\color[HTML]{000000} ${46.88}_{0.00}$} & {\color[HTML]{000000} ${45.51}_{0.00}$} & {\color[HTML]{00ba55} ${64.84}_{0.00}$} & {\color[HTML]{000000} ${31.25}_{0.00}$} & {\color[HTML]{00ba55} ${37.50}_{0.00}$} & {\color[HTML]{000000} ${40.23}_{0.00}$} & {\color[HTML]{000000} $37.71$} \\
{\color[HTML]{000000} } & \multirow{-4}{*}{{\color[HTML]{000000} \begin{tabular}[c]{@{}c@{}}Acc.\\ (\%)\end{tabular}}} & \multirow{-2}{*}{{\color[HTML]{000000} Test}} & w/ & \cellcolor[HTML]{EFEFEF}{\color[HTML]{00ba55} ${6.52}_{0.61}$} & \cellcolor[HTML]{EFEFEF}{\color[HTML]{000000} ${9.39}_{0.20}$} & \cellcolor[HTML]{EFEFEF}{\color[HTML]{00ba55} ${41.89}_{11.21}$} & \cellcolor[HTML]{EFEFEF}{\color[HTML]{00ba55} ${37.19}_{1.48}$} & \cellcolor[HTML]{EFEFEF}{\color[HTML]{000000} ${47.95}_{0.57}$} & \cellcolor[HTML]{EFEFEF}{\color[HTML]{000000} ${58.81}_{1.26}$} & \cellcolor[HTML]{EFEFEF}{\color[HTML]{00ba55} ${53.89}_{1.52}$} & \cellcolor[HTML]{EFEFEF}{\color[HTML]{00ba55} ${46.45}_{0.29}$} & \cellcolor[HTML]{EFEFEF}{\color[HTML]{000000} ${59.39}_{3.25}$} & \cellcolor[HTML]{EFEFEF}{\color[HTML]{00ba55} ${41.48}_{0.61}$} & \cellcolor[HTML]{EFEFEF}{\color[HTML]{000000} ${37.21}_{0.22}$} & \cellcolor[HTML]{EFEFEF}{\color[HTML]{00ba55} ${61.04}_{1.07}$} & \cellcolor[HTML]{EFEFEF}{\color[HTML]{00ba55} $41.77$} \\ \cmidrule(l){2-17} 
{\color[HTML]{000000} } & {\color[HTML]{000000} } & {\color[HTML]{000000} } & {\color[HTML]{000000} w/o} & {\color[HTML]{000000} ${4.70}_{0.00}$} & {\color[HTML]{00ba55} ${9.58}_{0.00}$} & {\color[HTML]{000000} ${24.11}_{0.00}$} & {\color[HTML]{00ba55} ${36.57}_{0.00}$} & {\color[HTML]{000000} ${47.91}_{0.00}$} & {\color[HTML]{000000} ${41.63}_{0.00}$} & {\color[HTML]{000000} ${34.16}_{0.00}$} & {\color[HTML]{000000} ${26.03}_{0.00}$} & {\color[HTML]{000000} ${54.25}_{0.00}$} & {\color[HTML]{000000} ${22.91}_{0.00}$} & {\color[HTML]{00ba55} ${46.68}_{0.00}$} & {\color[HTML]{000000} ${29.85}_{0.00}$} & {\color[HTML]{000000} $31.53$} \\
{\color[HTML]{000000} } & {\color[HTML]{000000} } & \multirow{-2}{*}{{\color[HTML]{000000} Val.}} & {\color[HTML]{000000} w} & {\color[HTML]{00ba55} ${5.25}_{0.57}$} & {\color[HTML]{000000} ${9.46}_{0.01}$} & {\color[HTML]{00ba55} ${27.57}_{5.24}$} & {\color[HTML]{000000} ${31.40}_{3.63}$} & {\color[HTML]{00ba55} ${50.31}_{0.61}$} & {\color[HTML]{00ba55} ${50.06}_{1.78}$} & {\color[HTML]{00ba55} ${51.44}_{2.19}$} & {\color[HTML]{00ba55} ${33.29}_{0.25}$} & {\color[HTML]{00ba55} ${61.13}_{2.01}$} & {\color[HTML]{00ba55} ${27.20}_{0.36}$} & {\color[HTML]{000000} ${38.58}_{0.10}$} & {\color[HTML]{00ba55} ${41.68}_{2.24}$} & {\color[HTML]{00ba55} $35.61$} \\ \cmidrule(l){3-17} 
{\color[HTML]{000000} } & {\color[HTML]{000000} } & {\color[HTML]{000000} } & {\color[HTML]{000000} w/o} & {\color[HTML]{00ba55} ${6.08}_{0.00}$} & {\color[HTML]{00ba55} ${6.50}_{0.00}$} & {\color[HTML]{000000} ${20.45}_{0.00}$} & {\color[HTML]{00ba55} ${34.84}_{0.00}$} & {\color[HTML]{000000} ${46.00}_{0.00}$} & {\color[HTML]{000000} ${38.83}_{0.00}$} & {\color[HTML]{000000} ${35.37}_{0.00}$} & {\color[HTML]{000000} ${28.04}_{0.00}$} & {\color[HTML]{00ba55} ${64.67}_{0.00}$} & {\color[HTML]{000000} ${22.73}_{0.00}$} & {\color[HTML]{00ba55} ${32.56}_{0.00}$} & {\color[HTML]{000000} ${29.22}_{0.00}$} & {\color[HTML]{000000} $30.44$} \\
{\color[HTML]{000000} } & \multirow{-4}{*}{{\color[HTML]{000000} \begin{tabular}[c]{@{}c@{}}MF1\\ (\%)\end{tabular}}} & \multirow{-2}{*}{{\color[HTML]{000000} Test}} & w/ & \cellcolor[HTML]{EFEFEF}{\color[HTML]{000000} ${5.68}_{1.01}$} & \cellcolor[HTML]{EFEFEF}{\color[HTML]{000000} ${5.78}_{0.11}$} & \cellcolor[HTML]{EFEFEF}{\color[HTML]{00ba55} ${27.89}_{5.96}$} & \cellcolor[HTML]{EFEFEF}{\color[HTML]{000000} ${27.69}_{3.78}$} & \cellcolor[HTML]{EFEFEF}{\color[HTML]{00ba55} ${46.66}_{0.47}$} & \cellcolor[HTML]{EFEFEF}{\color[HTML]{00ba55} ${49.68}_{0.71}$} & \cellcolor[HTML]{EFEFEF}{\color[HTML]{00ba55} ${48.89}_{3.86}$} & \cellcolor[HTML]{EFEFEF}{\color[HTML]{00ba55} ${29.66}_{0.32}$} & \cellcolor[HTML]{EFEFEF}{\color[HTML]{000000} ${58.07}_{4.31}$} & \cellcolor[HTML]{EFEFEF}{\color[HTML]{00ba55} ${29.69}_{0.45}$} & \cellcolor[HTML]{EFEFEF}{\color[HTML]{000000} ${32.17}_{0.22}$} & \cellcolor[HTML]{EFEFEF}{\color[HTML]{00ba55} ${54.16}_{2.63}$} & \cellcolor[HTML]{EFEFEF}{\color[HTML]{00ba55} $34.67$} \\ \cmidrule(l){2-17} 
{\color[HTML]{000000} } & {\color[HTML]{000000} } & {\color[HTML]{000000} } & {\color[HTML]{000000} w/o} & {\color[HTML]{00ba55} ${53.30}_{0.00}$} & {\color[HTML]{00ba55} ${46.81}_{0.00}$} & {\color[HTML]{000000} ${33.37}_{0.00}$} & {\color[HTML]{00ba55} ${13.58}_{0.00}$} & {\color[HTML]{000000} ${18.57}_{0.00}$} & {\color[HTML]{00ba55} ${22.00}_{0.00}$} & {\color[HTML]{000000} ${42.42}_{0.00}$} & {\color[HTML]{00ba55} ${13.50}_{0.00}$} & {\color[HTML]{000000} ${6.37}_{0.00}$} & {\color[HTML]{000000} ${20.26}_{0.00}$} & {\color[HTML]{00ba55} ${5.92}_{0.00}$} & {\color[HTML]{000000} ${49.98}_{0.00}$} & {\color[HTML]{000000} $27.17$} \\
{\color[HTML]{000000} } & {\color[HTML]{000000} } & \multirow{-2}{*}{{\color[HTML]{000000} Val.}} & {\color[HTML]{000000} w/} & {\color[HTML]{000000} ${58.20}_{2.84}$} & {\color[HTML]{000000} ${48.35}_{0.11}$} & {\color[HTML]{00ba55} ${17.65}_{7.90}$} & {\color[HTML]{000000} ${14.21}_{4.88}$} & {\color[HTML]{00ba55} ${17.31}_{0.55}$} & {\color[HTML]{000000} ${22.03}_{1.67}$} & {\color[HTML]{00ba55} ${17.86}_{2.16}$} & {\color[HTML]{000000} ${17.87}_{2.35}$} & {\color[HTML]{00ba55} ${5.43}_{1.65}$} & {\color[HTML]{00ba55} ${14.39}_{0.67}$} & {\color[HTML]{000000} ${6.90}_{0.26}$} & {\color[HTML]{00ba55} ${28.80}_{1.95}$} & {\color[HTML]{00ba55} $22.42$} \\ \cmidrule(l){3-17} 
{\color[HTML]{000000} } & {\color[HTML]{000000} } & {\color[HTML]{000000} } & {\color[HTML]{000000} w/o} & {\color[HTML]{00ba55} ${52.61}_{0.00}$} & {\color[HTML]{00ba55} ${51.74}_{0.00}$} & {\color[HTML]{000000} ${32.92}_{0.00}$} & {\color[HTML]{00ba55} ${14.46}_{0.00}$} & {\color[HTML]{00ba55} ${19.68}_{0.00}$} & {\color[HTML]{000000} ${27.22}_{0.00}$} & {\color[HTML]{000000} ${39.90}_{0.00}$} & {\color[HTML]{000000} ${10.44}_{0.00}$} & {\color[HTML]{000000} ${8.01}_{0.00}$} & {\color[HTML]{000000} ${22.48}_{0.00}$} & {\color[HTML]{00ba55} ${12.02}_{0.00}$} & {\color[HTML]{000000} ${49.70}_{0.00}$} & {\color[HTML]{000000} $28.43$} \\
\multirow{-13}{*}{{\color[HTML]{000000} GPT-2}} & \multirow{-4}{*}{{\color[HTML]{000000} \begin{tabular}[c]{@{}c@{}}$ECE_1$\\ (\%, $\downarrow$)\end{tabular}}} & \multirow{-2}{*}{{\color[HTML]{000000} Test}} & w/ & \cellcolor[HTML]{EFEFEF}{\color[HTML]{000000} ${57.49}_{2.32}$} & \cellcolor[HTML]{EFEFEF}{\color[HTML]{000000} ${54.24}_{0.30}$} & \cellcolor[HTML]{EFEFEF}{\color[HTML]{00ba55} ${17.41}_{10.77}$} & \cellcolor[HTML]{EFEFEF}{\color[HTML]{000000} ${20.08}_{5.50}$} & \cellcolor[HTML]{EFEFEF}{\color[HTML]{000000} ${20.17}_{0.46}$} & \cellcolor[HTML]{EFEFEF}{\color[HTML]{00ba55} ${25.34}_{0.77}$} & \cellcolor[HTML]{EFEFEF}{\color[HTML]{00ba55} ${21.52}_{3.83}$} & \cellcolor[HTML]{EFEFEF}{\color[HTML]{00ba55} ${8.81}_{0.48}$} & \cellcolor[HTML]{EFEFEF}{\color[HTML]{00ba55} ${3.94}_{2.29}$} & \cellcolor[HTML]{EFEFEF}{\color[HTML]{00ba55} ${11.62}_{0.67}$} & \cellcolor[HTML]{EFEFEF}{\color[HTML]{000000} ${12.48}_{0.25}$} & \cellcolor[HTML]{EFEFEF}{\color[HTML]{00ba55} ${14.84}_{2.55}$} & \cellcolor[HTML]{EFEFEF}{\color[HTML]{00ba55} $22.33$} \\ \bottomrule
\end{tabular}
}
\end{table*}

% k = 1 table
\begin{table*}[t]
\caption{Accuracy and Macro-F1 results ($\%$, $mean_{std}$, $k=1$). A better result is in {\color[HTML]{00ba55}green}. Notation is the same with the Tabel~\ref{tab:1}.}
\label{tab:K1}
\centering
\resizebox{2\columnwidth}{!}{
\begin{tabular}{@{}cccc|cccccccccccc|c@{}}
\toprule
\multicolumn{4}{c|}{{\color[HTML]{000000} Dataset}} & {\color[HTML]{000000} PS} & {\color[HTML]{000000} HS} & {\color[HTML]{000000} SE'14R} & {\color[HTML]{000000} SE'14L} & {\color[HTML]{000000} RTE} & {\color[HTML]{000000} MRPC} & {\color[HTML]{000000} Ethos} & {\color[HTML]{000000} FP} & {\color[HTML]{000000} SST2} & {\color[HTML]{000000} TEE} & {\color[HTML]{000000} TES} & {\color[HTML]{000000} TEH} & {\color[HTML]{000000} Mean} \\ \midrule
{\color[HTML]{000000} } & \multicolumn{3}{c|}{{\color[HTML]{000000} $\lambda$}} & {\color[HTML]{000000} 0.2} & {\color[HTML]{000000} 0.012} & {\color[HTML]{000000} 0.1} & {\color[HTML]{000000} 0.02} & {\color[HTML]{000000} 0.2} & {\color[HTML]{000000} 0.016} & {\color[HTML]{000000} 0.016} & {\color[HTML]{000000} 0.014} & {\color[HTML]{000000} 0.018} & {\color[HTML]{000000} 0.014} & {\color[HTML]{000000} 0.002} & {\color[HTML]{000000} 0.08} & {\color[HTML]{000000} ---} \\ \cmidrule(l){2-17} 
{\color[HTML]{000000} } & {\color[HTML]{000000} } & {\color[HTML]{000000} } & {\color[HTML]{000000} w/o} & {\color[HTML]{000000} ${39.43}_{0.99}$} & {\color[HTML]{000000} ${76.02}_{0.86}$} & {\color[HTML]{000000} ${57.85}_{1.45}$} & {\color[HTML]{000000} ${53.77}_{1.82}$} & {\color[HTML]{000000} ${50.41}_{1.15}$} & {\color[HTML]{00ba55} ${58.48}_{1.48}$} & {\color[HTML]{000000} ${50.32}_{1.81}$} & {\color[HTML]{000000} ${53.93}_{0.82}$} & {\color[HTML]{000000} ${68.20}_{0.52}$} & {\color[HTML]{000000} ${43.48}_{0.80}$} & {\color[HTML]{000000} ${43.83}_{2.45}$} & {\color[HTML]{000000} ${50.90}_{1.02}$} & {\color[HTML]{000000} $52.22$} \\
{\color[HTML]{000000} } & {\color[HTML]{000000} } & \multirow{-2}{*}{{\color[HTML]{000000} Val.}} & {\color[HTML]{000000} w/} & {\color[HTML]{00ba55} ${53.24}_{4.76}$} & {\color[HTML]{00ba55} ${76.86}_{0.54}$} & {\color[HTML]{00ba55} ${64.28}_{2.15}$} & {\color[HTML]{00ba55} ${55.57}_{1.05}$} & {\color[HTML]{00ba55} ${51.76}_{1.55}$} & {\color[HTML]{000000} ${58.28}_{1.35}$} & {\color[HTML]{00ba55} ${51.00}_{1.77}$} & {\color[HTML]{00ba55} ${53.96}_{1.27}$} & {\color[HTML]{00ba55} ${71.56}_{1.07}$} & {\color[HTML]{00ba55} ${44.32}_{0.50}$} & {\color[HTML]{00ba55} ${46.37}_{1.09}$} & {\color[HTML]{00ba55} ${52.13}_{0.96}$} & {\color[HTML]{00ba55} $56.61$} \\ \cmidrule(l){3-17} 
{\color[HTML]{000000} } & {\color[HTML]{000000} } & {\color[HTML]{000000} } & {\color[HTML]{000000} w/o} & {\color[HTML]{000000} ${35.08}_{1.74}$} & {\color[HTML]{000000} ${84.92}_{0.94}$} & {\color[HTML]{000000} ${53.66}_{1.12}$} & {\color[HTML]{000000} ${51.60}_{1.29}$} & {\color[HTML]{000000} ${50.52}_{1.31}$} & {\color[HTML]{000000} ${53.50}_{2.26}$} & {\color[HTML]{000000} ${50.40}_{1.37}$} & {\color[HTML]{00ba55} ${52.29}_{1.48}$} & {\color[HTML]{00ba55} ${73.89}_{0.84}$} & {\color[HTML]{000000} ${41.94}_{0.92}$} & {\color[HTML]{000000} ${44.80}_{1.26}$} & {\color[HTML]{000000} ${50.71}_{1.46}$} & {\color[HTML]{000000} $53.61$} \\
{\color[HTML]{000000} } & \multirow{-4}{*}{{\color[HTML]{000000} \begin{tabular}[c]{@{}c@{}}Acc.\\ (\%)\end{tabular}}} & \multirow{-2}{*}{{\color[HTML]{000000} Test}} & w/ & \cellcolor[HTML]{EFEFEF}{\color[HTML]{00ba55} ${46.31}_{3.18}$} & \cellcolor[HTML]{EFEFEF}{\color[HTML]{00ba55} ${85.64}_{1.06}$} & \cellcolor[HTML]{EFEFEF}{\color[HTML]{00ba55} ${58.26}_{1.49}$} & \cellcolor[HTML]{EFEFEF}{\color[HTML]{00ba55} ${53.71}_{1.08}$} & \cellcolor[HTML]{EFEFEF}{\color[HTML]{00ba55} ${51.84}_{0.51}$} & \cellcolor[HTML]{EFEFEF}{\color[HTML]{00ba55} ${54.28}_{1.10}$} & \cellcolor[HTML]{EFEFEF}{\color[HTML]{00ba55} ${50.88}_{2.16}$} & \cellcolor[HTML]{EFEFEF}{\color[HTML]{000000} ${51.96}_{1.10}$} & \cellcolor[HTML]{EFEFEF}{\color[HTML]{000000} ${73.85}_{1.67}$} & \cellcolor[HTML]{EFEFEF}{\color[HTML]{00ba55} ${42.06}_{0.93}$} & \cellcolor[HTML]{EFEFEF}{\color[HTML]{00ba55} ${45.23}_{1.75}$} & \cellcolor[HTML]{EFEFEF}{\color[HTML]{00ba55} ${52.11}_{1.38}$} & \cellcolor[HTML]{EFEFEF}{\color[HTML]{00ba55} $55.51$} \\ \cmidrule(l){2-17} 
{\color[HTML]{000000} } & {\color[HTML]{000000} } & {\color[HTML]{000000} } & {\color[HTML]{000000} w/o} & {\color[HTML]{00ba55} ${26.01}_{0.70}$} & {\color[HTML]{00ba55} ${24.73}_{0.66}$} & {\color[HTML]{000000} ${50.78}_{1.61}$} & {\color[HTML]{000000} ${52.26}_{1.85}$} & {\color[HTML]{00ba55} ${50.36}_{1.18}$} & {\color[HTML]{00ba55} ${51.11}_{1.43}$} & {\color[HTML]{000000} ${49.04}_{1.83}$} & {\color[HTML]{000000} ${38.25}_{1.01}$} & {\color[HTML]{000000} ${67.39}_{0.58}$} & {\color[HTML]{000000} ${26.71}_{1.08}$} & {\color[HTML]{000000} ${38.75}_{2.41}$} & {\color[HTML]{000000} ${49.42}_{0.85}$} & {\color[HTML]{000000} $43.73$} \\
{\color[HTML]{000000} } & {\color[HTML]{000000} } & \multirow{-2}{*}{{\color[HTML]{000000} Val.}} & {\color[HTML]{000000} w/} & {\color[HTML]{000000} ${23.18}_{1.34}$} & {\color[HTML]{000000} ${24.49}_{0.92}$} & {\color[HTML]{00ba55} ${52.49}_{2.89}$} & {\color[HTML]{00ba55} ${53.61}_{1.35}$} & {\color[HTML]{000000} ${48.28}_{4.78}$} & {\color[HTML]{000000} ${50.61}_{1.72}$} & {\color[HTML]{00ba55} ${49.88}_{1.81}$} & {\color[HTML]{00ba55} ${38.94}_{1.99}$} & {\color[HTML]{00ba55} ${71.02}_{1.16}$} & {\color[HTML]{00ba55} ${27.65}_{1.31}$} & {\color[HTML]{00ba55} ${41.43}_{1.61}$} & {\color[HTML]{00ba55} ${50.75}_{0.97}$} & {\color[HTML]{00ba55} $44.36$} \\ \cmidrule(l){3-17} 
{\color[HTML]{000000} } & {\color[HTML]{000000} } & {\color[HTML]{000000} } & {\color[HTML]{000000} w/o} & {\color[HTML]{00ba55} ${25.30}_{1.20}$} & {\color[HTML]{000000} ${24.73}_{0.71}$} & {\color[HTML]{00ba55} ${47.91}_{1.24}$} & {\color[HTML]{000000} ${50.62}_{1.25}$} & {\color[HTML]{00ba55} ${50.17}_{1.19}$} & {\color[HTML]{000000} ${49.97}_{2.33}$} & {\color[HTML]{000000} ${49.81}_{1.33}$} & {\color[HTML]{000000} ${37.90}_{1.98}$} & {\color[HTML]{00ba55} ${72.25}_{1.04}$} & {\color[HTML]{000000} ${24.67}_{1.43}$} & {\color[HTML]{000000} ${40.10}_{1.28}$} & {\color[HTML]{000000} ${48.45}_{1.61}$} & {\color[HTML]{000000} $43.49$} \\
{\color[HTML]{000000} } & \multirow{-4}{*}{{\color[HTML]{000000} \begin{tabular}[c]{@{}c@{}}MF1\\ (\%)\end{tabular}}} & \multirow{-2}{*}{{\color[HTML]{000000} Test}} & w/ & \cellcolor[HTML]{EFEFEF}{\color[HTML]{000000} ${24.76}_{1.72}$} & \cellcolor[HTML]{EFEFEF}{\color[HTML]{00ba55} ${25.13}_{1.02}$} & \cellcolor[HTML]{EFEFEF}{\color[HTML]{000000} ${47.53}_{2.14}$} & \cellcolor[HTML]{EFEFEF}{\color[HTML]{00ba55} ${52.18}_{0.89}$} & \cellcolor[HTML]{EFEFEF}{\color[HTML]{000000} ${47.10}_{4.38}$} & \cellcolor[HTML]{EFEFEF}{\color[HTML]{00ba55} ${50.65}_{1.07}$} & \cellcolor[HTML]{EFEFEF}{\color[HTML]{00ba55} ${50.36}_{2.14}$} & \cellcolor[HTML]{EFEFEF}{\color[HTML]{00ba55} ${38.93}_{1.27}$} & \cellcolor[HTML]{EFEFEF}{\color[HTML]{000000} ${72.22}_{1.96}$} & \cellcolor[HTML]{EFEFEF}{\color[HTML]{00ba55} ${24.97}_{0.92}$} & \cellcolor[HTML]{EFEFEF}{\color[HTML]{00ba55} ${40.45}_{1.73}$} & \cellcolor[HTML]{EFEFEF}{\color[HTML]{00ba55} ${49.94}_{1.40}$} & \cellcolor[HTML]{EFEFEF}{\color[HTML]{00ba55} $52.42$} \\ \cmidrule(l){2-17} 
{\color[HTML]{000000} } & {\color[HTML]{000000} } & {\color[HTML]{000000} } & {\color[HTML]{000000} w/o} & {\color[HTML]{000000} ${23.09}_{1.25}$} & {\color[HTML]{000000} ${22.67}_{0.82}$} & {\color[HTML]{00ba55} ${11.25}_{1.73}$} & {\color[HTML]{000000} ${17.01}_{1.66}$} & {\color[HTML]{000000} ${49.33}_{1.14}$} & {\color[HTML]{00ba55} ${35.31}_{1.39}$} & {\color[HTML]{000000} ${47.42}_{1.82}$} & {\color[HTML]{000000} ${27.54}_{0.97}$} & {\color[HTML]{000000} ${5.33}_{1.23}$} & {\color[HTML]{000000} ${44.33}_{1.12}$} & {\color[HTML]{000000} ${28.71}_{2.16}$} & {\color[HTML]{000000} ${45.64}_{1.19}$} & {\color[HTML]{000000} $29.80$} \\
{\color[HTML]{000000} } & {\color[HTML]{000000} } & \multirow{-2}{*}{{\color[HTML]{000000} Val.}} & {\color[HTML]{000000} w/} & {\color[HTML]{00ba55} ${16.64}_{4.79}$} & {\color[HTML]{00ba55} ${21.81}_{0.59}$} & {\color[HTML]{000000} ${13.33}_{2.62}$} & {\color[HTML]{00ba55} ${15.57}_{0.64}$} & {\color[HTML]{00ba55} ${26.82}_{5.71}$} & {\color[HTML]{000000} ${35.56}_{1.89}$} & {\color[HTML]{00ba55} ${46.63}_{1.82}$} & {\color[HTML]{00ba55} ${27.39}_{1.11}$} & {\color[HTML]{00ba55} ${3.72}_{1.19}$} & {\color[HTML]{00ba55} ${42.88}_{0.69}$} & {\color[HTML]{00ba55} ${26.36}_{1.40}$} & {\color[HTML]{00ba55} ${43.95}_{1.10}$} & {\color[HTML]{00ba55} $26.72$} \\ \cmidrule(l){3-17} 
{\color[HTML]{000000} } & {\color[HTML]{000000} } & {\color[HTML]{000000} } & {\color[HTML]{000000} w/o} & {\color[HTML]{000000} ${28.18}_{1.95}$} & {\color[HTML]{000000} ${13.97}_{0.95}$} & {\color[HTML]{00ba55} ${12.05}_{1.52}$} & {\color[HTML]{000000} ${15.95}_{1.05}$} & {\color[HTML]{000000} ${49.20}_{1.34}$} & {\color[HTML]{000000} ${39.94}_{2.33}$} & {\color[HTML]{000000} ${47.35}_{1.40}$} & {\color[HTML]{000000} ${28.24}_{1.42}$} & {\color[HTML]{00ba55} ${4.60}_{0.95}$} & {\color[HTML]{000000} ${45.34}_{0.90}$} & {\color[HTML]{000000} ${27.63}_{0.85}$} & {\color[HTML]{000000} ${46.98}_{1.37}$} & {\color[HTML]{000000} $29.95$} \\
\multirow{-13}{*}{{\color[HTML]{000000} GPT-J}} & \multirow{-4}{*}{{\color[HTML]{000000} \begin{tabular}[c]{@{}c@{}}$ECE_1$\\ (\%, $\downarrow$)\end{tabular}}} & \multirow{-2}{*}{{\color[HTML]{000000} Test}} & w/ & \cellcolor[HTML]{EFEFEF}{\color[HTML]{00ba55} ${20.85}_{5.28}$} & \cellcolor[HTML]{EFEFEF}{\color[HTML]{00ba55} ${13.21}_{1.07}$} & \cellcolor[HTML]{EFEFEF}{\color[HTML]{000000} ${15.45}_{1.84}$} & \cellcolor[HTML]{EFEFEF}{\color[HTML]{00ba55} ${14.28}_{0.93}$} & \cellcolor[HTML]{EFEFEF}{\color[HTML]{00ba55} ${25.64}_{3.69}$} & \cellcolor[HTML]{EFEFEF}{\color[HTML]{00ba55} ${35.56}_{1.89}$} & \cellcolor[HTML]{EFEFEF}{\color[HTML]{00ba55} ${46.76}_{2.10}$} & \cellcolor[HTML]{EFEFEF}{\color[HTML]{00ba55} ${28.13}_{0.97}$} & \cellcolor[HTML]{EFEFEF}{\color[HTML]{000000} ${5.53}_{1.44}$} & \cellcolor[HTML]{EFEFEF}{\color[HTML]{00ba55} ${44.71}_{0.93}$} & \cellcolor[HTML]{EFEFEF}{\color[HTML]{00ba55} ${27.08}_{1.93}$} & \cellcolor[HTML]{EFEFEF}{\color[HTML]{00ba55} ${45.28}_{1.33}$} & \cellcolor[HTML]{EFEFEF}{\color[HTML]{00ba55} $26.87$} \\ \midrule
{\color[HTML]{000000} } & \multicolumn{3}{c|}{{\color[HTML]{000000} $\lambda$}} & {\color[HTML]{000000} 0.02} & {\color[HTML]{000000} 0.002} & {\color[HTML]{000000} 0.016} & {\color[HTML]{000000} 0.004} & {\color[HTML]{000000} 0.002} & {\color[HTML]{000000} 0.002} & {\color[HTML]{000000} 0.2} & {\color[HTML]{000000} 0.06} & {\color[HTML]{000000} 0.02} & {\color[HTML]{000000} 0.02} & {\color[HTML]{000000} 0.06} & {\color[HTML]{000000} 0.1} & {\color[HTML]{000000} ---} \\ \cmidrule(l){2-17} 
{\color[HTML]{000000} } & {\color[HTML]{000000} } & {\color[HTML]{000000} } & {\color[HTML]{000000} w/o} & {\color[HTML]{000000} ${42.42}_{0.97}$} & {\color[HTML]{00ba55} ${41.37}_{1.11}$} & {\color[HTML]{000000} ${45.68}_{1.09}$} & {\color[HTML]{000000} ${39.69}_{1.69}$} & {\color[HTML]{00ba55} ${50.62}_{1.35}$} & {\color[HTML]{000000} ${68.71}_{0.20}$} & {\color[HTML]{000000} ${47.26}_{0.72}$} & {\color[HTML]{000000} ${45.45}_{1.81}$} & {\color[HTML]{000000} ${57.79}_{1.23}$} & {\color[HTML]{000000} ${28.11}_{0.82}$} & {\color[HTML]{000000} ${41.45}_{1.36}$} & {\color[HTML]{000000} ${42.70}_{0.32}$} & {\color[HTML]{000000} $45.94$} \\
{\color[HTML]{000000} } & {\color[HTML]{000000} } & \multirow{-2}{*}{{\color[HTML]{000000} Val.}} & {\color[HTML]{000000} w/} & {\color[HTML]{00ba55} ${45.55}_{3.18}$} & {\color[HTML]{000000} ${37.34}_{1.33}$} & {\color[HTML]{00ba55} ${48.75}_{1.80}$} & {\color[HTML]{00ba55} ${41.66}_{1.04}$} & {\color[HTML]{000000} ${50.18}_{2.11}$} & {\color[HTML]{00ba55} ${68.81}_{0.18}$} & {\color[HTML]{00ba55} ${53.35}_{2.99}$} & {\color[HTML]{00ba55} ${55.33}_{2.72}$} & {\color[HTML]{00ba55} ${59.26}_{1.38}$} & {\color[HTML]{00ba55} ${32.73}_{0.53}$} & {\color[HTML]{00ba55} ${45.25}_{1.34}$} & {\color[HTML]{00ba55} ${48.41}_{2.15}$} & {\color[HTML]{00ba55} $48.88$} \\ \cmidrule(l){3-17} 
{\color[HTML]{000000} } & {\color[HTML]{000000} } & {\color[HTML]{000000} } & {\color[HTML]{000000} w/o} & {\color[HTML]{000000} ${35.07}_{1.44}$} & {\color[HTML]{00ba55} ${39.64}_{0.61}$} & {\color[HTML]{000000} ${43.24}_{1.43}$} & {\color[HTML]{000000} ${38.47}_{0.97}$} & {\color[HTML]{000000} ${50.22}_{1.27}$} & {\color[HTML]{00ba55} ${62.94}_{0.42}$} & {\color[HTML]{000000} ${48.65}_{0.59}$} & {\color[HTML]{000000} ${45.09}_{1.51}$} & {\color[HTML]{000000} ${59.76}_{1.67}$} & {\color[HTML]{000000} ${27.61}_{1.20}$} & {\color[HTML]{000000} ${37.98}_{1.24}$} & {\color[HTML]{000000} ${44.37}_{0.81}$} & {\color[HTML]{000000} $44.42$} \\
{\color[HTML]{000000} } & \multirow{-4}{*}{{\color[HTML]{000000} \begin{tabular}[c]{@{}c@{}}Acc.\\ (\%)\end{tabular}}} & \multirow{-2}{*}{{\color[HTML]{000000} Test}} & w/ & \cellcolor[HTML]{EFEFEF}{\color[HTML]{00ba55} ${40.08}_{2.91}$} & \cellcolor[HTML]{EFEFEF}{\color[HTML]{000000} ${34.92}_{1.12}$} & \cellcolor[HTML]{EFEFEF}{\color[HTML]{00ba55} ${44.36}_{0.95}$} & \cellcolor[HTML]{EFEFEF}{\color[HTML]{00ba55} ${39.41}_{1.30}$} & \cellcolor[HTML]{EFEFEF}{\color[HTML]{00ba55} ${50.58}_{2.31}$} & \cellcolor[HTML]{EFEFEF}{\color[HTML]{000000} ${62.47}_{0.29}$} & \cellcolor[HTML]{EFEFEF}{\color[HTML]{00ba55} ${51.69}_{2.95}$} & \cellcolor[HTML]{EFEFEF}{\color[HTML]{00ba55} ${52.12}_{1.60}$} & \cellcolor[HTML]{EFEFEF}{\color[HTML]{00ba55} ${60.64}_{1.78}$} & \cellcolor[HTML]{EFEFEF}{\color[HTML]{00ba55} ${30.55}_{0.79}$} & \cellcolor[HTML]{EFEFEF}{\color[HTML]{00ba55} ${43.88}_{1.53}$} & \cellcolor[HTML]{EFEFEF}{\color[HTML]{00ba55} ${55.35}_{0.79}$} & \cellcolor[HTML]{EFEFEF}{\color[HTML]{00ba55} $47.17$} \\ \cmidrule(l){2-17} 
{\color[HTML]{000000} } & {\color[HTML]{000000} } & {\color[HTML]{000000} } & {\color[HTML]{000000} w/o} & {\color[HTML]{00ba55} ${26.44}_{0.73}$} & {\color[HTML]{00ba55} ${20.32}_{1.44}$} & {\color[HTML]{000000} ${32.96}_{0.78}$} & {\color[HTML]{000000} ${33.69}_{1.92}$} & {\color[HTML]{00ba55} ${50.59}_{1.35}$} & {\color[HTML]{000000} ${41.67}_{0.39}$} & {\color[HTML]{000000} ${43.18}_{0.82}$} & {\color[HTML]{000000} ${30.19}_{1.51}$} & {\color[HTML]{000000} ${56.71}_{1.25}$} & {\color[HTML]{000000} ${22.59}_{0.81}$} & {\color[HTML]{00ba55} ${36.97}_{1.57}$} & {\color[HTML]{000000} ${32.23}_{0.53}$} & {\color[HTML]{000000} $35.63$} \\
{\color[HTML]{000000} } & {\color[HTML]{000000} } & \multirow{-2}{*}{{\color[HTML]{000000} Val.}} & {\color[HTML]{000000} w} & {\color[HTML]{000000} ${26.37}_{1.41}$} & {\color[HTML]{000000} ${18.95}_{1.29}$} & {\color[HTML]{00ba55} ${35.97}_{1.24}$} & {\color[HTML]{00ba55} ${35.61}_{1.25}$} & {\color[HTML]{000000} ${50.13}_{2.10}$} & {\color[HTML]{00ba55} ${42.28}_{0.48}$} & {\color[HTML]{00ba55} ${47.26}_{2.81}$} & {\color[HTML]{00ba55} ${34.47}_{0.77}$} & {\color[HTML]{00ba55} ${59.22}_{1.39}$} & {\color[HTML]{00ba55} ${22.87}_{1.07}$} & {\color[HTML]{000000} ${30.49}_{1.97}$} & {\color[HTML]{00ba55} ${51.68}_{2.03}$} & {\color[HTML]{00ba55} $37.94$} \\ \cmidrule(l){3-17} 
{\color[HTML]{000000} } & {\color[HTML]{000000} } & {\color[HTML]{000000} } & {\color[HTML]{000000} w/o} & {\color[HTML]{000000} ${23.97}_{1.23}$} & {\color[HTML]{00ba55} ${17.45}_{0.85}$} & {\color[HTML]{000000} ${33.27}_{1.93}$} & {\color[HTML]{000000} ${36.32}_{1.03}$} & {\color[HTML]{000000} ${49.85}_{1.26}$} & {\color[HTML]{00ba55} ${39.41}_{0.66}$} & {\color[HTML]{000000} ${43.80}_{0.77}$} & {\color[HTML]{000000} ${30.92}_{1.38}$} & {\color[HTML]{00ba55} ${59.63}_{1.61}$} & {\color[HTML]{00ba55} ${22.32}_{1.38}$} & {\color[HTML]{00ba55} ${36.21}_{1.30}$} & {\color[HTML]{000000} ${38.79}_{1.06}$} & {\color[HTML]{000000} $36.00$} \\
{\color[HTML]{000000} } & \multirow{-4}{*}{{\color[HTML]{000000} \begin{tabular}[c]{@{}c@{}}MF1\\ (\%)\end{tabular}}} & \multirow{-2}{*}{{\color[HTML]{000000} Test}} & w/ & \cellcolor[HTML]{EFEFEF}{\color[HTML]{00ba55} ${26.31}_{1.58}$} & \cellcolor[HTML]{EFEFEF}{\color[HTML]{000000} ${15.87}_{1.14}$} & \cellcolor[HTML]{EFEFEF}{\color[HTML]{00ba55} ${33.78}_{1.33}$} & \cellcolor[HTML]{EFEFEF}{\color[HTML]{00ba55} ${37.11}_{1.53}$} & \cellcolor[HTML]{EFEFEF}{\color[HTML]{00ba55} ${50.18}_{2.40}$} & \cellcolor[HTML]{EFEFEF}{\color[HTML]{000000} ${39.10}_{0.41}$} & \cellcolor[HTML]{EFEFEF}{\color[HTML]{00ba55} ${44.93}_{3.66}$} & \cellcolor[HTML]{EFEFEF}{\color[HTML]{00ba55} ${34.58}_{1.06}$} & \cellcolor[HTML]{EFEFEF}{\color[HTML]{000000} ${59.52}_{1.82}$} & \cellcolor[HTML]{EFEFEF}{\color[HTML]{000000} ${21.69}_{1.08}$} & \cellcolor[HTML]{EFEFEF}{\color[HTML]{000000} ${33.40}_{1.36}$} & \cellcolor[HTML]{EFEFEF}{\color[HTML]{00ba55} ${50.57}_{1.50}$} & \cellcolor[HTML]{EFEFEF}{\color[HTML]{00ba55} $37.25$} \\ \cmidrule(l){2-17} 
{\color[HTML]{000000} } & {\color[HTML]{000000} } & {\color[HTML]{000000} } & {\color[HTML]{000000} w/o} & {\color[HTML]{000000} ${14.45}_{1.49}$} & {\color[HTML]{00ba55} ${25.96}_{1.14}$} & {\color[HTML]{00ba55} ${16.68}_{1.33}$} & {\color[HTML]{000000} ${25.55}_{1.99}$} & {\color[HTML]{00ba55} ${42.92}_{1.41}$} & {\color[HTML]{000000} ${19.22}_{0.49}$} & {\color[HTML]{000000} ${35.57}_{1.01}$} & {\color[HTML]{000000} ${21.11}_{1.76}$} & {\color[HTML]{000000} ${5.40}_{1.16}$} & {\color[HTML]{000000} ${31.62}_{1.17}$} & {\color[HTML]{00ba55} ${20.72}_{1.70}$} & {\color[HTML]{000000} ${46.15}_{0.38}$} & {\color[HTML]{000000} $25.44$} \\
{\color[HTML]{000000} } & {\color[HTML]{000000} } & \multirow{-2}{*}{{\color[HTML]{000000} Val.}} & {\color[HTML]{000000} w/} & {\color[HTML]{00ba55} ${9.64}_{1.86}$} & {\color[HTML]{000000} ${30.37}_{1.36}$} & {\color[HTML]{000000} ${17.27}_{2.01}$} & {\color[HTML]{00ba55} ${23.65}_{0.78}$} & {\color[HTML]{000000} ${43.31}_{2.03}$} & {\color[HTML]{00ba55} ${17.87}_{0.49}$} & {\color[HTML]{00ba55} ${19.53}_{3.89}$} & {\color[HTML]{00ba55} ${14.80}_{0.62}$} & {\color[HTML]{00ba55} ${3.94}_{0.41}$} & {\color[HTML]{00ba55} ${30.66}_{0.71}$} & {\color[HTML]{000000} ${24.11}_{2.54}$} & {\color[HTML]{00ba55} ${26.44}_{2.47}$} & {\color[HTML]{00ba55} $21.80$} \\ \cmidrule(l){3-17} 
{\color[HTML]{000000} } & {\color[HTML]{000000} } & {\color[HTML]{000000} } & {\color[HTML]{000000} w/o} & {\color[HTML]{000000} ${18.82}_{1.59}$} & {\color[HTML]{00ba55} ${26.53}_{0.96}$} & {\color[HTML]{00ba55} ${18.64}_{1.41}$} & {\color[HTML]{000000} ${24.56}_{0.96}$} & {\color[HTML]{000000} ${43.56}_{1.28}$} & {\color[HTML]{000000} ${24.09}_{0.58}$} & {\color[HTML]{000000} ${35.15}_{0.76}$} & {\color[HTML]{000000} ${22.13}_{1.42}$} & {\color[HTML]{00ba55} ${2.83}_{1.04}$} & {\color[HTML]{00ba55} ${32.31}_{1.34}$} & {\color[HTML]{000000} ${22.45}_{1.38}$} & {\color[HTML]{000000} ${39.96}_{0.77}$} & {\color[HTML]{000000} $25.92$} \\
\multirow{-13}{*}{{\color[HTML]{000000} GPT-2}} & \multirow{-4}{*}{{\color[HTML]{000000} \begin{tabular}[c]{@{}c@{}}$ECE_1$\\ (\%, $\downarrow$)\end{tabular}}} & \multirow{-2}{*}{{\color[HTML]{000000} Test}} & w/ & \cellcolor[HTML]{EFEFEF}{\color[HTML]{00ba55} ${13.20}_{2.03}$} & \cellcolor[HTML]{EFEFEF}{\color[HTML]{000000} ${32.02}_{1.47}$} & \cellcolor[HTML]{EFEFEF}{\color[HTML]{000000} ${20.45}_{1.07}$} & \cellcolor[HTML]{EFEFEF}{\color[HTML]{00ba55} ${23.83}_{1.22}$} & \cellcolor[HTML]{EFEFEF}{\color[HTML]{00ba55} ${43.07}_{2.30}$} & \cellcolor[HTML]{EFEFEF}{\color[HTML]{00ba55} ${23.01}_{0.87}$} & \cellcolor[HTML]{EFEFEF}{\color[HTML]{00ba55} ${23.65}_{6.52}$} & \cellcolor[HTML]{EFEFEF}{\color[HTML]{00ba55} ${14.48}_{1.15}$} & \cellcolor[HTML]{EFEFEF}{\color[HTML]{000000} ${3.22}_{1.50}$} & \cellcolor[HTML]{EFEFEF}{\color[HTML]{000000} ${33.43}_{1.60}$} & \cellcolor[HTML]{EFEFEF}{\color[HTML]{00ba55} ${19.37}_{1.64}$} & \cellcolor[HTML]{EFEFEF}{\color[HTML]{00ba55} ${23.89}_{1.50}$} & \cellcolor[HTML]{EFEFEF}{\color[HTML]{00ba55} $22.80$} \\ \bottomrule
\end{tabular}
}
\end{table*}

% k = 2 table
\begin{table*}[t]
\caption{Accuracy and Macro-F1 results ($\%$, $mean_{std}$, $k=2$). A better result is in {\color[HTML]{00ba55}green}. Notation is the same with the Tabel~\ref{tab:1}.}
\label{tab:K2}
\centering
\resizebox{2\columnwidth}{!}{
\begin{tabular}{@{}cccc|cccccccccccc|c@{}}
\toprule
\multicolumn{4}{c|}{{\color[HTML]{000000} Dataset}} & {\color[HTML]{000000} PS} & {\color[HTML]{000000} HS} & {\color[HTML]{000000} SE'14R} & {\color[HTML]{000000} SE'14L} & {\color[HTML]{000000} RTE} & {\color[HTML]{000000} MRPC} & {\color[HTML]{000000} Ethos} & {\color[HTML]{000000} FP} & {\color[HTML]{000000} SST2} & {\color[HTML]{000000} TEE} & {\color[HTML]{000000} TES} & {\color[HTML]{000000} TEH} & {\color[HTML]{000000} Mean} \\ \midrule
{\color[HTML]{000000} } & \multicolumn{3}{c|}{{\color[HTML]{000000} $\lambda$}} & {\color[HTML]{000000} 0.2} & {\color[HTML]{000000} 0.2} & {\color[HTML]{000000} 0.1} & {\color[HTML]{000000} 0.1} & {\color[HTML]{000000} 0.08} & {\color[HTML]{000000} 0.2} & {\color[HTML]{000000} 0.014} & {\color[HTML]{000000} 0.08} & {\color[HTML]{000000} 0.016} & {\color[HTML]{000000} 0.018} & {\color[HTML]{000000} 0.02} & {\color[HTML]{000000} 0.008} & {\color[HTML]{000000} ---} \\ \cmidrule(l){2-17} 
{\color[HTML]{000000} } & {\color[HTML]{000000} } & {\color[HTML]{000000} } & {\color[HTML]{000000} w/o} & {\color[HTML]{000000} ${42.54}_{1.49}$} & {\color[HTML]{000000} ${72.03}_{0.66}$} & {\color[HTML]{000000} ${41.50}_{0.97}$} & {\color[HTML]{000000} ${40.88}_{0.49}$} & {\color[HTML]{00ba55} ${51.31}_{1.19}$} & {\color[HTML]{000000} ${55.74}_{1.24}$} & {\color[HTML]{000000} ${49.81}_{2.28}$} & {\color[HTML]{00ba55} ${61.84}_{0.60}$} & {\color[HTML]{000000} ${72.05}_{0.96}$} & {\color[HTML]{000000} ${44.47}_{0.62}$} & {\color[HTML]{000000} ${48.85}_{0.75}$} & {\color[HTML]{000000} ${49.08}_{1.16}$} & {\color[HTML]{000000} $52.51$} \\
{\color[HTML]{000000} } & {\color[HTML]{000000} } & \multirow{-2}{*}{{\color[HTML]{000000} Val.}} & {\color[HTML]{000000} w/} & {\color[HTML]{00ba55} ${57.11}_{4.74}$} & {\color[HTML]{00ba55} ${77.34}_{3.29}$} & {\color[HTML]{00ba55} ${58.57}_{5.39}$} & {\color[HTML]{00ba55} ${51.25}_{2.71}$} & {\color[HTML]{000000} ${50.76}_{2.09}$} & {\color[HTML]{00ba55} ${58.98}_{6.13}$} & {\color[HTML]{00ba55} ${51.37}_{1.15}$} & {\color[HTML]{000000} ${60.47}_{1.77}$} & {\color[HTML]{00ba55} ${73.38}_{0.70}$} & {\color[HTML]{00ba55} ${44.67}_{0.63}$} & {\color[HTML]{00ba55} ${49.14}_{0.78}$} & {\color[HTML]{00ba55} ${50.62}_{2.11}$} & {\color[HTML]{00ba55} $56.97$} \\ \cmidrule(l){3-17} 
{\color[HTML]{000000} } & {\color[HTML]{000000} } & {\color[HTML]{000000} } & {\color[HTML]{000000} w/o} & {\color[HTML]{000000} ${39.23}_{0.98}$} & {\color[HTML]{000000} ${82.42}_{1.21}$} & {\color[HTML]{000000} ${36.69}_{1.30}$} & {\color[HTML]{000000} ${39.26}_{1.67}$} & {\color[HTML]{000000} ${48.13}_{2.60}$} & {\color[HTML]{000000} ${50.97}_{1.43}$} & {\color[HTML]{000000} ${51.41}_{1.93}$} & {\color[HTML]{00ba55} ${60.77}_{0.64}$} & {\color[HTML]{000000} ${72.60}_{1.66}$} & {\color[HTML]{00ba55} ${44.28}_{0.73}$} & {\color[HTML]{00ba55} ${48.79}_{1.00}$} & {\color[HTML]{00ba55} ${51.68}_{1.95}$} & {\color[HTML]{000000} $52.19$} \\
{\color[HTML]{000000} } & \multirow{-4}{*}{{\color[HTML]{000000} \begin{tabular}[c]{@{}c@{}}Acc.\\ (\%)\end{tabular}}} & \multirow{-2}{*}{{\color[HTML]{000000} Test}} & w/ & \cellcolor[HTML]{EFEFEF}{\color[HTML]{00ba55} ${50.95}_{4.52}$} & \cellcolor[HTML]{EFEFEF}{\color[HTML]{00ba55} ${82.99}_{5.51}$} & \cellcolor[HTML]{EFEFEF}{\color[HTML]{00ba55} ${53.91}_{3.42}$} & \cellcolor[HTML]{EFEFEF}{\color[HTML]{00ba55} ${49.30}_{3.09}$} & \cellcolor[HTML]{EFEFEF}{\color[HTML]{00ba55} ${49.45}_{1.30}$} & \cellcolor[HTML]{EFEFEF}{\color[HTML]{00ba55} ${53.43}_{5.23}$} & \cellcolor[HTML]{EFEFEF}{\color[HTML]{00ba55} ${51.46}_{1.62}$} & \cellcolor[HTML]{EFEFEF}{\color[HTML]{000000} ${59.51}_{1.22}$} & \cellcolor[HTML]{EFEFEF}{\color[HTML]{00ba55} ${72.83}_{1.36}$} & \cellcolor[HTML]{EFEFEF}{\color[HTML]{000000} ${44.21}_{0.53}$} & \cellcolor[HTML]{EFEFEF}{\color[HTML]{000000} ${48.77}_{1.20}$} & \cellcolor[HTML]{EFEFEF}{\color[HTML]{000000} ${51.02}_{1.84}$} & \cellcolor[HTML]{EFEFEF}{\color[HTML]{00ba55} $55.65$} \\ \cmidrule(l){2-17} 
{\color[HTML]{000000} } & {\color[HTML]{000000} } & {\color[HTML]{000000} } & {\color[HTML]{000000} w/o} & {\color[HTML]{00ba55} ${25.17}_{1.10}$} & {\color[HTML]{00ba55} ${25.03}_{0.33}$} & {\color[HTML]{000000} ${37.42}_{1.13}$} & {\color[HTML]{000000} ${39.11}_{0.83}$} & {\color[HTML]{000000} ${50.44}_{1.23}$} & {\color[HTML]{00ba55} ${51.35}_{1.45}$} & {\color[HTML]{000000} ${49.78}_{2.28}$} & {\color[HTML]{000000} ${31.83}_{1.31}$} & {\color[HTML]{000000} ${71.32}_{1.07}$} & {\color[HTML]{00ba55} ${25.43}_{1.03}$} & {\color[HTML]{000000} ${36.69}_{1.16}$} & {\color[HTML]{000000} ${49.05}_{1.17}$} & {\color[HTML]{000000} $41.05$} \\
{\color[HTML]{000000} } & {\color[HTML]{000000} } & \multirow{-2}{*}{{\color[HTML]{000000} Val.}} & {\color[HTML]{000000} w/} & {\color[HTML]{000000} ${22.75}_{1.84}$} & {\color[HTML]{000000} ${24.15}_{0.56}$} & {\color[HTML]{00ba55} ${50.06}_{5.11}$} & {\color[HTML]{00ba55} ${49.26}_{2.50}$} & {\color[HTML]{00ba55} ${50.47}_{2.22}$} & {\color[HTML]{000000} ${48.95}_{0.99}$} & {\color[HTML]{00ba55} ${51.36}_{1.15}$} & {\color[HTML]{00ba55} ${35.39}_{3.22}$} & {\color[HTML]{00ba55} ${72.77}_{0.78}$} & {\color[HTML]{000000} ${25.16}_{0.84}$} & {\color[HTML]{00ba55} ${37.72}_{1.56}$} & {\color[HTML]{00ba55} ${50.56}_{2.12}$} & {\color[HTML]{00ba55} $43.22$} \\ \cmidrule(l){3-17} 
{\color[HTML]{000000} } & {\color[HTML]{000000} } & {\color[HTML]{000000} } & {\color[HTML]{000000} w/o} & {\color[HTML]{00ba55} ${25.51}_{1.25}$} & {\color[HTML]{000000} ${24.95}_{0.85}$} & {\color[HTML]{000000} ${34.99}_{1.38}$} & {\color[HTML]{000000} ${39.75}_{1.63}$} & {\color[HTML]{000000} ${47.82}_{2.64}$} & {\color[HTML]{00ba55} ${49.56}_{1.48}$} & {\color[HTML]{000000} ${51.18}_{1.96}$} & {\color[HTML]{000000} ${30.66}_{0.98}$} & {\color[HTML]{000000} ${70.34}_{1.87}$} & {\color[HTML]{000000} ${24.05}_{0.97}$} & {\color[HTML]{00ba55} ${36.78}_{1.28}$} & {\color[HTML]{00ba55} ${51.49}_{1.96}$} & {\color[HTML]{000000} $40.59$} \\
{\color[HTML]{000000} } & \multirow{-4}{*}{{\color[HTML]{000000} \begin{tabular}[c]{@{}c@{}}MF1\\ (\%)\end{tabular}}} & \multirow{-2}{*}{{\color[HTML]{000000} Test}} & w/ & \cellcolor[HTML]{EFEFEF}{\color[HTML]{000000} ${22.98}_{1.69}$} & \cellcolor[HTML]{EFEFEF}{\color[HTML]{00ba55} ${25.10}_{0.79}$} & \cellcolor[HTML]{EFEFEF}{\color[HTML]{00ba55} ${47.51}_{2.96}$} & \cellcolor[HTML]{EFEFEF}{\color[HTML]{00ba55} ${48.50}_{3.12}$} & \cellcolor[HTML]{EFEFEF}{\color[HTML]{00ba55} ${49.35}_{1.25}$} & \cellcolor[HTML]{EFEFEF}{\color[HTML]{000000} ${46.51}_{1.56}$} & \cellcolor[HTML]{EFEFEF}{\color[HTML]{00ba55} ${51.19}_{1.62}$} & \cellcolor[HTML]{EFEFEF}{\color[HTML]{00ba55} ${32.85}_{2.38}$} & \cellcolor[HTML]{EFEFEF}{\color[HTML]{00ba55} ${70.54}_{1.76}$} & \cellcolor[HTML]{EFEFEF}{\color[HTML]{00ba55} ${24.05}_{0.67}$} & \cellcolor[HTML]{EFEFEF}{\color[HTML]{000000} ${36.63}_{0.53}$} & \cellcolor[HTML]{EFEFEF}{\color[HTML]{000000} ${50.87}_{1.80}$} & \cellcolor[HTML]{EFEFEF}{\color[HTML]{00ba55} $42.17$} \\ \cmidrule(l){2-17} 
{\color[HTML]{000000} } & {\color[HTML]{000000} } & {\color[HTML]{000000} } & {\color[HTML]{000000} w/o} & {\color[HTML]{000000} ${15.11}_{1.29}$} & {\color[HTML]{000000} ${21.52}_{0.58}$} & {\color[HTML]{000000} ${23.45}_{1.46}$} & {\color[HTML]{000000} ${27.07}_{0.89}$} & {\color[HTML]{00ba55} ${39.26}_{0.83}$} & {\color[HTML]{000000} ${23.82}_{1.59}$} & {\color[HTML]{000000} ${38.51}_{2.24}$} & {\color[HTML]{000000} ${22.54}_{0.77}$} & {\color[HTML]{00ba55} ${3.56}_{0.89}$} & {\color[HTML]{000000} ${44.33}_{0.66}$} & {\color[HTML]{000000} ${19.22}_{0.62}$} & {\color[HTML]{000000} ${40.33}_{1.30}$} & {\color[HTML]{000000} $26.56$} \\
{\color[HTML]{000000} } & {\color[HTML]{000000} } & \multirow{-2}{*}{{\color[HTML]{000000} Val.}} & {\color[HTML]{000000} w/} & {\color[HTML]{00ba55} ${13.69}_{3.31}$} & {\color[HTML]{00ba55} ${11.25}_{5.10}$} & {\color[HTML]{00ba55} ${8.03}_{4.49}$} & {\color[HTML]{00ba55} ${15.54}_{3.79}$} & {\color[HTML]{000000} ${39.53}_{1.65}$} & {\color[HTML]{00ba55} ${12.12}_{2.82}$} & {\color[HTML]{00ba55} ${37.12}_{0.98}$} & {\color[HTML]{00ba55} ${21.10}_{1.70}$} & {\color[HTML]{000000} ${4.80}_{0.74}$} & {\color[HTML]{00ba55} ${43.87}_{0.96}$} & {\color[HTML]{00ba55} ${18.64}_{1.08}$} & {\color[HTML]{00ba55} ${38.66}_{2.22}$} & {\color[HTML]{00ba55} $22.03$} \\ \cmidrule(l){3-17} 
{\color[HTML]{000000} } & {\color[HTML]{000000} } & {\color[HTML]{000000} } & {\color[HTML]{000000} w/o} & {\color[HTML]{000000} ${18.59}_{1.01}$} & {\color[HTML]{000000} ${13.15}_{1.15}$} & {\color[HTML]{000000} ${27.87}_{1.35}$} & {\color[HTML]{000000} ${26.71}_{1.83}$} & {\color[HTML]{000000} ${41.77}_{2.91}$} & {\color[HTML]{000000} ${28.29}_{1.47}$} & {\color[HTML]{000000} ${36.62}_{1.76}$} & {\color[HTML]{000000} ${22.97}_{0.87}$} & {\color[HTML]{00ba55} ${4.77}_{0.84}$} & {\color[HTML]{000000} ${44.53}_{0.98}$} & {\color[HTML]{00ba55} ${19.92}_{1.25}$} & {\color[HTML]{00ba55} ${36.63}_{1.86}$} & {\color[HTML]{000000} $26.82$} \\
\multirow{-13}{*}{{\color[HTML]{000000} GPT-J}} & \multirow{-4}{*}{{\color[HTML]{000000} \begin{tabular}[c]{@{}c@{}}$ECE_1$\\ (\%, $\downarrow$)\end{tabular}}} & \multirow{-2}{*}{{\color[HTML]{000000} Test}} & w/ & \cellcolor[HTML]{EFEFEF}{\color[HTML]{00ba55} ${13.49}_{2.88}$} & \cellcolor[HTML]{EFEFEF}{\color[HTML]{00ba55} ${6.52}_{2.25}$} & \cellcolor[HTML]{EFEFEF}{\color[HTML]{00ba55} ${11.44}_{3.91}$} & \cellcolor[HTML]{EFEFEF}{\color[HTML]{00ba55} ${15.80}_{2.76}$} & \cellcolor[HTML]{EFEFEF}{\color[HTML]{00ba55} ${41.33}_{2.22}$} & \cellcolor[HTML]{EFEFEF}{\color[HTML]{00ba55} ${20.24}_{6.56}$} & \cellcolor[HTML]{EFEFEF}{\color[HTML]{00ba55} ${36.61}_{1.67}$} & \cellcolor[HTML]{EFEFEF}{\color[HTML]{00ba55} ${21.88}_{1.75}$} & \cellcolor[HTML]{EFEFEF}{\color[HTML]{000000} ${5.32}_{1.16}$} & \cellcolor[HTML]{EFEFEF}{\color[HTML]{00ba55} ${44.40}_{0.77}$} & \cellcolor[HTML]{EFEFEF}{\color[HTML]{000000} ${19.98}_{0.70}$} & \cellcolor[HTML]{EFEFEF}{\color[HTML]{000000} ${37.48}_{1.92}$} & \cellcolor[HTML]{EFEFEF}{\color[HTML]{00ba55} $22.87$} \\ \midrule
{\color[HTML]{000000} } & \multicolumn{3}{c|}{{\color[HTML]{000000} $\lambda$}} & {\color[HTML]{000000} 0.02} & {\color[HTML]{000000} 0.1} & {\color[HTML]{000000} 0.08} & {\color[HTML]{000000} 0.08} & {\color[HTML]{000000} 0.006} & {\color[HTML]{000000} 0.002} & {\color[HTML]{000000} 0.2} & {\color[HTML]{000000} 0.04} & {\color[HTML]{000000} 0.02} & {\color[HTML]{000000} 0.02} & {\color[HTML]{000000} 0.04} & {\color[HTML]{000000} 0.2} & {\color[HTML]{000000} ---} \\ \cmidrule(l){2-17} 
{\color[HTML]{000000} } & {\color[HTML]{000000} } & {\color[HTML]{000000} } & {\color[HTML]{000000} w/o} & {\color[HTML]{000000} ${41.86}_{1.78}$} & {\color[HTML]{000000} ${48.24}_{1.37}$} & {\color[HTML]{000000} ${46.99}_{0.84}$} & {\color[HTML]{000000} ${42.42}_{1.10}$} & {\color[HTML]{000000} ${50.88}_{1.05}$} & {\color[HTML]{000000} ${68.55}_{0.16}$} & {\color[HTML]{000000} ${45.85}_{1.27}$} & {\color[HTML]{000000} ${49.10}_{1.15}$} & {\color[HTML]{000000} ${56.56}_{1.31}$} & {\color[HTML]{000000} ${29.06}_{0.71}$} & {\color[HTML]{000000} ${39.39}_{1.51}$} & {\color[HTML]{000000} ${42.81}_{0.63}$} & {\color[HTML]{000000} $46.81$} \\
{\color[HTML]{000000} } & {\color[HTML]{000000} } & \multirow{-2}{*}{{\color[HTML]{000000} Val.}} & {\color[HTML]{000000} w/} & {\color[HTML]{00ba55} ${47.23}_{3.35}$} & {\color[HTML]{00ba55} ${48.73}_{6.40}$} & {\color[HTML]{00ba55} ${52.85}_{5.89}$} & {\color[HTML]{00ba55} ${46.62}_{1.59}$} & {\color[HTML]{00ba55} ${51.80}_{1.48}$} & {\color[HTML]{00ba55} ${68.89}_{0.43}$} & {\color[HTML]{00ba55} ${52.84}_{2.43}$} & {\color[HTML]{00ba55} ${61.31}_{0.81}$} & {\color[HTML]{00ba55} ${59.45}_{1.20}$} & {\color[HTML]{00ba55} ${30.29}_{0.82}$} & {\color[HTML]{00ba55} ${45.70}_{1.46}$} & {\color[HTML]{00ba55} ${51.09}_{4.07}$} & {\color[HTML]{00ba55} $51.40$} \\ \cmidrule(l){3-17} 
{\color[HTML]{000000} } & {\color[HTML]{000000} } & {\color[HTML]{000000} } & {\color[HTML]{000000} w/o} & {\color[HTML]{000000} ${38.26}_{1.37}$} & {\color[HTML]{00ba55} ${52.04}_{0.91}$} & {\color[HTML]{000000} ${42.52}_{1.15}$} & {\color[HTML]{000000} ${39.71}_{1.47}$} & {\color[HTML]{000000} ${49.67}_{0.80}$} & {\color[HTML]{000000} ${62.77}_{0.39}$} & {\color[HTML]{000000} ${48.26}_{1.00}$} & {\color[HTML]{000000} ${47.42}_{1.45}$} & {\color[HTML]{000000} ${58.96}_{1.07}$} & {\color[HTML]{000000} ${28.47}_{0.97}$} & {\color[HTML]{000000} ${32.11}_{1.37}$} & {\color[HTML]{000000} ${44.74}_{0.82}$} & {\color[HTML]{000000} $45.41$} \\
{\color[HTML]{000000} } & \multirow{-4}{*}{{\color[HTML]{000000} \begin{tabular}[c]{@{}c@{}}Acc.\\ (\%)\end{tabular}}} & \multirow{-2}{*}{{\color[HTML]{000000} Test}} & w/ & \cellcolor[HTML]{EFEFEF}{\color[HTML]{00ba55} ${40.95}_{1.84}$} & \cellcolor[HTML]{EFEFEF}{\color[HTML]{000000} ${45.52}_{3.99}$} & \cellcolor[HTML]{EFEFEF}{\color[HTML]{00ba55} ${50.12}_{3.58}$} & \cellcolor[HTML]{EFEFEF}{\color[HTML]{00ba55} ${40.31}_{0.90}$} & \cellcolor[HTML]{EFEFEF}{\color[HTML]{00ba55} ${50.64}_{1.87}$} & \cellcolor[HTML]{EFEFEF}{\color[HTML]{00ba55} ${57.34}_{1.01}$} & \cellcolor[HTML]{EFEFEF}{\color[HTML]{00ba55} ${51.76}_{2.27}$} & \cellcolor[HTML]{EFEFEF}{\color[HTML]{00ba55} ${59.41}_{1.11}$} & \cellcolor[HTML]{EFEFEF}{\color[HTML]{00ba55} ${61.33}_{1.51}$} & \cellcolor[HTML]{EFEFEF}{\color[HTML]{00ba55} ${28.49}_{1.04}$} & \cellcolor[HTML]{EFEFEF}{\color[HTML]{00ba55} ${40.29}_{3.18}$} & \cellcolor[HTML]{EFEFEF}{\color[HTML]{00ba55} ${53.01}_{3.31}$} & \cellcolor[HTML]{EFEFEF}{\color[HTML]{00ba55} $48.26$} \\ \cmidrule(l){2-17} 
{\color[HTML]{000000} } & {\color[HTML]{000000} } & {\color[HTML]{000000} } & {\color[HTML]{000000} w/o} & {\color[HTML]{000000} ${25.20}_{1.70}$} & {\color[HTML]{000000} ${21.61}_{1.20}$} & {\color[HTML]{00ba55} ${36.33}_{1.10}$} & {\color[HTML]{00ba55} ${36.50}_{0.51}$} & {\color[HTML]{000000} ${50.38}_{1.03}$} & {\color[HTML]{000000} ${41.38}_{0.39}$} & {\color[HTML]{000000} ${41.44}_{1.75}$} & {\color[HTML]{00ba55} ${33.91}_{1.20}$} & {\color[HTML]{000000} ${55.30}_{1.30}$} & {\color[HTML]{00ba55} ${20.40}_{1.31}$} & {\color[HTML]{00ba55} ${35.73}_{1.63}$} & {\color[HTML]{000000} ${33.12}_{1.17}$} & {\color[HTML]{000000} $35.94$} \\
{\color[HTML]{000000} } & {\color[HTML]{000000} } & \multirow{-2}{*}{{\color[HTML]{000000} Val.}} & {\color[HTML]{000000} w} & {\color[HTML]{00ba55} ${25.40}_{0.74}$} & {\color[HTML]{00ba55} ${21.77}_{1.89}$} & {\color[HTML]{000000} ${32.89}_{2.37}$} & {\color[HTML]{000000} ${28.58}_{2.09}$} & {\color[HTML]{00ba55} ${51.21}_{1.53}$} & {\color[HTML]{00ba55} ${42.70}_{0.82}$} & {\color[HTML]{00ba55} ${48.94}_{1.40}$} & {\color[HTML]{000000} ${31.66}_{2.16}$} & {\color[HTML]{00ba55} ${59.17}_{1.11}$} & {\color[HTML]{000000} ${19.12}_{0.69}$} & {\color[HTML]{000000} ${30.66}_{1.33}$} & {\color[HTML]{00ba55} ${46.38}_{4.14}$} & {\color[HTML]{00ba55} $36.54$} \\ \cmidrule(l){3-17} 
{\color[HTML]{000000} } & {\color[HTML]{000000} } & {\color[HTML]{000000} } & {\color[HTML]{000000} w/o} & {\color[HTML]{000000} ${24.91}_{1.27}$} & {\color[HTML]{00ba55} ${20.54}_{0.43}$} & {\color[HTML]{00ba55} ${35.57}_{0.99}$} & {\color[HTML]{00ba55} ${38.02}_{1.35}$} & {\color[HTML]{000000} ${48.42}_{0.93}$} & {\color[HTML]{000000} ${39.80}_{0.37}$} & {\color[HTML]{000000} ${42.83}_{1.35}$} & {\color[HTML]{00ba55} ${33.29}_{1.18}$} & {\color[HTML]{000000} ${58.84}_{1.10}$} & {\color[HTML]{00ba55} ${20.96}_{1.15}$} & {\color[HTML]{000000} ${31.99}_{1.36}$} & {\color[HTML]{000000} ${40.79}_{1.13}$} & {\color[HTML]{000000} $36.33$} \\
{\color[HTML]{000000} } & \multirow{-4}{*}{{\color[HTML]{000000} \begin{tabular}[c]{@{}c@{}}MF1\\ (\%)\end{tabular}}} & \multirow{-2}{*}{{\color[HTML]{000000} Test}} & w/ & \cellcolor[HTML]{EFEFEF}{\color[HTML]{00ba55} ${25.63}_{1.20}$} & \cellcolor[HTML]{EFEFEF}{\color[HTML]{000000} ${18.73}_{0.73}$} & \cellcolor[HTML]{EFEFEF}{\color[HTML]{000000} ${31.81}_{1.94}$} & \cellcolor[HTML]{EFEFEF}{\color[HTML]{000000} ${28.97}_{2.76}$} & \cellcolor[HTML]{EFEFEF}{\color[HTML]{00ba55} ${49.44}_{1.83}$} & \cellcolor[HTML]{EFEFEF}{\color[HTML]{00ba55} ${44.06}_{1.11}$} & \cellcolor[HTML]{EFEFEF}{\color[HTML]{00ba55} ${45.99}_{4.55}$} & \cellcolor[HTML]{EFEFEF}{\color[HTML]{000000} ${33.03}_{1.60}$} & \cellcolor[HTML]{EFEFEF}{\color[HTML]{00ba55} ${59.72}_{1.13}$} & \cellcolor[HTML]{EFEFEF}{\color[HTML]{000000} ${18.40}_{1.10}$} & \cellcolor[HTML]{EFEFEF}{\color[HTML]{00ba55} ${33.02}_{1.48}$} & \cellcolor[HTML]{EFEFEF}{\color[HTML]{00ba55} ${50.92}_{1.60}$} & \cellcolor[HTML]{EFEFEF}{\color[HTML]{00ba55} $36.64$} \\ \cmidrule(l){2-17} 
{\color[HTML]{000000} } & {\color[HTML]{000000} } & {\color[HTML]{000000} } & {\color[HTML]{000000} w/o} & {\color[HTML]{000000} ${10.21}_{1.71}$} & {\color[HTML]{00ba55} ${25.07}_{1.21}$} & {\color[HTML]{000000} ${14.93}_{1.00}$} & {\color[HTML]{00ba55} ${21.00}_{1.10}$} & {\color[HTML]{000000} ${35.37}_{1.19}$} & {\color[HTML]{000000} ${20.19}_{0.40}$} & {\color[HTML]{000000} ${41.29}_{1.34}$} & {\color[HTML]{000000} ${15.87}_{1.07}$} & {\color[HTML]{000000} ${6.43}_{1.22}$} & {\color[HTML]{00ba55} ${30.94}_{0.72}$} & {\color[HTML]{000000} ${18.02}_{1.74}$} & {\color[HTML]{000000} ${46.90}_{0.60}$} & {\color[HTML]{000000} $23.85$} \\
{\color[HTML]{000000} } & {\color[HTML]{000000} } & \multirow{-2}{*}{{\color[HTML]{000000} Val.}} & {\color[HTML]{000000} w/} & {\color[HTML]{00ba55} ${5.92}_{1.26}$} & {\color[HTML]{000000} ${29.23}_{4.69}$} & {\color[HTML]{00ba55} ${12.70}_{1.65}$} & {\color[HTML]{000000} ${22.24}_{1.86}$} & {\color[HTML]{00ba55} ${34.69}_{1.23}$} & {\color[HTML]{00ba55} ${19.18}_{0.61}$} & {\color[HTML]{00ba55} ${21.02}_{2.77}$} & {\color[HTML]{00ba55} ${10.53}_{2.79}$} & {\color[HTML]{00ba55} ${3.97}_{1.86}$} & {\color[HTML]{000000} ${35.02}_{0.56}$} & {\color[HTML]{00ba55} ${16.78}_{1.71}$} & {\color[HTML]{00ba55} ${27.18}_{4.94}$} & {\color[HTML]{00ba55} $19.87$} \\ \cmidrule(l){3-17} 
{\color[HTML]{000000} } & {\color[HTML]{000000} } & {\color[HTML]{000000} } & {\color[HTML]{000000} w/o} & {\color[HTML]{000000} ${12.48}_{1.44}$} & {\color[HTML]{00ba55} ${19.52}_{0.68}$} & {\color[HTML]{000000} ${17.95}_{1.41}$} & {\color[HTML]{00ba55} ${20.35}_{1.46}$} & {\color[HTML]{000000} ${36.73}_{0.99}$} & {\color[HTML]{000000} ${23.63}_{0.44}$} & {\color[HTML]{000000} ${39.93}_{1.27}$} & {\color[HTML]{000000} ${17.40}_{1.25}$} & {\color[HTML]{000000} ${4.08}_{1.05}$} & {\color[HTML]{00ba55} ${31.48}_{1.10}$} & {\color[HTML]{000000} ${24.73}_{1.44}$} & {\color[HTML]{000000} ${42.57}_{1.24}$} & {\color[HTML]{000000} $24.24$} \\
\multirow{-13}{*}{{\color[HTML]{000000} GPT-2}} & \multirow{-4}{*}{{\color[HTML]{000000} \begin{tabular}[c]{@{}c@{}}$ECE_1$\\ (\%, $\downarrow$)\end{tabular}}} & \multirow{-2}{*}{{\color[HTML]{000000} Test}} & w/ & \cellcolor[HTML]{EFEFEF}{\color[HTML]{00ba55} ${9.58}_{1.56}$} & \cellcolor[HTML]{EFEFEF}{\color[HTML]{000000} ${33.12}_{4.08}$} & \cellcolor[HTML]{EFEFEF}{\color[HTML]{00ba55} ${16.31}_{1.69}$} & \cellcolor[HTML]{EFEFEF}{\color[HTML]{000000} ${28.04}_{3.60}$} & \cellcolor[HTML]{EFEFEF}{\color[HTML]{00ba55} ${36.06}_{1.95}$} & \cellcolor[HTML]{EFEFEF}{\color[HTML]{00ba55} ${17.86}_{0.86}$} & \cellcolor[HTML]{EFEFEF}{\color[HTML]{00ba55} ${24.58}_{6.04}$} & \cellcolor[HTML]{EFEFEF}{\color[HTML]{00ba55} ${8.99}_{2.05}$} & \cellcolor[HTML]{EFEFEF}{\color[HTML]{00ba55} ${2.09}_{0.84}$} & \cellcolor[HTML]{EFEFEF}{\color[HTML]{000000} ${37.68}_{1.94}$} & \cellcolor[HTML]{EFEFEF}{\color[HTML]{00ba55} ${15.45}_{1.96}$} & \cellcolor[HTML]{EFEFEF}{\color[HTML]{00ba55} ${21.12}_{2.08}$} & \cellcolor[HTML]{EFEFEF}{\color[HTML]{00ba55} $20.91$} \\ \bottomrule
\end{tabular}
}
\end{table*}

% k = 8 table
\begin{table*}[t]
\caption{Accuracy and Macro-F1 results ($\%$, $mean_{std}$, $k=8$). A better result is in {\color[HTML]{00ba55}green}. Notation is the same with the Tabel~\ref{tab:1}. Some experiments can't be conducted due to the length of the input sequence.}
\label{tab:K8}
\centering
\resizebox{2\columnwidth}{!}{
\begin{tabular}{@{}cccc|cccccccccccc|c@{}}
\toprule
\multicolumn{4}{c|}{{\color[HTML]{000000} Dataset}} & {\color[HTML]{000000} PS} & {\color[HTML]{000000} HS} & {\color[HTML]{000000} SE'14R} & {\color[HTML]{000000} SE'14L} & {\color[HTML]{000000} RTE} & {\color[HTML]{000000} MRPC} & {\color[HTML]{000000} Ethos} & {\color[HTML]{000000} FP} & {\color[HTML]{000000} SST2} & {\color[HTML]{000000} TEE} & {\color[HTML]{000000} TES} & {\color[HTML]{000000} TEH} & {\color[HTML]{000000} Mean} \\ \midrule
{\color[HTML]{000000} } & \multicolumn{3}{c|}{{\color[HTML]{000000} $\lambda$}} & {\color[HTML]{000000} 0.2} & {\color[HTML]{000000} 0.2} & {\color[HTML]{000000} 0.1} & {\color[HTML]{000000} 0.1} & {\color[HTML]{000000} 0.1} & {\color[HTML]{000000} 0.2} & {\color[HTML]{000000} 0.004} & {\color[HTML]{000000} 0.012} & {\color[HTML]{000000} 0.002} & {\color[HTML]{000000} 0.006} & {\color[HTML]{000000} 0.08} & {\color[HTML]{000000} 0.002} & {\color[HTML]{000000} ---} \\ \cmidrule(l){2-17} 
{\color[HTML]{000000} } & {\color[HTML]{000000} } & {\color[HTML]{000000} } & {\color[HTML]{000000} w/o} & {\color[HTML]{000000} ${27.32}_{0.84}$} & {\color[HTML]{000000} ${68.87}_{1.30}$} & {\color[HTML]{000000} ${35.61}_{0.57}$} & {\color[HTML]{000000} ${33.57}_{0.74}$} & {\color[HTML]{000000} ${50.12}_{0.29}$} & {\color[HTML]{000000} ${35.86}_{0.55}$} & {\color[HTML]{000000} ${57.29}_{1.69}$} & {\color[HTML]{000000} ${62.34}_{0.05}$} & {\color[HTML]{00ba55} ${83.96}_{0.86}$} & {\color[HTML]{000000} ${47.64}_{0.52}$} & {\color[HTML]{000000} ${51.13}_{0.66}$} & {\color[HTML]{000000} ${50.62}_{1.62}$} & {\color[HTML]{000000} $50.37$} \\
{\color[HTML]{000000} } & {\color[HTML]{000000} } & \multirow{-2}{*}{{\color[HTML]{000000} Val.}} & {\color[HTML]{000000} w/} & {\color[HTML]{00ba55} ${62.01}_{4.95}$} & {\color[HTML]{00ba55} ${85.29}_{0.95}$} & {\color[HTML]{00ba55} ${45.84}_{11.13}$} & {\color[HTML]{00ba55} ${40.84}_{6.41}$} & {\color[HTML]{00ba55} ${50.49}_{0.89}$} & {\color[HTML]{00ba55} ${58.87}_{7.93}$} & {\color[HTML]{00ba55} ${57.95}_{1.35}$} & {\color[HTML]{00ba55} ${62.44}_{0.08}$} & {\color[HTML]{000000} ${83.89}_{0.79}$} & {\color[HTML]{00ba55} ${47.81}_{0.90}$} & {\color[HTML]{00ba55} ${51.35}_{1.67}$} & {\color[HTML]{00ba55} ${52.05}_{0.75}$} & {\color[HTML]{00ba55} $58.24$} \\ \cmidrule(l){3-17} 
{\color[HTML]{000000} } & {\color[HTML]{000000} } & {\color[HTML]{000000} } & {\color[HTML]{000000} w/o} & {\color[HTML]{000000} ${25.70}_{0.84}$} & {\color[HTML]{000000} ${77.10}_{1.75}$} & {\color[HTML]{000000} ${33.30}_{0.63}$} & {\color[HTML]{000000} ${30.80}_{0.71}$} & {\color[HTML]{000000} ${46.35}_{1.39}$} & {\color[HTML]{000000} ${39.35}_{0.84}$} & {\color[HTML]{00ba55} ${61.67}_{1.38}$} & {\color[HTML]{000000} ${61.76}_{0.06}$} & {\color[HTML]{00ba55} ${80.59}_{1.02}$} & {\color[HTML]{000000} ${46.37}_{0.50}$} & {\color[HTML]{00ba55} ${50.68}_{0.35}$} & {\color[HTML]{000000} ${54.57}_{1.41}$} & {\color[HTML]{000000} $50.69$} \\
{\color[HTML]{000000} } & \multirow{-4}{*}{{\color[HTML]{000000} \begin{tabular}[c]{@{}c@{}}Acc.\\ (\%)\end{tabular}}} & \multirow{-2}{*}{{\color[HTML]{000000} Test}} & w/ & \cellcolor[HTML]{EFEFEF}{\color[HTML]{00ba55} ${58.54}_{3.27}$} & \cellcolor[HTML]{EFEFEF}{\color[HTML]{00ba55} ${89.74}_{3.97}$} & \cellcolor[HTML]{EFEFEF}{\color[HTML]{00ba55} ${41.68}_{6.66}$} & \cellcolor[HTML]{EFEFEF}{\color[HTML]{00ba55} ${35.66}_{4.65}$} & \cellcolor[HTML]{EFEFEF}{\color[HTML]{00ba55} ${47.53}_{1.26}$} & \cellcolor[HTML]{EFEFEF}{\color[HTML]{00ba55} ${54.57}_{8.48}$} & \cellcolor[HTML]{EFEFEF}{\color[HTML]{000000} ${61.46}_{0.96}$} & \cellcolor[HTML]{EFEFEF}{\color[HTML]{00ba55} ${61.78}_{0.08}$} & \cellcolor[HTML]{EFEFEF}{\color[HTML]{000000} ${79.78}_{0.69}$} & \cellcolor[HTML]{EFEFEF}{\color[HTML]{00ba55} ${46.42}_{0.63}$} & \cellcolor[HTML]{EFEFEF}{\color[HTML]{000000} ${50.62}_{0.41}$} & \cellcolor[HTML]{EFEFEF}{\color[HTML]{00ba55} ${54.68}_{0.92}$} & \cellcolor[HTML]{EFEFEF}{\color[HTML]{00ba55} $56.87$} \\ \cmidrule(l){2-17} 
{\color[HTML]{000000} } & {\color[HTML]{000000} } & {\color[HTML]{000000} } & {\color[HTML]{000000} w/o} & {\color[HTML]{000000} ${16.07}_{0.47}$} & {\color[HTML]{000000} ${23.66}_{0.32}$} & {\color[HTML]{000000} ${30.15}_{1.14}$} & {\color[HTML]{000000} ${24.48}_{1.40}$} & {\color[HTML]{000000} ${45.66}_{0.49}$} & {\color[HTML]{000000} ${33.37}_{0.56}$} & {\color[HTML]{000000} ${56.68}_{1.81}$} & {\color[HTML]{000000} ${25.91}_{0.26}$} & {\color[HTML]{00ba55} ${83.80}_{0.91}$} & {\color[HTML]{000000} ${27.20}_{0.59}$} & {\color[HTML]{000000} ${33.72}_{1.47}$} & {\color[HTML]{000000} ${48.52}_{1.98}$} & {\color[HTML]{000000} $37.44$} \\
{\color[HTML]{000000} } & {\color[HTML]{000000} } & \multirow{-2}{*}{{\color[HTML]{000000} Val.}} & {\color[HTML]{000000} w/} & {\color[HTML]{00ba55} ${20.91}_{1.79}$} & {\color[HTML]{00ba55} ${27.76}_{1.35}$} & {\color[HTML]{00ba55} ${41.70}_{10.30}$} & {\color[HTML]{00ba55} ${35.28}_{8.07}$} & {\color[HTML]{00ba55} ${45.89}_{3.44}$} & {\color[HTML]{00ba55} ${46.69}_{2.90}$} & \cellcolor[HTML]{FFFFFF}{\color[HTML]{00ba55} ${57.45}_{1.37}$} & {\color[HTML]{00ba55} ${26.24}_{0.36}$} & {\color[HTML]{000000} ${83.72}_{0.82}$} & {\color[HTML]{00ba55} ${27.39}_{0.84}$} & {\color[HTML]{00ba55} ${36.84}_{5.55}$} & {\color[HTML]{00ba55} ${49.94}_{0.99}$} & {\color[HTML]{00ba55} $41.65$} \\ \cmidrule(l){3-17} 
{\color[HTML]{000000} } & {\color[HTML]{000000} } & {\color[HTML]{000000} } & {\color[HTML]{000000} w/o} & {\color[HTML]{000000} ${16.01}_{0.94}$} & {\color[HTML]{00ba55} ${26.26}_{1.29}$} & {\color[HTML]{000000} ${28.72}_{0.67}$} & {\color[HTML]{000000} ${27.21}_{0.78}$} & {\color[HTML]{00ba55} ${42.58}_{1.37}$} & {\color[HTML]{000000} ${33.58}_{0.98}$} & {\color[HTML]{00ba55} ${60.88}_{1.44}$} & {\color[HTML]{000000} ${25.60}_{0.25}$} & {\color[HTML]{00ba55} ${78.98}_{1.18}$} & {\color[HTML]{000000} ${24.50}_{0.70}$} & {\color[HTML]{000000} ${33.19}_{0.40}$} & {\color[HTML]{000000} ${54.36}_{1.42}$} & {\color[HTML]{000000} $37.66$} \\
{\color[HTML]{000000} } & \multirow{-4}{*}{{\color[HTML]{000000} \begin{tabular}[c]{@{}c@{}}MF1\\ (\%)\end{tabular}}} & \multirow{-2}{*}{{\color[HTML]{000000} Test}} & w/ & \cellcolor[HTML]{EFEFEF}{\color[HTML]{00ba55} ${20.81}_{1.17}$} & \cellcolor[HTML]{EFEFEF}{\color[HTML]{000000} ${24.29}_{0.56}$} & \cellcolor[HTML]{EFEFEF}{\color[HTML]{00ba55} ${39.15}_{6.23}$} & \cellcolor[HTML]{EFEFEF}{\color[HTML]{00ba55} ${34.28}_{5.55}$} & \cellcolor[HTML]{EFEFEF}{\color[HTML]{000000} ${42.47}_{4.45}$} & \cellcolor[HTML]{EFEFEF}{\color[HTML]{00ba55} ${42.45}_{4.20}$} & \cellcolor[HTML]{EFEFEF}{\color[HTML]{000000} ${60.52}_{1.02}$} & \cellcolor[HTML]{EFEFEF}{\color[HTML]{00ba55} ${25.67}_{0.31}$} & \cellcolor[HTML]{EFEFEF}{\color[HTML]{000000} ${78.06}_{0.79}$} & \cellcolor[HTML]{EFEFEF}{\color[HTML]{00ba55} ${24.80}_{0.92}$} & \cellcolor[HTML]{EFEFEF}{\color[HTML]{00ba55} ${33.85}_{3.11}$} & \cellcolor[HTML]{EFEFEF}{\color[HTML]{00ba55} ${54.45}_{0.96}$} & \cellcolor[HTML]{EFEFEF}{\color[HTML]{00ba55} $40.07$} \\ \cmidrule(l){2-17} 
{\color[HTML]{000000} } & {\color[HTML]{000000} } & {\color[HTML]{000000} } & {\color[HTML]{000000} w/o} & {\color[HTML]{000000} ${42.98}_{0.85}$} & {\color[HTML]{000000} ${10.43}_{1.20}$} & {\color[HTML]{000000} ${32.79}_{0.53}$} & {\color[HTML]{000000} ${42.43}_{0.99}$} & {\color[HTML]{00ba55} ${21.59}_{0.34}$} & {\color[HTML]{000000} ${33.19}_{0.72}$} & {\color[HTML]{000000} ${13.63}_{1.44}$} & {\color[HTML]{000000} ${24.54}_{0.13}$} & {\color[HTML]{00ba55} ${12.61}_{0.98}$} & {\color[HTML]{000000} ${40.10}_{0.65}$} & {\color[HTML]{000000} ${13.99}_{0.17}$} & {\color[HTML]{000000} ${21.11}_{1.51}$} & {\color[HTML]{000000} $25.78$} \\
{\color[HTML]{000000} } & {\color[HTML]{000000} } & \multirow{-2}{*}{{\color[HTML]{000000} Val.}} & {\color[HTML]{000000} w/} & {\color[HTML]{00ba55} ${11.81}_{4.56}$} & {\color[HTML]{00ba55} ${9.51}_{2.14}$} & {\color[HTML]{00ba55} ${19.27}_{11.45}$} & {\color[HTML]{00ba55} ${25.96}_{10.94}$} & {\color[HTML]{000000} ${23.13}_{2.74}$} & {\color[HTML]{00ba55} ${8.10}_{2.84}$} & {\color[HTML]{00ba55} ${12.47}_{1.21}$} & {\color[HTML]{00ba55} ${24.33}_{0.69}$} & {\color[HTML]{000000} ${12.80}_{0.75}$} & {\color[HTML]{00ba55} ${39.95}_{1.00}$} & {\color[HTML]{00ba55} ${7.25}_{4.79}$} & {\color[HTML]{00ba55} ${20.01}_{0.66}$} & {\color[HTML]{00ba55} $17.88$} \\ \cmidrule(l){3-17} 
{\color[HTML]{000000} } & {\color[HTML]{000000} } & {\color[HTML]{000000} } & {\color[HTML]{000000} w/o} & {\color[HTML]{000000} ${44.92}_{0.80}$} & {\color[HTML]{000000} ${7.93}_{1.41}$} & {\color[HTML]{000000} ${36.15}_{0.98}$} & {\color[HTML]{000000} ${43.80}_{1.01}$} & {\color[HTML]{00ba55} ${25.13}_{1.62}$} & {\color[HTML]{000000} ${29.83}_{0.74}$} & {\color[HTML]{00ba55} ${9.35}_{1.45}$} & {\color[HTML]{000000} ${25.00}_{0.23}$} & {\color[HTML]{000000} ${10.52}_{0.94}$} & {\color[HTML]{000000} ${41.83}_{0.54}$} & {\color[HTML]{000000} ${14.13}_{0.43}$} & {\color[HTML]{000000} ${16.78}_{1.45}$} & {\color[HTML]{000000} $25.45$} \\
\multirow{-13}{*}{{\color[HTML]{000000} GPT-J}} & \multirow{-4}{*}{{\color[HTML]{000000} \begin{tabular}[c]{@{}c@{}}$ECE_1$\\ (\%, $\downarrow$)\end{tabular}}} & \multirow{-2}{*}{{\color[HTML]{000000} Test}} & w/ & \cellcolor[HTML]{EFEFEF}{\color[HTML]{00ba55} ${8.86}_{3.57}$} & \cellcolor[HTML]{EFEFEF}{\color[HTML]{00ba55} ${6.04}_{1.94}$} & \cellcolor[HTML]{EFEFEF}{\color[HTML]{00ba55} ${22.26}_{9.50}$} & \cellcolor[HTML]{EFEFEF}{\color[HTML]{00ba55} ${29.44}_{8.54}$} & \cellcolor[HTML]{EFEFEF}{\color[HTML]{000000} ${28.55}_{6.11}$} & \cellcolor[HTML]{EFEFEF}{\color[HTML]{00ba55} ${15.28}_{7.58}$} & \cellcolor[HTML]{EFEFEF}{\color[HTML]{000000} ${9.77}_{0.86}$} & \cellcolor[HTML]{EFEFEF}{\color[HTML]{00ba55} ${24.92}_{0.52}$} & \cellcolor[HTML]{EFEFEF}{\color[HTML]{00ba55} ${9.88}_{0.66}$} & \cellcolor[HTML]{EFEFEF}{\color[HTML]{00ba55} ${41.83}_{0.80}$} & \cellcolor[HTML]{EFEFEF}{\color[HTML]{00ba55} ${9.24}_{2.47}$} & \cellcolor[HTML]{EFEFEF}{\color[HTML]{00ba55} ${16.77}_{0.95}$} & \cellcolor[HTML]{EFEFEF}{\color[HTML]{00ba55} $18.57$} \\ \bottomrule
\end{tabular}
}
\end{table*}

% k = 16 table
\begin{table*}[t]
\caption{Accuracy and Macro-F1 results ($\%$, $mean_{std}$, $k=16$). A better result is in {\color[HTML]{00ba55}green}. Notation is the same with the Tabel~\ref{tab:1}. Some experiments can't be conducted due to the length of the input sequence.}
\label{tab:K16}
\centering
\resizebox{2\columnwidth}{!}{
\begin{tabular}{@{}cccc|cccccccccc|c@{}}
\toprule
\multicolumn{4}{c|}{{\color[HTML]{000000} Dataset}} & {\color[HTML]{000000} PS} & {\color[HTML]{000000} HS} & {\color[HTML]{000000} SE'14R} & {\color[HTML]{000000} SE'14L} & {\color[HTML]{000000} RTE} & {\color[HTML]{000000} FP} & {\color[HTML]{000000} SST2} & {\color[HTML]{000000} TEE} & {\color[HTML]{000000} TES} & {\color[HTML]{000000} TEH} & {\color[HTML]{000000} Mean} \\ \midrule
{\color[HTML]{000000} } & \multicolumn{3}{c|}{{\color[HTML]{000000} $\lambda$}} & {\color[HTML]{000000} 0.2} & {\color[HTML]{000000} 0.2} & {\color[HTML]{000000} 0.1} & {\color[HTML]{000000} 0.1} & {\color[HTML]{000000} 0.2} & {\color[HTML]{000000} 0.1} & {\color[HTML]{000000} 0.004} & {\color[HTML]{000000} 0.008} & {\color[HTML]{000000} 0.08} & {\color[HTML]{000000} 0.002} & {\color[HTML]{000000} ---} \\ \cmidrule(l){2-15} 
{\color[HTML]{000000} } & {\color[HTML]{000000} } & {\color[HTML]{000000} } & {\color[HTML]{000000} w/o} & {\color[HTML]{000000} ${18.52}_{0.78}$} & {\color[HTML]{000000} ${70.02}_{0.52}$} & {\color[HTML]{000000} ${40.86}_{0.90}$} & {\color[HTML]{000000} ${34.59}_{0.41}$} & {\color[HTML]{000000} ${48.91}_{0.62}$} & {\color[HTML]{000000} ${62.42}_{0.10}$} & {\color[HTML]{000000} ${88.24}_{0.70}$} & {\color[HTML]{000000} ${49.22}_{0.47}$} & {\color[HTML]{000000} ${51.84}_{0.97}$} & {\color[HTML]{000000} ${54.59}_{0.87}$} & {\color[HTML]{000000} $51.92$} \\
{\color[HTML]{000000} } & {\color[HTML]{000000} } & \multirow{-2}{*}{{\color[HTML]{000000} Val.}} & {\color[HTML]{000000} w/} & {\color[HTML]{00ba55} ${61.78}_{5.94}$} & {\color[HTML]{00ba55} ${85.98}_{0.61}$} & {\color[HTML]{00ba55} ${46.84}_{12.92}$} & {\color[HTML]{00ba55} ${43.01}_{8.67}$} & {\color[HTML]{00ba55} ${50.76}_{1.15}$} & {\color[HTML]{00ba55} ${62.66}_{0.35}$} & {\color[HTML]{00ba55} ${88.83}_{0.71}$} & {\color[HTML]{00ba55} ${49.30}_{0.53}$} & {\color[HTML]{00ba55} ${53.11}_{1.99}$} & {\color[HTML]{00ba55} ${55.20}_{0.91}$} & {\color[HTML]{00ba55} $59.75$} \\ \cmidrule(l){3-15} 
{\color[HTML]{000000} } & {\color[HTML]{000000} } & {\color[HTML]{000000} } & {\color[HTML]{000000} w/o} & {\color[HTML]{000000} ${18.75}_{0.38}$} & {\color[HTML]{000000} ${74.97}_{1.43}$} & {\color[HTML]{000000} ${34.98}_{1.10}$} & {\color[HTML]{000000} ${30.00}_{0.58}$} & {\color[HTML]{000000} ${45.95}_{0.45}$} & {\color[HTML]{000000} ${61.82}_{0.09}$} & {\color[HTML]{000000} ${81.88}_{1.00}$} & {\color[HTML]{00ba55} ${47.11}_{0.35}$} & {\color[HTML]{000000} ${51.70}_{0.42}$} & {\color[HTML]{000000} ${55.90}_{0.81}$} & {\color[HTML]{000000} $50.31$} \\
{\color[HTML]{000000} } & \multirow{-4}{*}{{\color[HTML]{000000} \begin{tabular}[c]{@{}c@{}}Acc.\\ (\%)\end{tabular}}} & \multirow{-2}{*}{{\color[HTML]{000000} Test}} & w/ & \cellcolor[HTML]{EFEFEF}{\color[HTML]{00ba55} ${59.97}_{2.40}$} & \cellcolor[HTML]{EFEFEF}{\color[HTML]{00ba55} ${89.96}_{4.15}$} & \cellcolor[HTML]{EFEFEF}{\color[HTML]{00ba55} ${42.56}_{9.98}$} & \cellcolor[HTML]{EFEFEF}{\color[HTML]{00ba55} ${38.55}_{7.65}$} & \cellcolor[HTML]{EFEFEF}{\color[HTML]{00ba55} ${52.96}_{1.39}$} & \cellcolor[HTML]{EFEFEF}{\color[HTML]{00ba55} ${61.92}_{0.49}$} & \cellcolor[HTML]{EFEFEF}{\color[HTML]{00ba55} ${82.20}_{0.86}$} & \cellcolor[HTML]{EFEFEF}{\color[HTML]{000000} ${47.06}_{0.65}$} & \cellcolor[HTML]{EFEFEF}{\color[HTML]{00ba55} ${52.13}_{0.88}$} & \cellcolor[HTML]{EFEFEF}{\color[HTML]{00ba55} ${56.29}_{1.21}$} & \cellcolor[HTML]{EFEFEF}{\color[HTML]{00ba55} $58.36$} \\ \cmidrule(l){2-15} 
{\color[HTML]{000000} } & {\color[HTML]{000000} } & {\color[HTML]{000000} } & {\color[HTML]{000000} w/o} & {\color[HTML]{000000} ${10.33}_{0.46}$} & {\color[HTML]{00ba55} ${30.44}_{0.84}$} & {\color[HTML]{000000} ${32.79}_{0.78}$} & {\color[HTML]{000000} ${24.15}_{0.66}$} & {\color[HTML]{000000} ${38.73}_{0.71}$} & {\color[HTML]{000000} ${26.08}_{0.46}$} & {\color[HTML]{000000} ${88.22}_{0.70}$} & {\color[HTML]{000000} ${28.75}_{0.40}$} & {\color[HTML]{000000} ${36.01}_{1.76}$} & {\color[HTML]{000000} ${53.61}_{0.88}$} & {\color[HTML]{000000} $36.91$} \\
{\color[HTML]{000000} } & {\color[HTML]{000000} } & \multirow{-2}{*}{{\color[HTML]{000000} Val.}} & {\color[HTML]{000000} w/} & {\color[HTML]{00ba55} ${20.00}_{0.71}$} & {\color[HTML]{000000} ${23.54}_{0.57}$} & {\color[HTML]{00ba55} ${41.11}_{11.38}$} & {\color[HTML]{00ba55} ${35.07}_{10.38}$} & {\color[HTML]{00ba55} ${42.85}_{7.50}$} & {\color[HTML]{00ba55} ${27.18}_{1.53}$} & {\color[HTML]{00ba55} ${88.81}_{0.72}$} & {\color[HTML]{00ba55} ${29.04}_{0.55}$} & {\color[HTML]{00ba55} ${40.22}_{6.60}$} & {\color[HTML]{00ba55} ${54.02}_{1.02}$} & {\color[HTML]{00ba55} $40.18$} \\ \cmidrule(l){3-15} 
{\color[HTML]{000000} } & {\color[HTML]{000000} } & {\color[HTML]{000000} } & {\color[HTML]{000000} w/o} & {\color[HTML]{000000} ${9.61}_{0.26}$} & {\color[HTML]{00ba55} ${27.64}_{1.69}$} & {\color[HTML]{000000} ${28.80}_{0.97}$} & {\color[HTML]{000000} ${25.05}_{0.94}$} & {\color[HTML]{000000} ${35.57}_{0.69}$} & {\color[HTML]{000000} ${25.80}_{0.31}$} & {\color[HTML]{000000} ${80.33}_{1.12}$} & {\color[HTML]{00ba55} ${25.50}_{0.54}$} & {\color[HTML]{000000} ${35.44}_{0.90}$} & {\color[HTML]{000000} ${55.76}_{0.81}$} & {\color[HTML]{000000} $34.95$} \\
{\color[HTML]{000000} } & \multirow{-4}{*}{{\color[HTML]{000000} \begin{tabular}[c]{@{}c@{}}MF1\\ (\%)\end{tabular}}} & \multirow{-2}{*}{{\color[HTML]{000000} Test}} & w/ & \cellcolor[HTML]{EFEFEF}{\color[HTML]{00ba55} ${19.70}_{1.28}$} & \cellcolor[HTML]{EFEFEF}{\color[HTML]{000000} ${23.96}_{0.08}$} & \cellcolor[HTML]{EFEFEF}{\color[HTML]{00ba55} ${37.93}_{9.63}$} & \cellcolor[HTML]{EFEFEF}{\color[HTML]{00ba55} ${36.06}_{8.97}$} & \cellcolor[HTML]{EFEFEF}{\color[HTML]{00ba55} ${41.17}_{5.54}$} & \cellcolor[HTML]{EFEFEF}{\color[HTML]{00ba55} ${26.27}_{1.96}$} & \cellcolor[HTML]{EFEFEF}{\color[HTML]{00ba55} ${80.61}_{1.12}$} & \cellcolor[HTML]{EFEFEF}{\color[HTML]{000000} ${25.45}_{0.95}$} & \cellcolor[HTML]{EFEFEF}{\color[HTML]{00ba55} ${37.20}_{3.58}$} & \cellcolor[HTML]{EFEFEF}{\color[HTML]{00ba55} ${56.13}_{1.25}$} & \cellcolor[HTML]{EFEFEF}{\color[HTML]{00ba55} $38.45$} \\ \cmidrule(l){2-15} 
{\color[HTML]{000000} } & {\color[HTML]{000000} } & {\color[HTML]{000000} } & {\color[HTML]{000000} w/o} & {\color[HTML]{000000} ${54.29}_{0.95}$} & {\color[HTML]{00ba55} ${4.01}_{0.77}$} & {\color[HTML]{000000} ${27.53}_{1.27}$} & {\color[HTML]{000000} ${40.90}_{0.52}$} & {\color[HTML]{000000} ${26.13}_{0.54}$} & {\color[HTML]{000000} ${26.34}_{0.14}$} & {\color[HTML]{00ba55} ${14.43}_{0.75}$} & {\color[HTML]{000000} ${38.97}_{0.58}$} & {\color[HTML]{000000} ${11.46}_{0.99}$} & {\color[HTML]{000000} ${13.66}_{0.72}$} & {\color[HTML]{000000} $25.77$} \\
{\color[HTML]{000000} } & {\color[HTML]{000000} } & \multirow{-2}{*}{{\color[HTML]{000000} Val.}} & {\color[HTML]{000000} w/} & {\color[HTML]{00ba55} ${11.75}_{5.60}$} & {\color[HTML]{000000} ${9.70}_{2.72}$} & {\color[HTML]{00ba55} ${18.56}_{10.81}$} & {\color[HTML]{00ba55} ${23.33}_{13.20}$} & {\color[HTML]{00ba55} ${21.01}_{13.24}$} & {\color[HTML]{00ba55} ${18.07}_{5.22}$} & {\color[HTML]{000000} ${15.49}_{0.58}$} & {\color[HTML]{00ba55} ${38.61}_{0.69}$} & {\color[HTML]{00ba55} ${6.12}_{2.85}$} & {\color[HTML]{00ba55} ${13.25}_{0.68}$} & {\color[HTML]{00ba55} $17.59$} \\ \cmidrule(l){3-15} 
{\color[HTML]{000000} } & {\color[HTML]{000000} } & {\color[HTML]{000000} } & {\color[HTML]{000000} w/o} & {\color[HTML]{000000} ${53.72}_{0.37}$} & {\color[HTML]{00ba55} ${4.51}_{0.87}$} & {\color[HTML]{000000} ${34.09}_{1.27}$} & {\color[HTML]{000000} ${43.06}_{0.62}$} & {\color[HTML]{000000} ${29.93}_{0.68}$} & {\color[HTML]{000000} ${26.90}_{0.19}$} & {\color[HTML]{00ba55} ${9.96}_{1.00}$} & {\color[HTML]{00ba55} ${40.84}_{0.45}$} & {\color[HTML]{000000} ${11.24}_{0.41}$} & {\color[HTML]{000000} ${13.45}_{0.86}$} & {\color[HTML]{000000} $26.77$} \\
\multirow{-13}{*}{{\color[HTML]{000000} GPT-J}} & \multirow{-4}{*}{{\color[HTML]{000000} \begin{tabular}[c]{@{}c@{}}$ECE_1$\\ (\%, $\downarrow$)\end{tabular}}} & \multirow{-2}{*}{{\color[HTML]{000000} Test}} & w/ & \cellcolor[HTML]{EFEFEF}{\color[HTML]{00ba55} ${8.33}_{4.47}$} & \cellcolor[HTML]{EFEFEF}{\color[HTML]{000000} ${5.63}_{2.27}$} & \cellcolor[HTML]{EFEFEF}{\color[HTML]{00ba55} ${20.29}_{12.35}$} & \cellcolor[HTML]{EFEFEF}{\color[HTML]{00ba55} ${24.94}_{10.05}$} & \cellcolor[HTML]{EFEFEF}{\color[HTML]{00ba55} ${17.53}_{7.30}$} & \cellcolor[HTML]{EFEFEF}{\color[HTML]{00ba55} ${20.44}_{3.90}$} & \cellcolor[HTML]{EFEFEF}{\color[HTML]{000000} ${9.99}_{0.89}$} & \cellcolor[HTML]{EFEFEF}{\color[HTML]{000000} ${41.00}_{0.83}$} & \cellcolor[HTML]{EFEFEF}{\color[HTML]{00ba55} ${5.19}_{2.52}$} & \cellcolor[HTML]{EFEFEF}{\color[HTML]{00ba55} ${13.03}_{1.33}$} & \cellcolor[HTML]{EFEFEF}{\color[HTML]{00ba55} $16.64$} \\ \bottomrule
\end{tabular}
}
\end{table*}

\begin{figure}[t]
    \centering
    \centerline{\includegraphics[width=\linewidth]{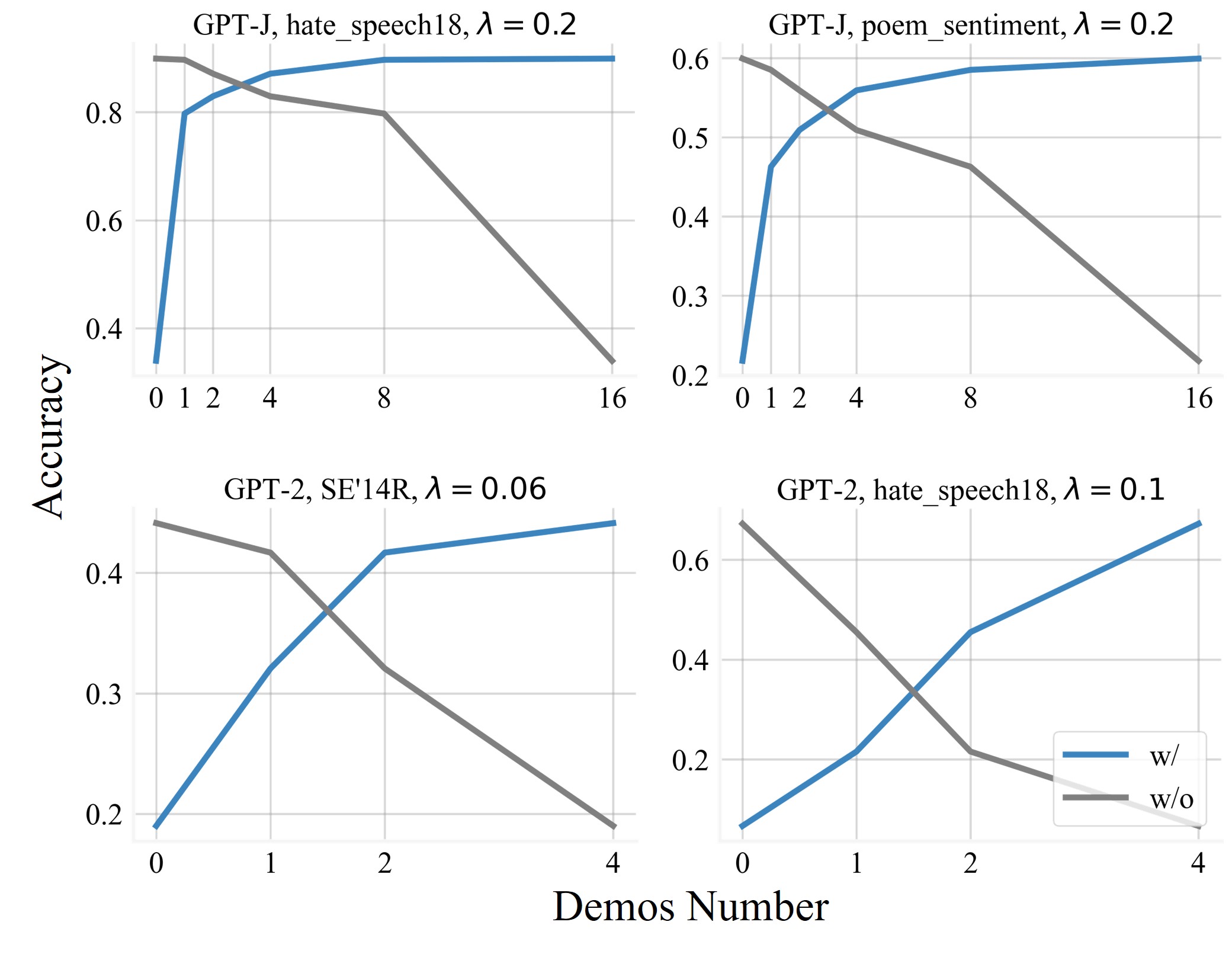}}
    \caption{The relationship between demos quantity and accuracy in some cases. \M\ can make the model learn from the demos correctly.}
    \label{fig:demo}
\end{figure}

\begin{figure}[t]
    \centering
    \centerline{\includegraphics[width=\linewidth]{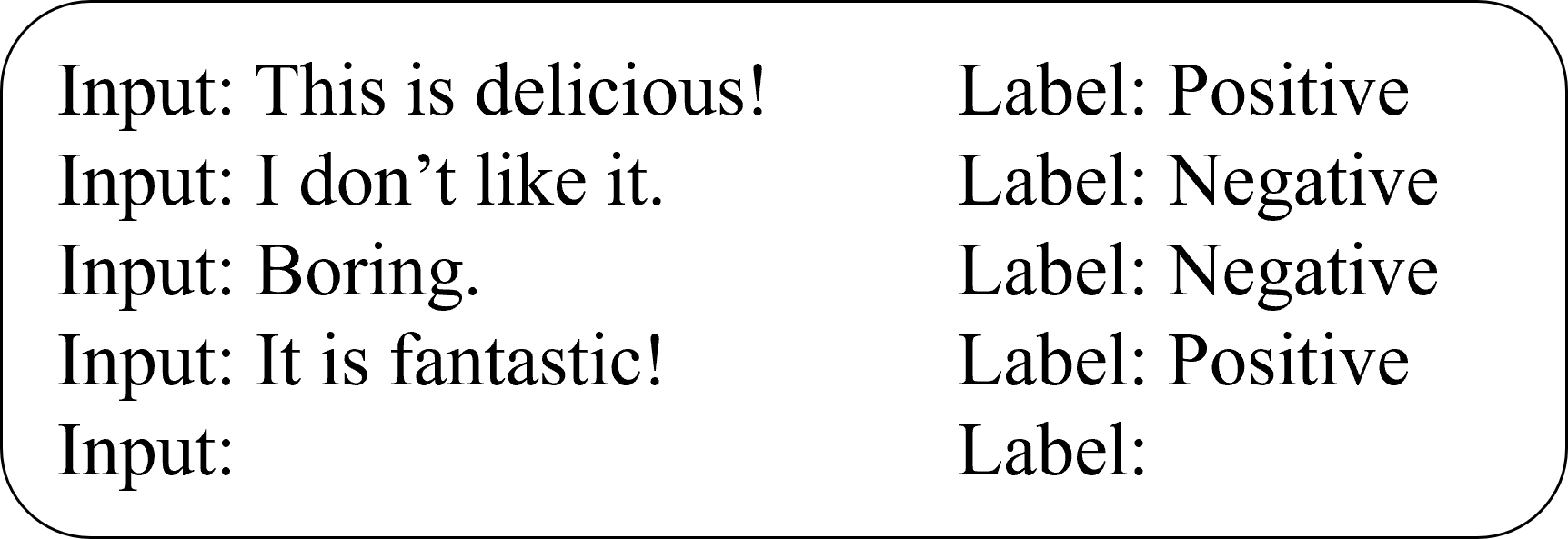}}
    \caption{An example of inputs in the normalized label entropy calculation.}
    \label{fig:epte}
\end{figure}

\section{Searching and Stability of $\lambda$}
\label{Appendix:lambda}

We select a set of candidates of $\lambda$ as \{0, 0.002, 0.004, 0.006, 0.008, 0.01, 0.012, 0.014, 0.016, 0.018, 0.02, 0.04, 0.06, 0.08, 0.1, 0.2, 0.3, 0.4, 0.5, 0.6, 0.7, 0.8, 0.9, 1.0\}, test the performance of \M{} with this set of $\lambda$, and select the one with the best accuracy as the optimal $\lambda$.

\begin{figure*}[t]
    \centering
    \centerline{\includegraphics[width=1\linewidth]{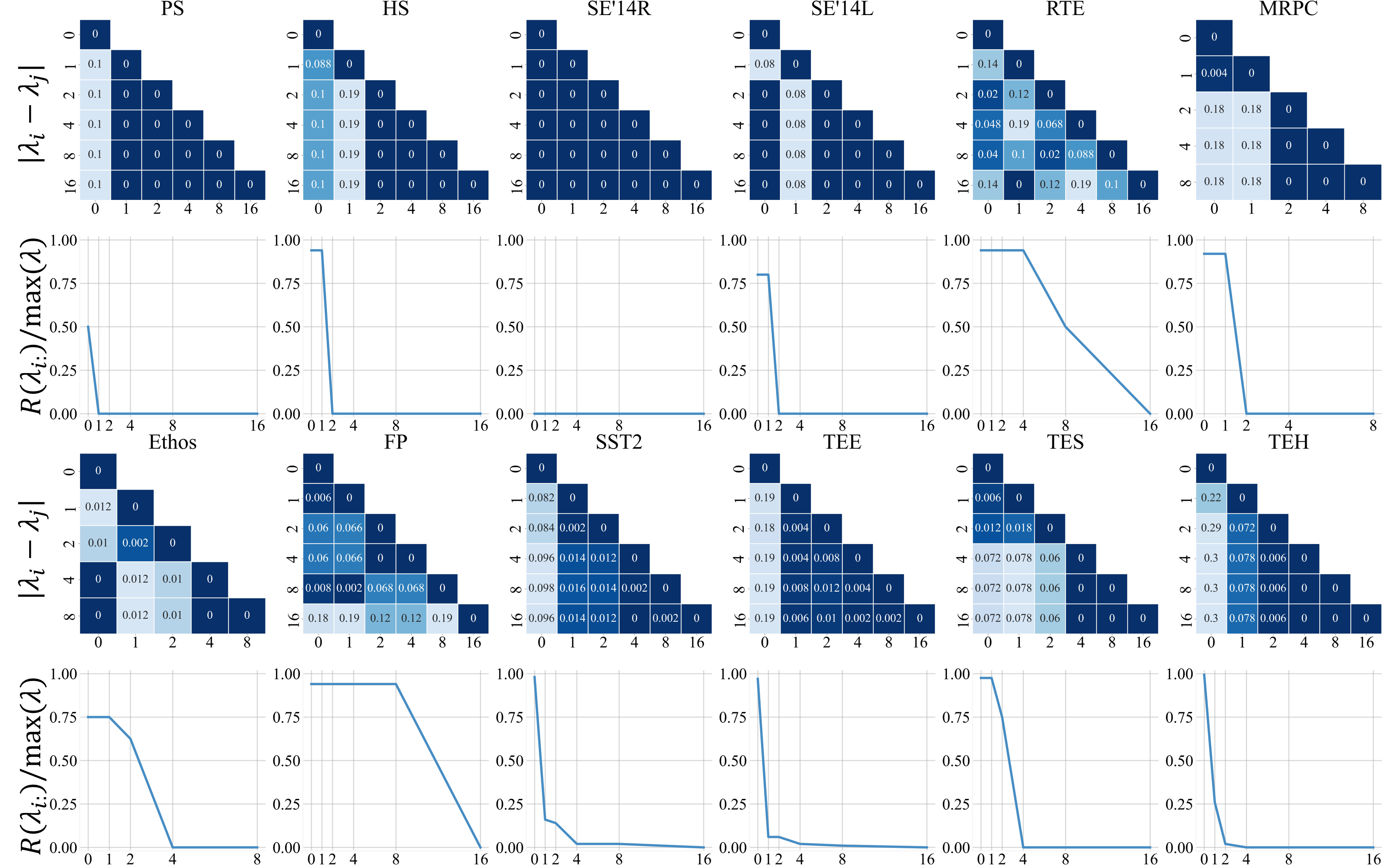}}
    \caption{The stability of the optimal $\lambda$ on various $k$ for GPT-J. Upper figure: the distance between the optimal $\lambda$ of different demo numbers; Lower figure: the normalized remaining range.}
    \label{fig:lamJ}
\end{figure*}

\begin{figure*}[t]
    \centering
    \centerline{\includegraphics[width=1\linewidth]{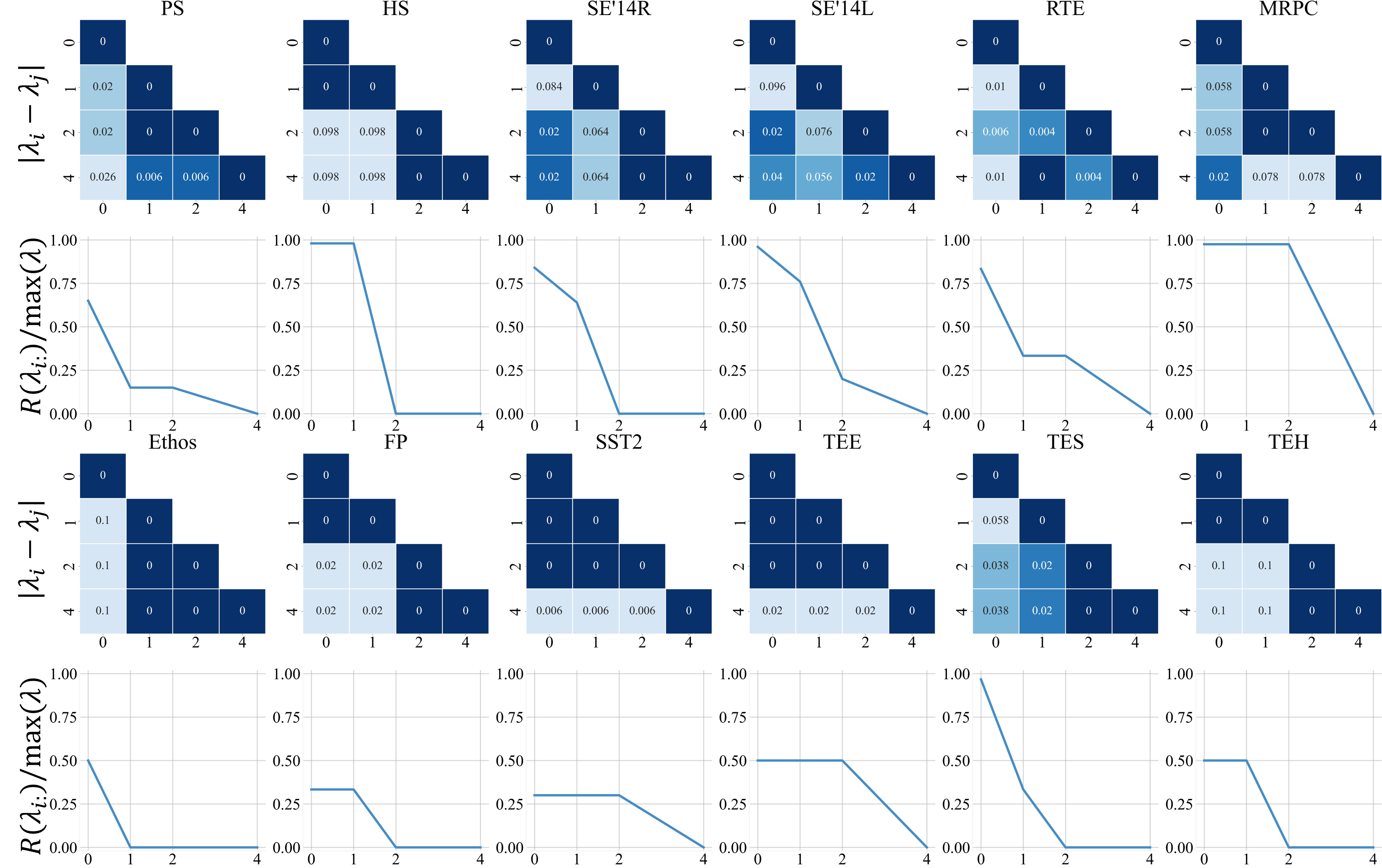}}
    \caption{The stability of the optimal $\lambda$ on various $k$ for GPT-2. Upper figure: the distance between the optimal $\lambda$ of different demo numbers; Lower figure: the normalized remaining range.}
    \label{fig:lam2}
\end{figure*}

Moreover, we find that given a dataset and model, $\lambda$ remains relatively stable w.r.t the demo number $k$, especially when $k$ is large. As shown in Fig. \ref{fig:lamJ} and Fig. \ref{fig:lam2}, we calculate $|\lambda_i - \lambda_j|$, the distance between the optimal $\lambda$ of different demo numbers as the heatmap, and the normalized remaining range $R(\lambda_{i:})/\max(\lambda)$, where the $R(\lambda_{i:})$ is the range of all the optimal $\lambda_c$ where the demo number $c\geq i$, and the $\max(\lambda)$ is the maximum of all the optimal $\lambda$ on the dataset. It characterizes the dispersion of $\lambda$ when the demo number is greater than the given one $i$.

The results show that while the $k$ is increasing, the distance and the normalized remaining range quickly zero out, which supports our hypothesis: \M{} is a bridge from the pre-training to ICL. For that during this process, the demo number should have a minimal impact.

\section{Calculation of $H_n^t$ and $H_n^l$}
\label{Appendix:entropy_calcu}

\paragraph{Normalized Token Entropy.} We calculate the normalized token entropy $H_n^t$ of the predicted probability distribution of tokens with an empty input as:
\begin{equation}
    H_n^t = -\frac{\sum_{l\in\mathbb{V}} P_\theta(l\vert x_\varnothing) \ln{P_\theta(l \vert x_\varnothing)}}{\ln{\vert \mathbb{V} \vert}},
\end{equation}

\noindent where the $P_\theta(\cdot)$ is an LM with a vocabulary space $\mathbb{V}$ and size $\vert \mathbb{V} \vert$, the $x_\varnothing$ is an empty input, such as ``~'', or ``Label:~''. The model outputs a global probability distribution of tokens when the $x_\varnothing$ is given, and we calculate the normalized entropy on the distribution.

\paragraph{Normalized Label Entropy.} We calculate the normalized label entropy $H_n^l$ of the model prediction probability distribution on the label space with 4 demos and an empty queue from a dataset (Fig. \ref{fig:epte} is an example) as:
    
\begin{equation}
H_n^l = -\frac{\sum_{l\in\mathbb{U}} r_{P, \theta}(l\vert x_*) \ln{r_{P, \theta}(l\vert x_*)}}{\ln{\vert \mathbb{U} \vert}},
\end{equation}

\noindent where the $r_{P, \theta}(l|x_*)$ is the predicted frequency of label $l$ given the aforementioned query-less only-demos input $x_*$. We use 512 tries to estimate this frequency. The $\mathbb{U}$ and $|\mathbb{U}|$ are the label space and label space size, respectively. 

\section{Calculation of $ECE_1$}
\label{Appendix:ece1}

Given a prediction set $\mathcal{Y}=\{(\hat{y_i}, z_i, y_i)\}_{i=1}^n$ of predictions $\hat{y_i}$, predicted confidence $z_i\in [0, 1]$, and supervised label $y_i$, in the calculation of $ECE_1$, we first divide the confidence space $(0, 1)$ into $m=10$ equal bins $B_j=[0.1(j-1), 0.1j]_{j=1}^m$ according to the $z_i$ as shown in Fig. \ref{fig:confexam}. The amount of prediction in each bin is denoted as $|B_j|$. For each bin, we calculate the accuracy $\alpha_j=\frac{1}{|B_j|}\sum_{\hat{y_i}\in B_j, \hat{y_i}=y_1}1$ of predictions in the $j$th bin, and assign a standard accuracy based on the average of confidence in the bin, that is, $\alpha^s_j = \frac{1}{|B_j|}\sum_{z_i\in B_j}z_i$. The $ECE_1$ can be described as:

\begin{equation}
    ECE_1 = \sum_{j = 1}^m\frac{|B_j|}{n}\vert\alpha_j-\alpha_j^s\vert,
\end{equation}

In terms of implementation, we use the \verb|MulticlassCalibrationError| module given by the TorchMetrics \footnote{\url{lightning.ai/docs/torchmetrics/stable/classification/calibration_error.html\#multiclasscalibrationerror}} with \verb|n_bins=10|, \verb|norm='l1'|. 

\section{Complete Results: Reliability Diagrams and Confident Distribution}
\label{Appendix:confd}

\begin{figure*}[t]
    \centering
    \centerline{\includegraphics[width=1\linewidth]{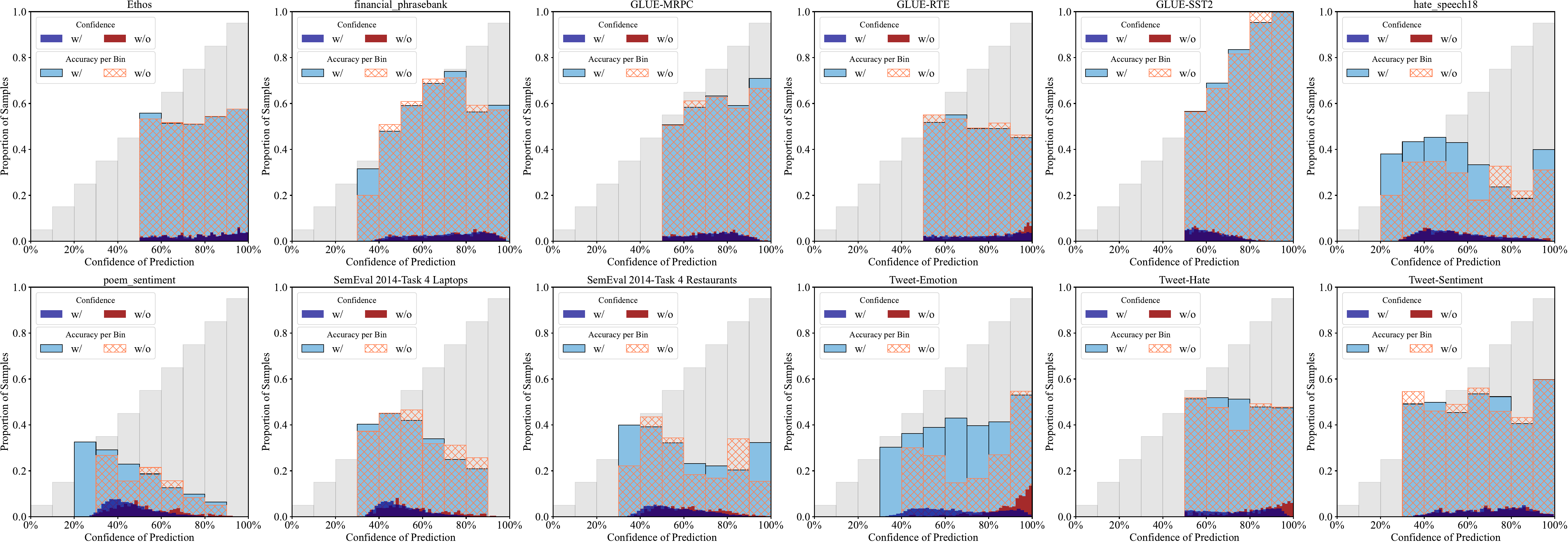}}
    \caption{The reliability diagrams of GPT-J ($k=0$).}
    \label{fig:relJ0}
\end{figure*}

\begin{figure*}[t]
    \centering
    \centerline{\includegraphics[width=1\linewidth]{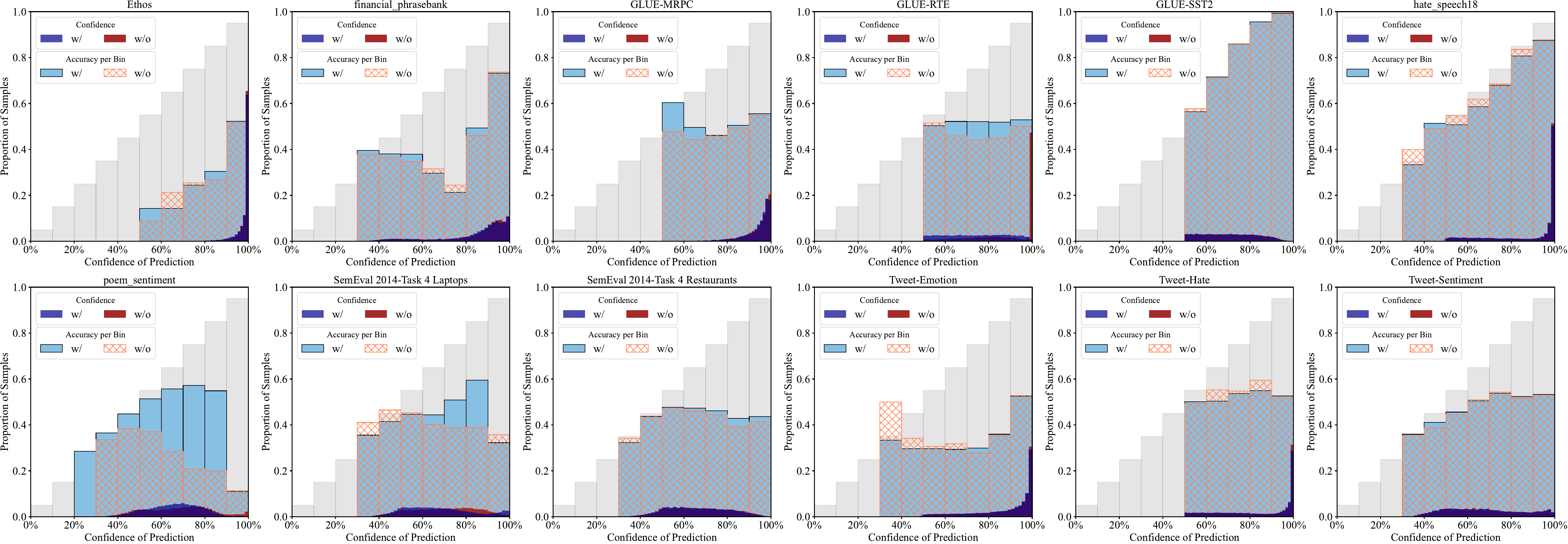}}
    \caption{The reliability diagrams of GPT-J ($k=1$).}
    \label{fig:relJ1}
\end{figure*}

\begin{figure*}[t]
    \centering
    \centerline{\includegraphics[width=1\linewidth]{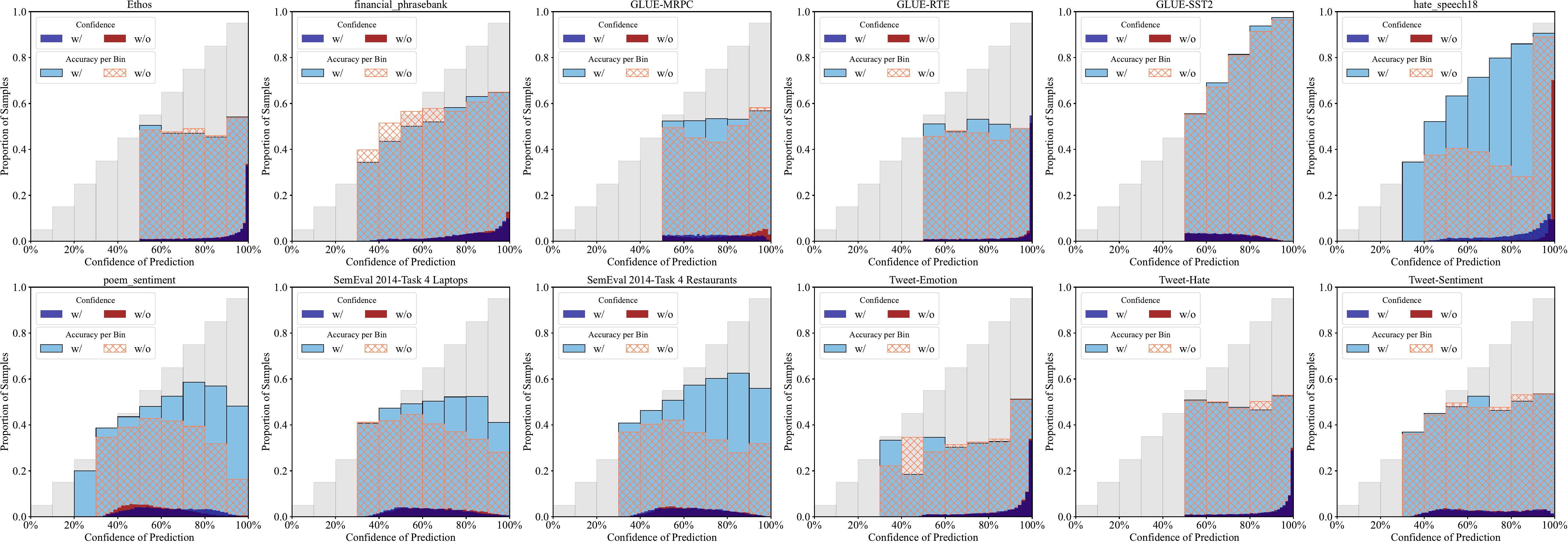}}
    \caption{The reliability diagrams of GPT-J ($k=2$).}
    \label{fig:relJ2}
\end{figure*}

\begin{figure*}[t]
    \centering
    \centerline{\includegraphics[width=1\linewidth]{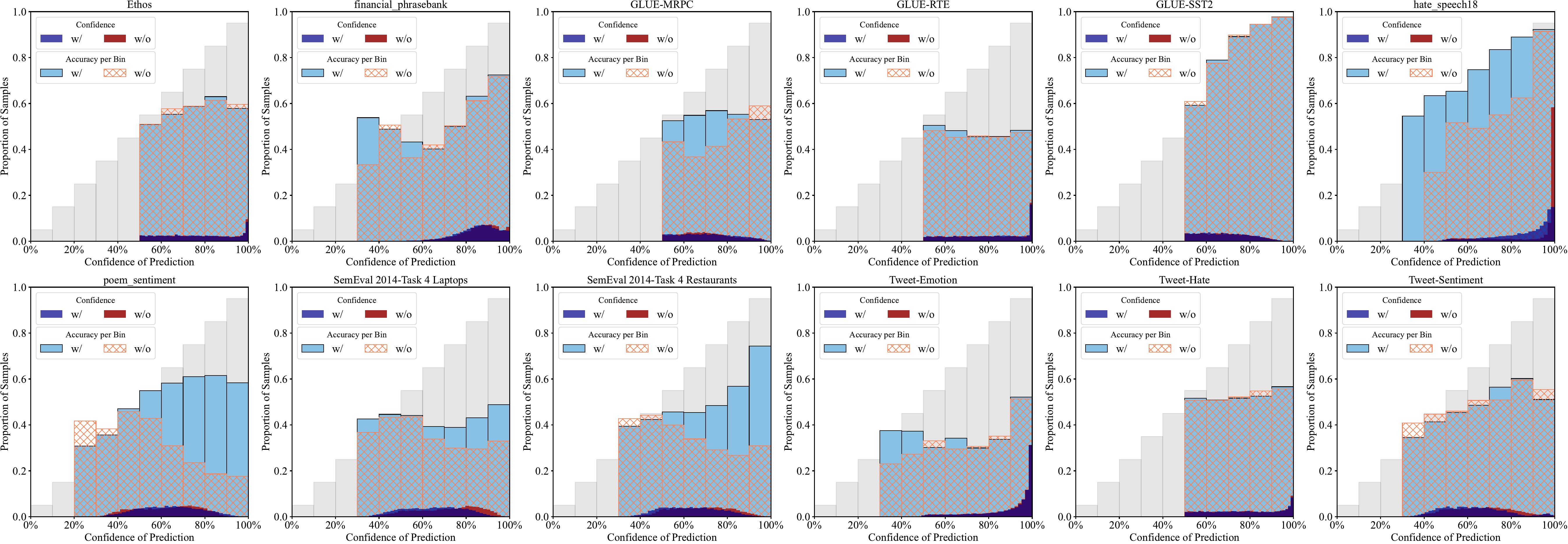}}
    \caption{The reliability diagrams of GPT-J ($k=4$).}
    \label{fig:relJ4}
\end{figure*}

\begin{figure*}[t]
    \centering
    \centerline{\includegraphics[width=1\linewidth]{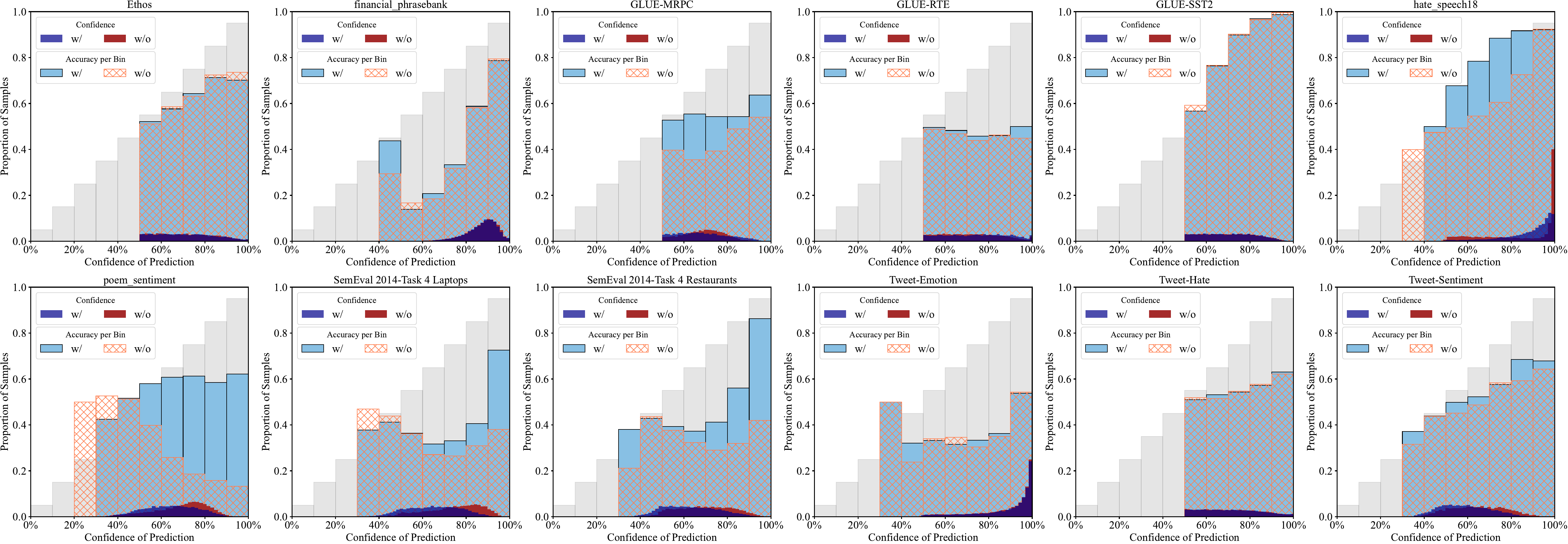}}
    \caption{The reliability diagrams of GPT-J ($k=8$).}
    \label{fig:relJ8}
\end{figure*}

\begin{figure*}[t]
    \centering
    \centerline{\includegraphics[width=1\linewidth]{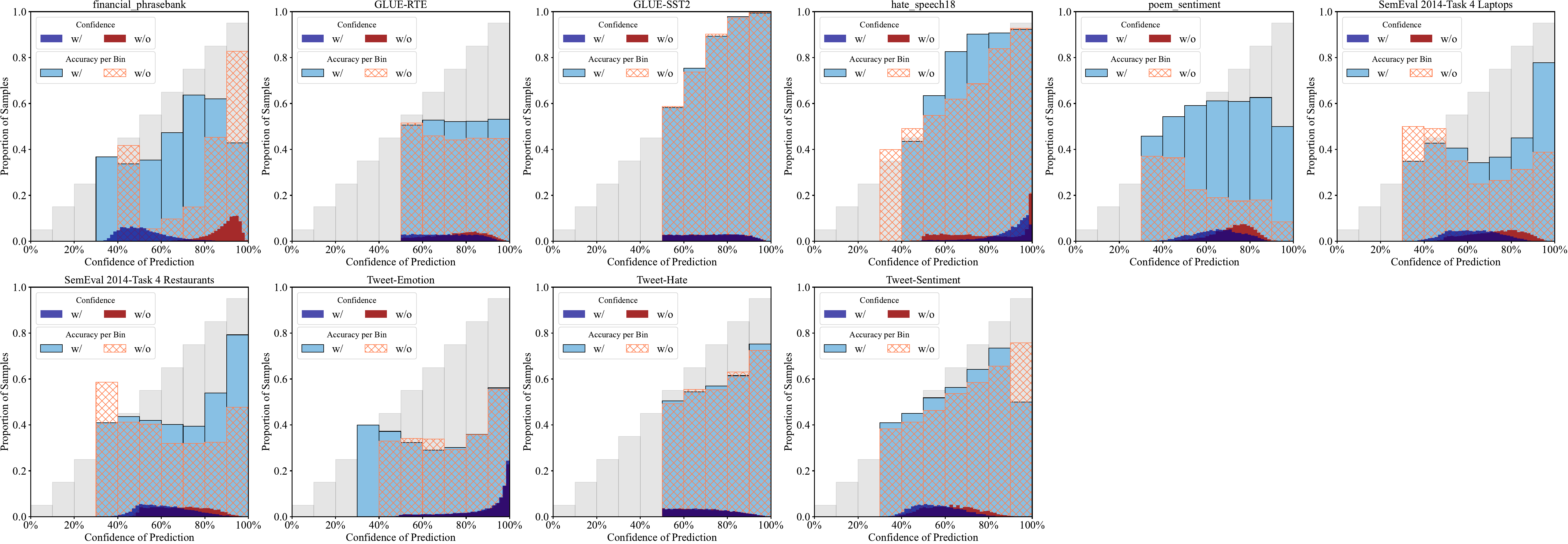}}
    \caption{The reliability diagrams of GPT-J ($k=16$).}
    \label{fig:relJ16}
\end{figure*}

\begin{figure*}[t]
    \centering
    \centerline{\includegraphics[width=1\linewidth]{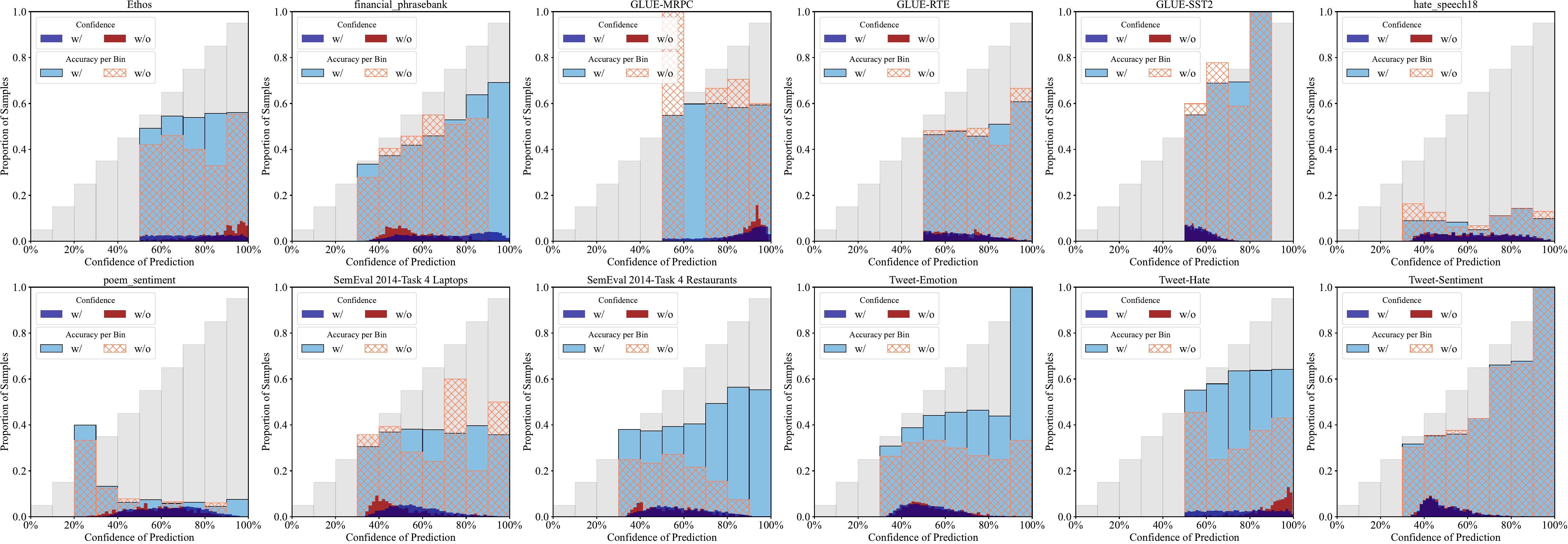}}
    \caption{The reliability diagrams of GPT-2 ($k=0$).}
    \label{fig:rel20}
\end{figure*}

\begin{figure*}[t]
    \centering
    \centerline{\includegraphics[width=1\linewidth]{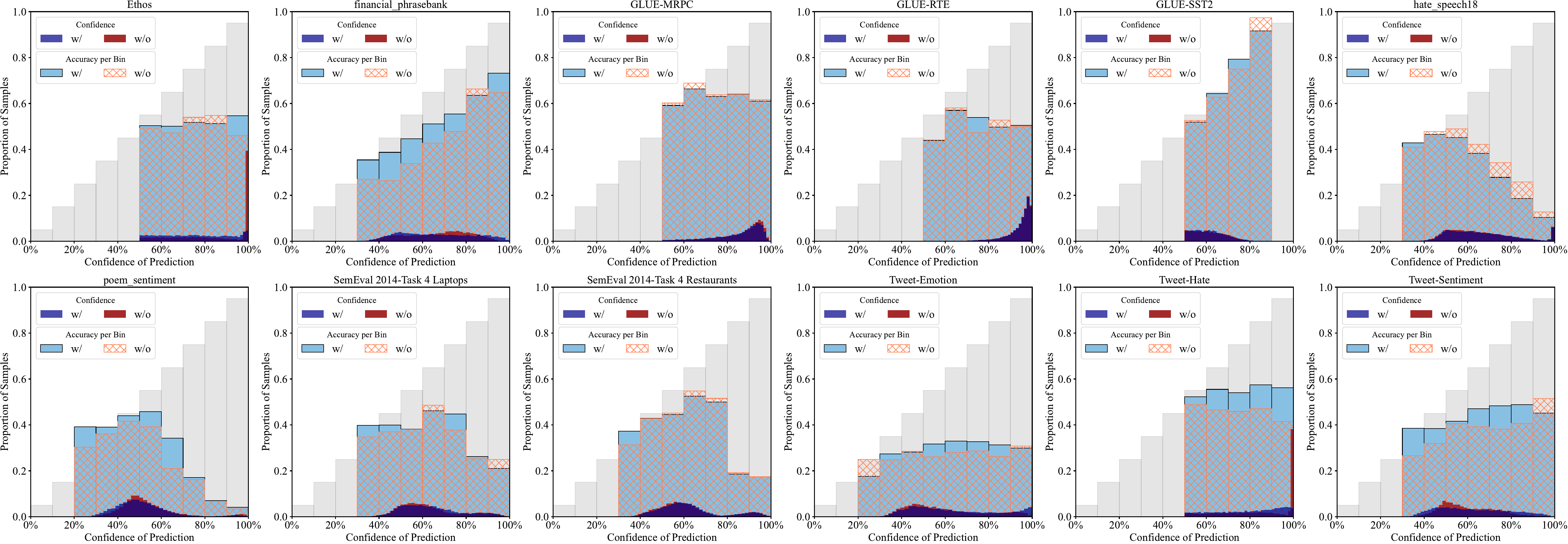}}
    \caption{The reliability diagrams of GPT-2 ($k=1$).}
    \label{fig:rel21}
\end{figure*}

\begin{figure*}[t]
    \centering
    \centerline{\includegraphics[width=1\linewidth]{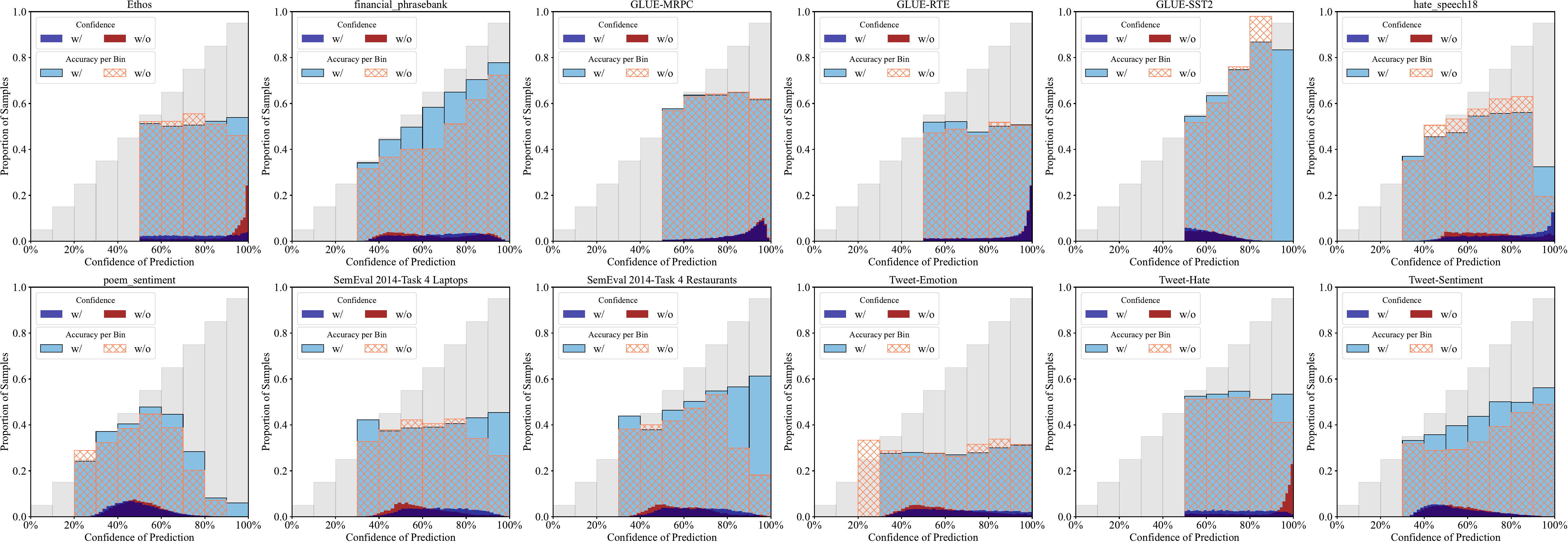}}
    \caption{The reliability diagrams of GPT-2 ($k=2$).}
    \label{fig:rel22}
\end{figure*}

\begin{figure*}[t]
    \centering
    \centerline{\includegraphics[width=1\linewidth]{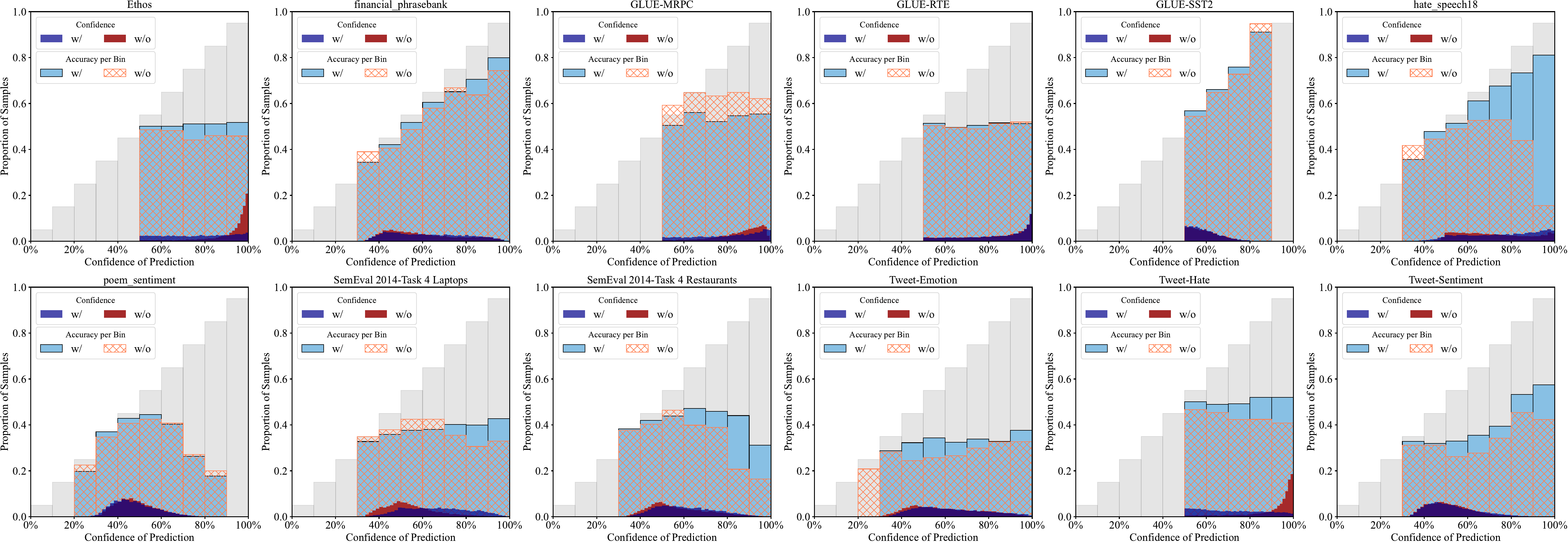}}
    \caption{The reliability diagrams of GPT-2 ($k=4$).}
    \label{fig:rel24}
\end{figure*}

\textbf{Reliability diagrams}: as shown in Fig. \ref{fig:relJ0} - \ref{fig:relJ16} for GPT-J, and Fig. \ref{fig:rel20} - \ref{fig:rel24} for GPT-2.

\begin{figure*}[t]
    \centering
    \centerline{\includegraphics[width=1\linewidth]{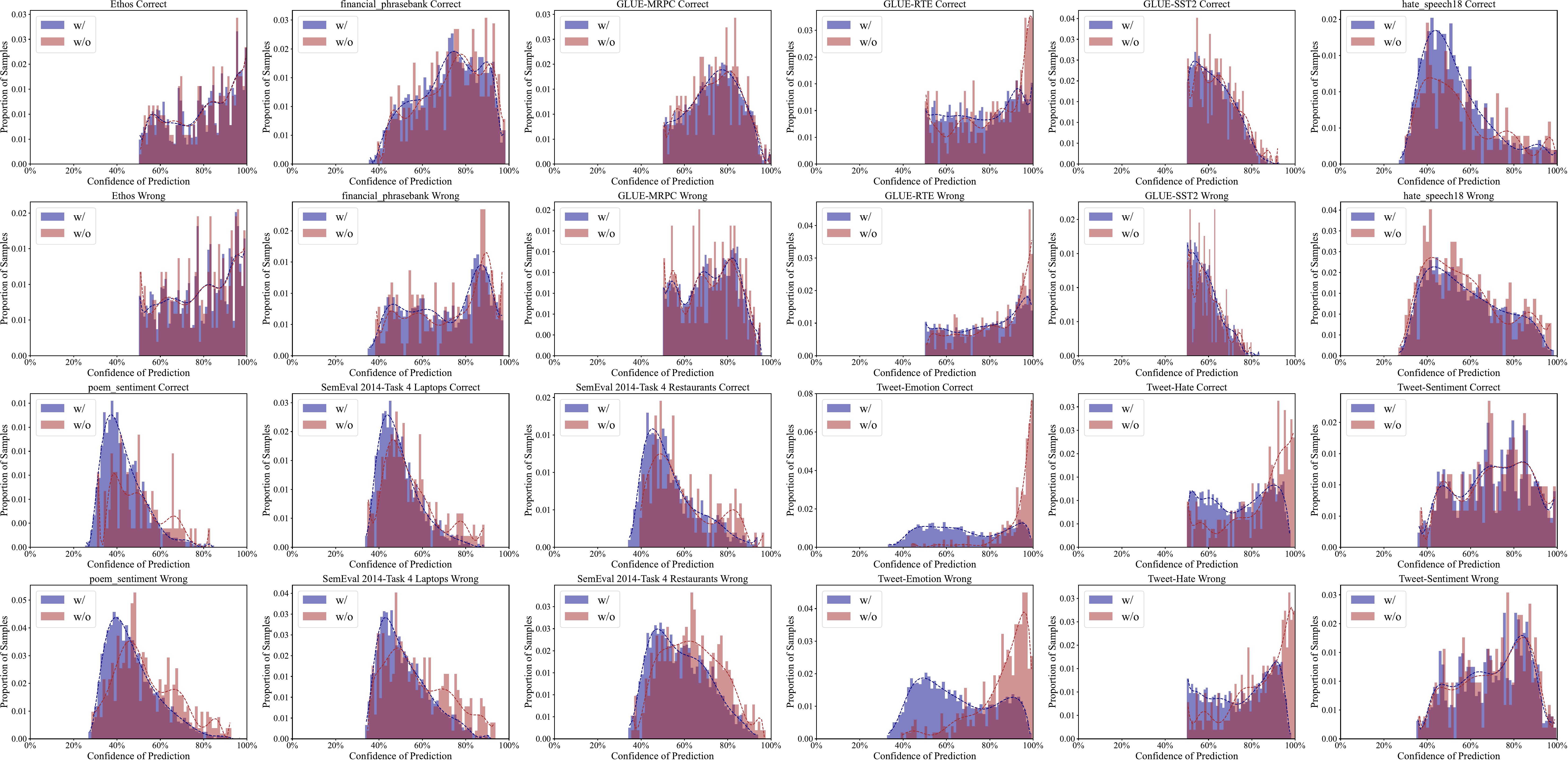}}
    \caption{The confidence distribution of GPT-J ($k=0$).}
    \label{fig:disJ0}
\end{figure*}

\begin{figure*}[t]
    \centering
    \centerline{\includegraphics[width=1\linewidth]{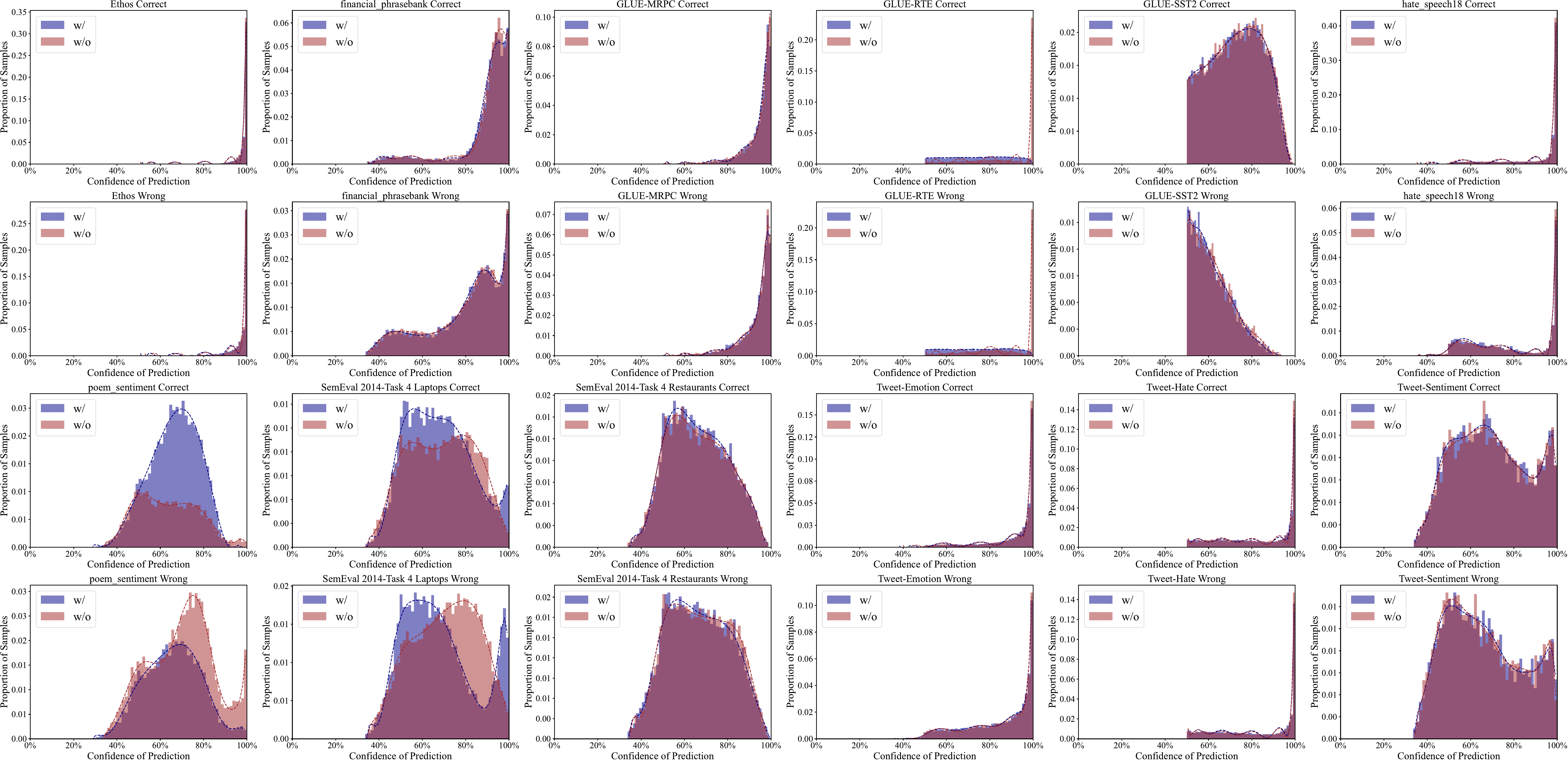}}
    \caption{The confidence distribution of GPT-J ($k=1$).}
    \label{fig:disJ1}
\end{figure*}

\begin{figure*}[t]
    \centering
    \centerline{\includegraphics[width=1\linewidth]{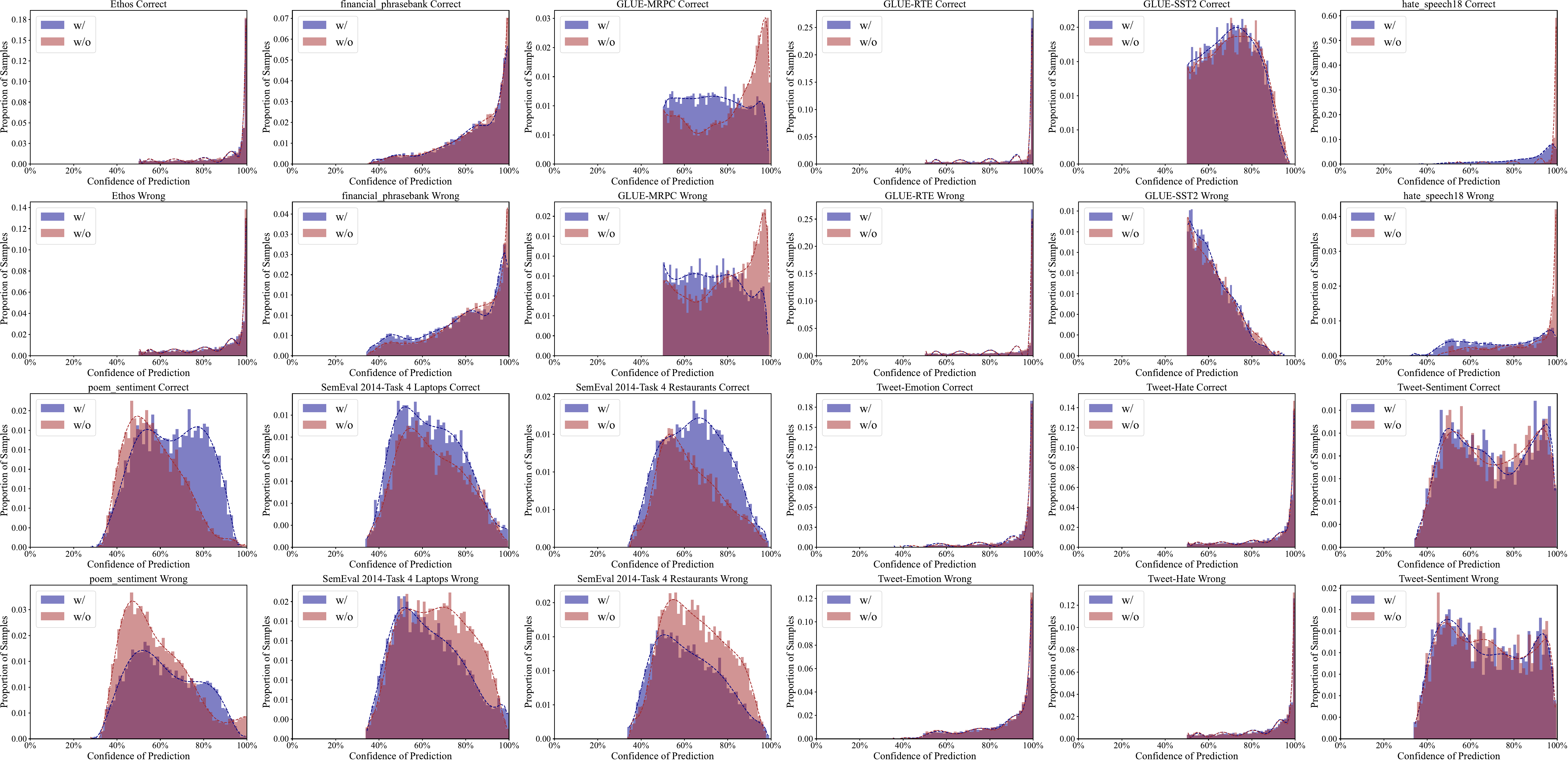}}
    \caption{The confidence distribution of GPT-J ($k=2$).}
    \label{fig:disJ2}
\end{figure*}

\begin{figure*}[t]
    \centering
    \centerline{\includegraphics[width=1\linewidth]{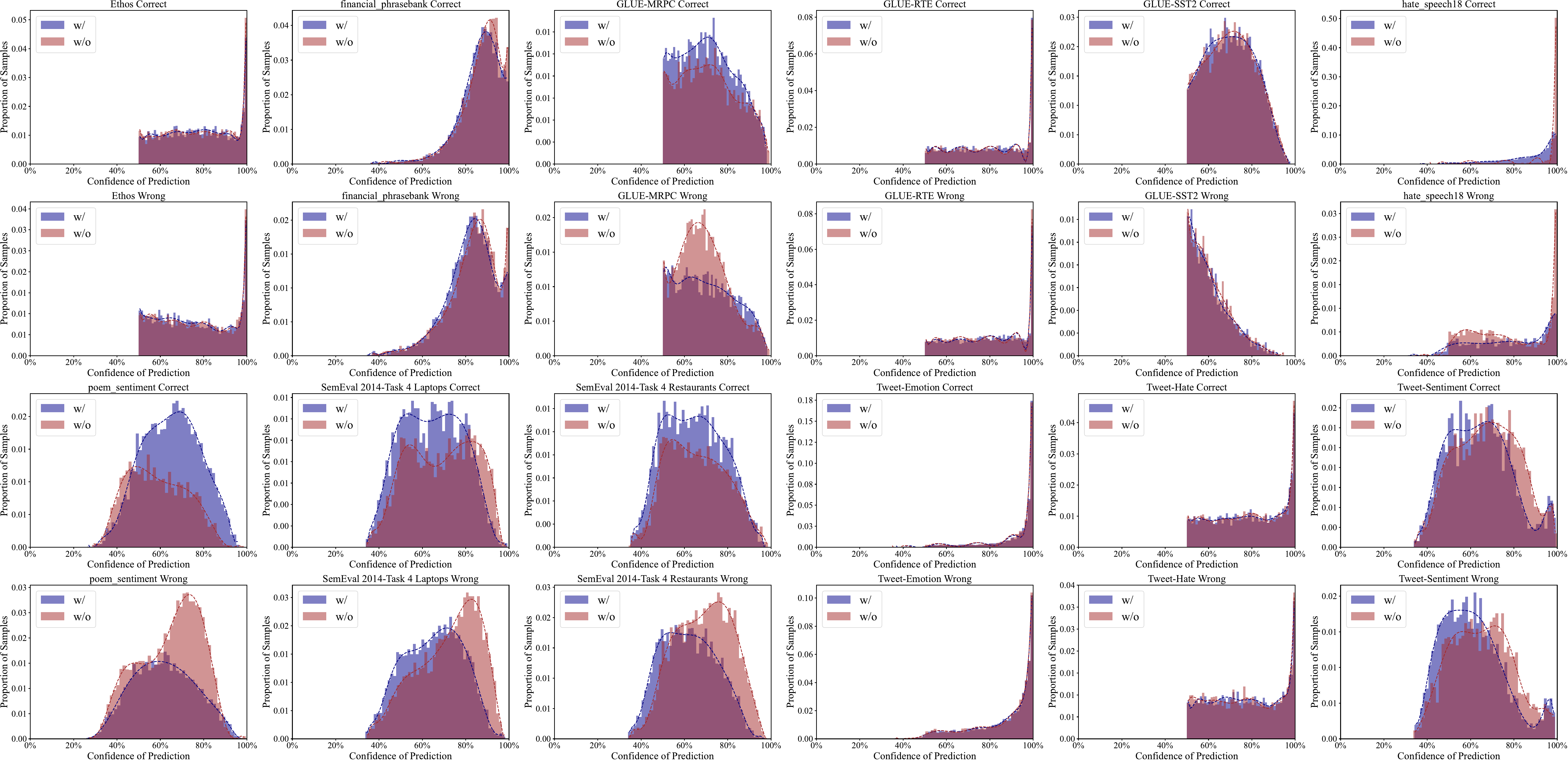}}
    \caption{The confidence distribution of GPT-J ($k=4$).}
    \label{fig:disJ4}
\end{figure*}

\begin{figure*}[t]
    \centering
    \centerline{\includegraphics[width=1\linewidth]{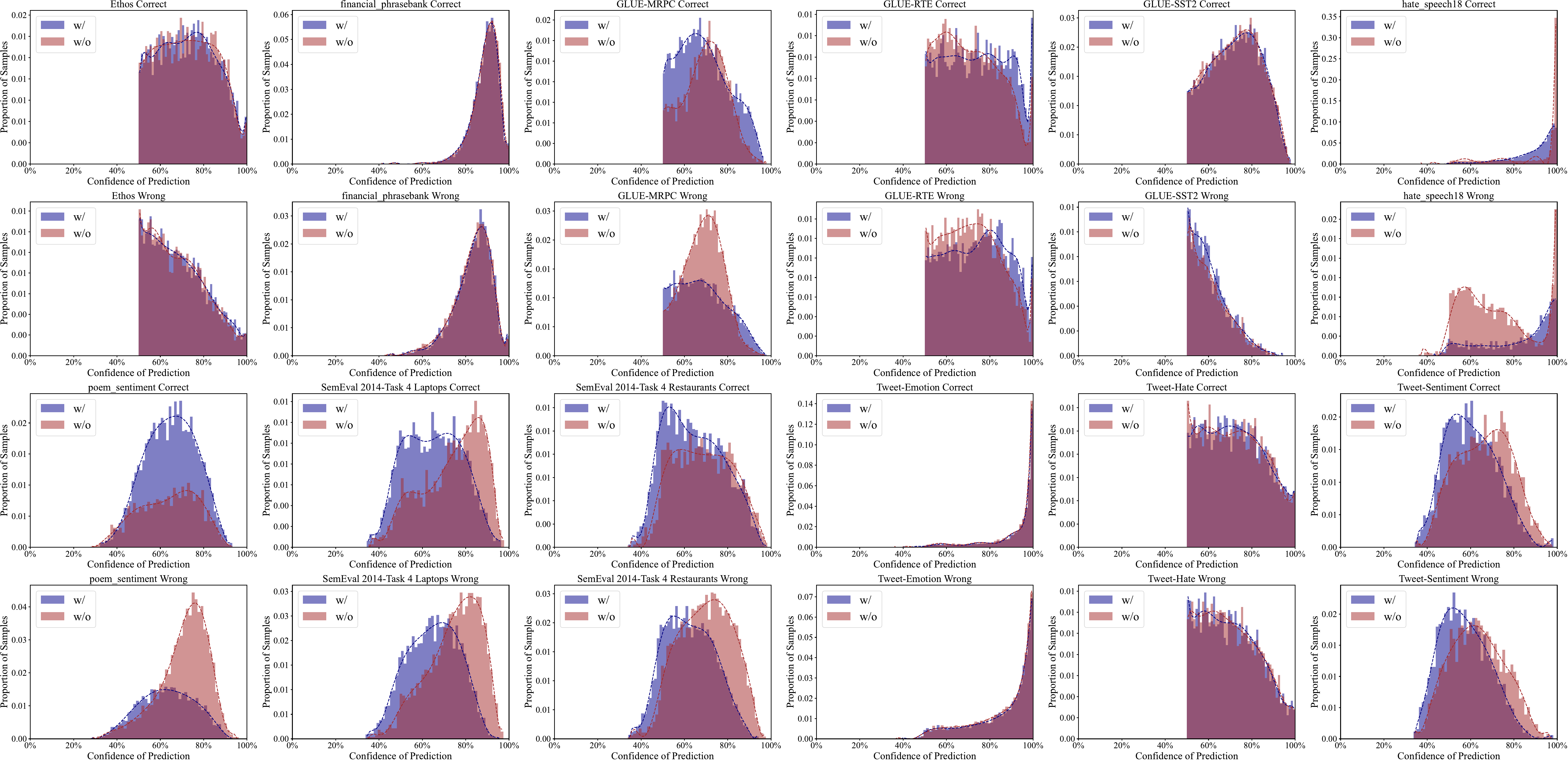}}
    \caption{The confidence distribution of GPT-J ($k=8$).}
    \label{fig:disJ8}
\end{figure*}

\begin{figure*}[t]
    \centering
    \centerline{\includegraphics[width=1\linewidth]{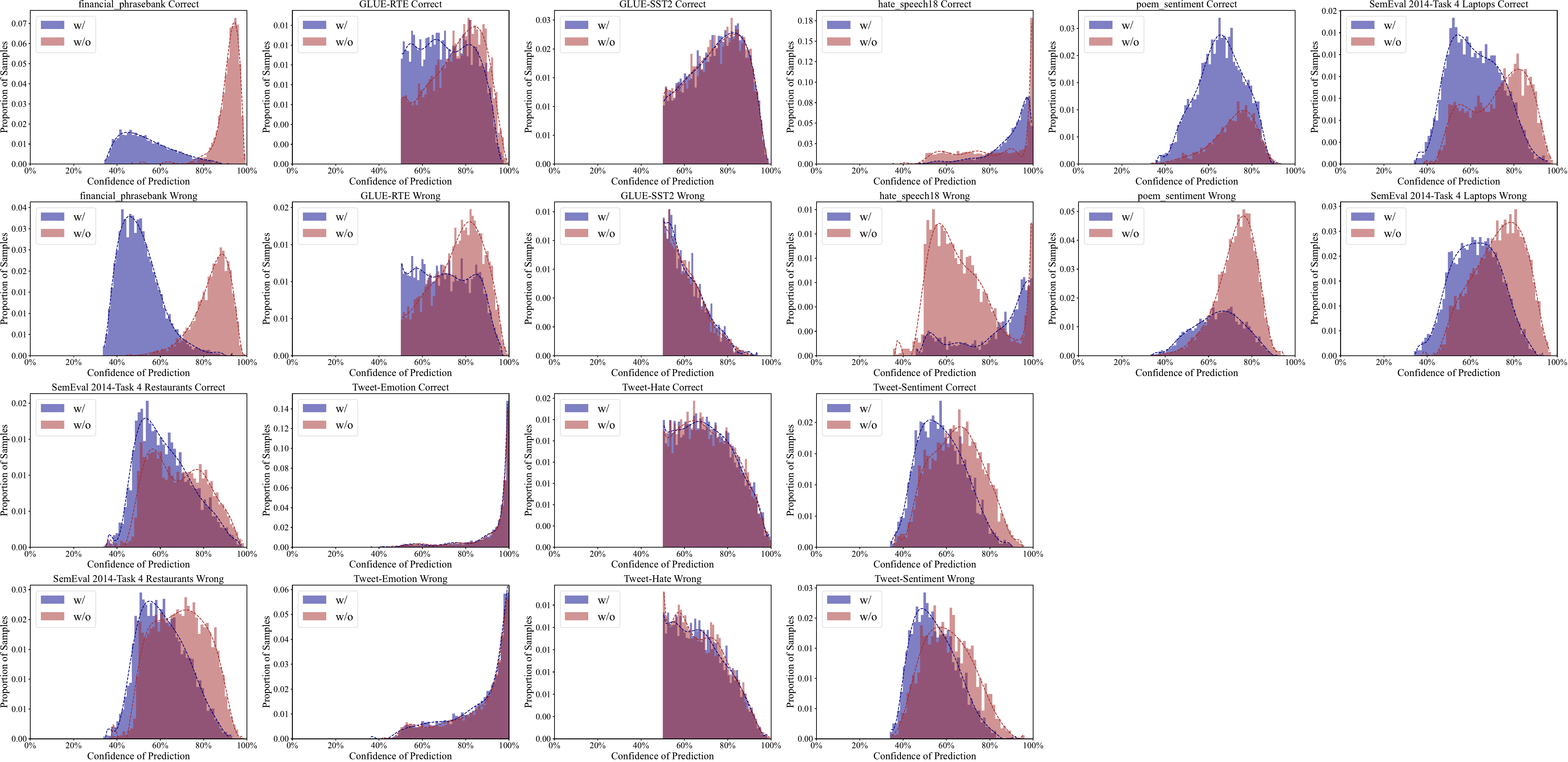}}
    \caption{The confidence distribution of GPT-J ($k=16$).}
    \label{fig:disJ16}
\end{figure*}

\begin{figure*}[t]
    \centering
    \centerline{\includegraphics[width=1\linewidth]{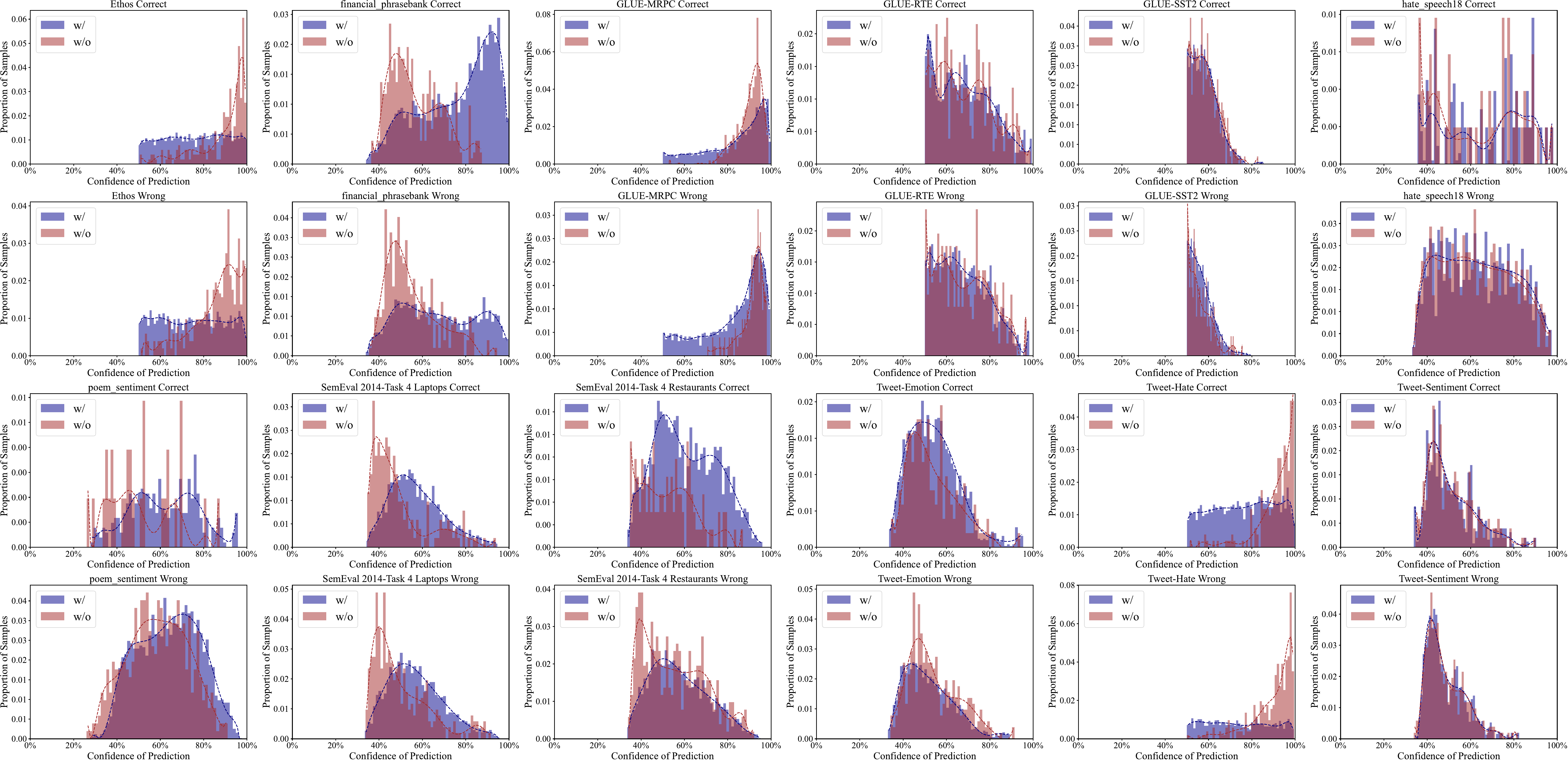}}
    \caption{The confidence distribution of GPT-2 ($k=0$).}
    \label{fig:dis20}
\end{figure*}

\begin{figure*}[t]
    \centering
    \centerline{\includegraphics[width=1\linewidth]{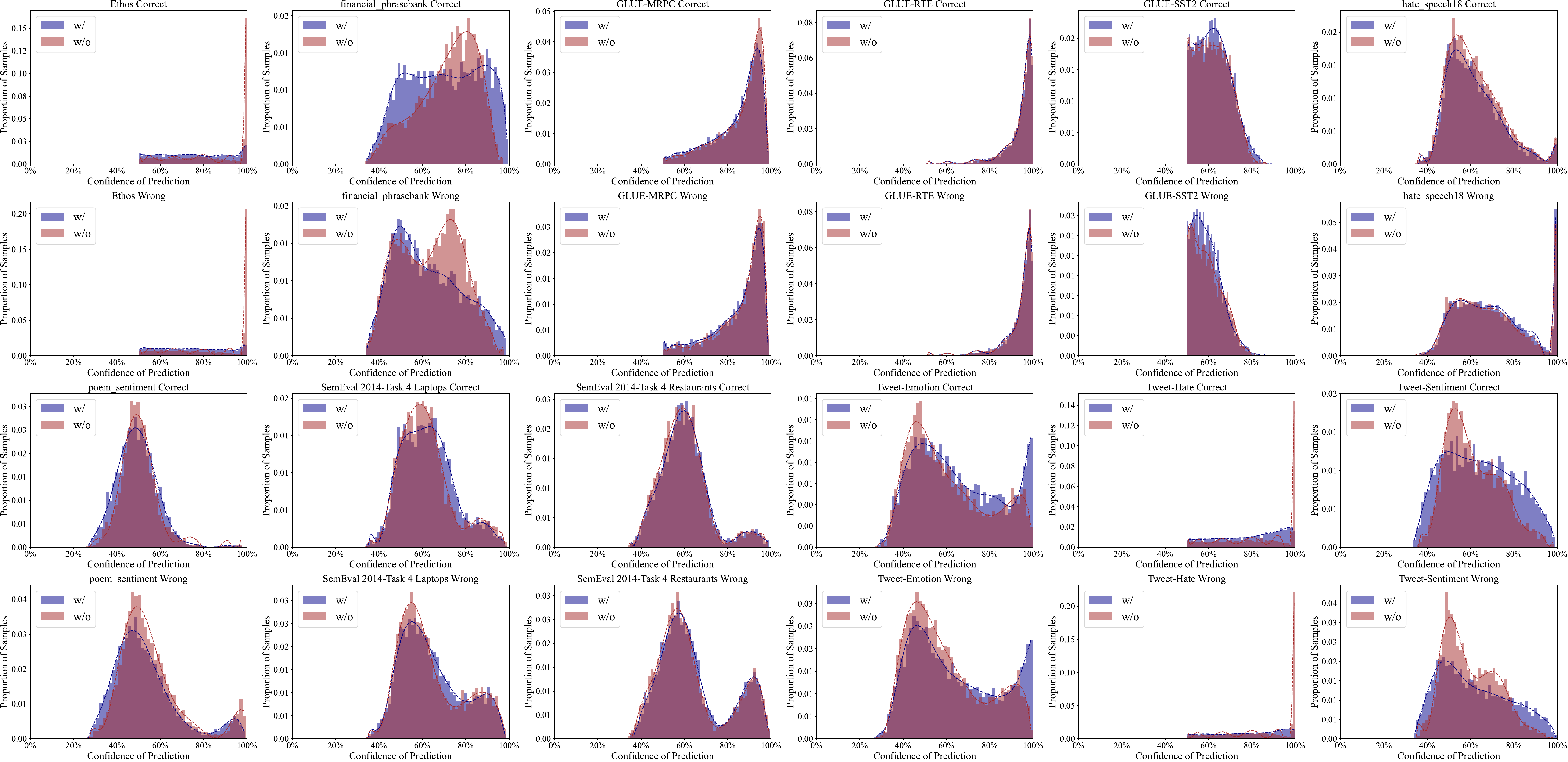}}
    \caption{The confidence distribution of GPT-2 ($k=1$).}
    \label{fig:dis21}
\end{figure*}

\begin{figure*}[t]
    \centering
    \centerline{\includegraphics[width=1\linewidth]{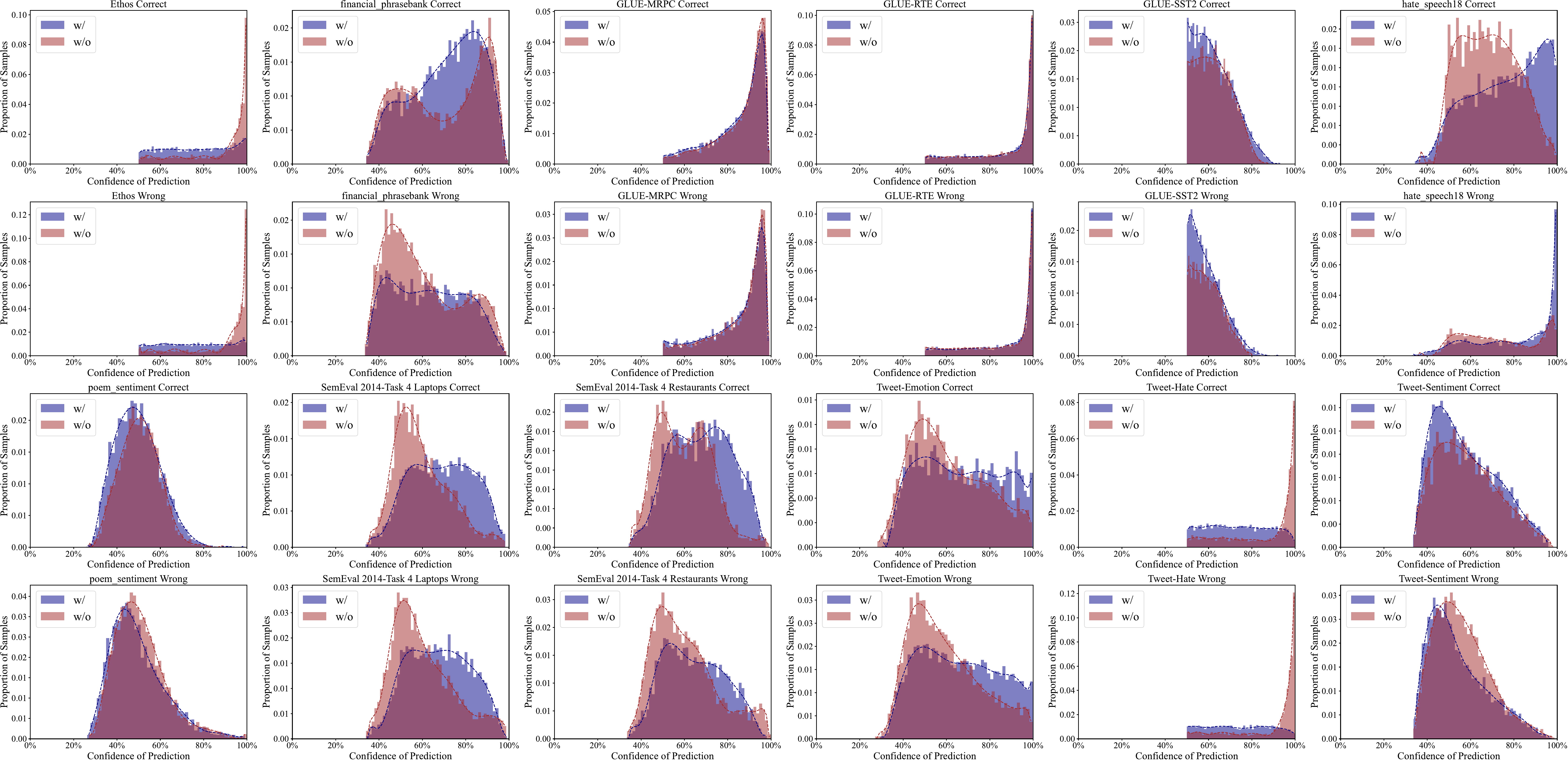}}
    \caption{The confidence distribution of GPT-2 ($k=2$).}
    \label{fig:dis22}
\end{figure*}

\begin{figure*}[t]
    \centering
    \centerline{\includegraphics[width=1\linewidth]{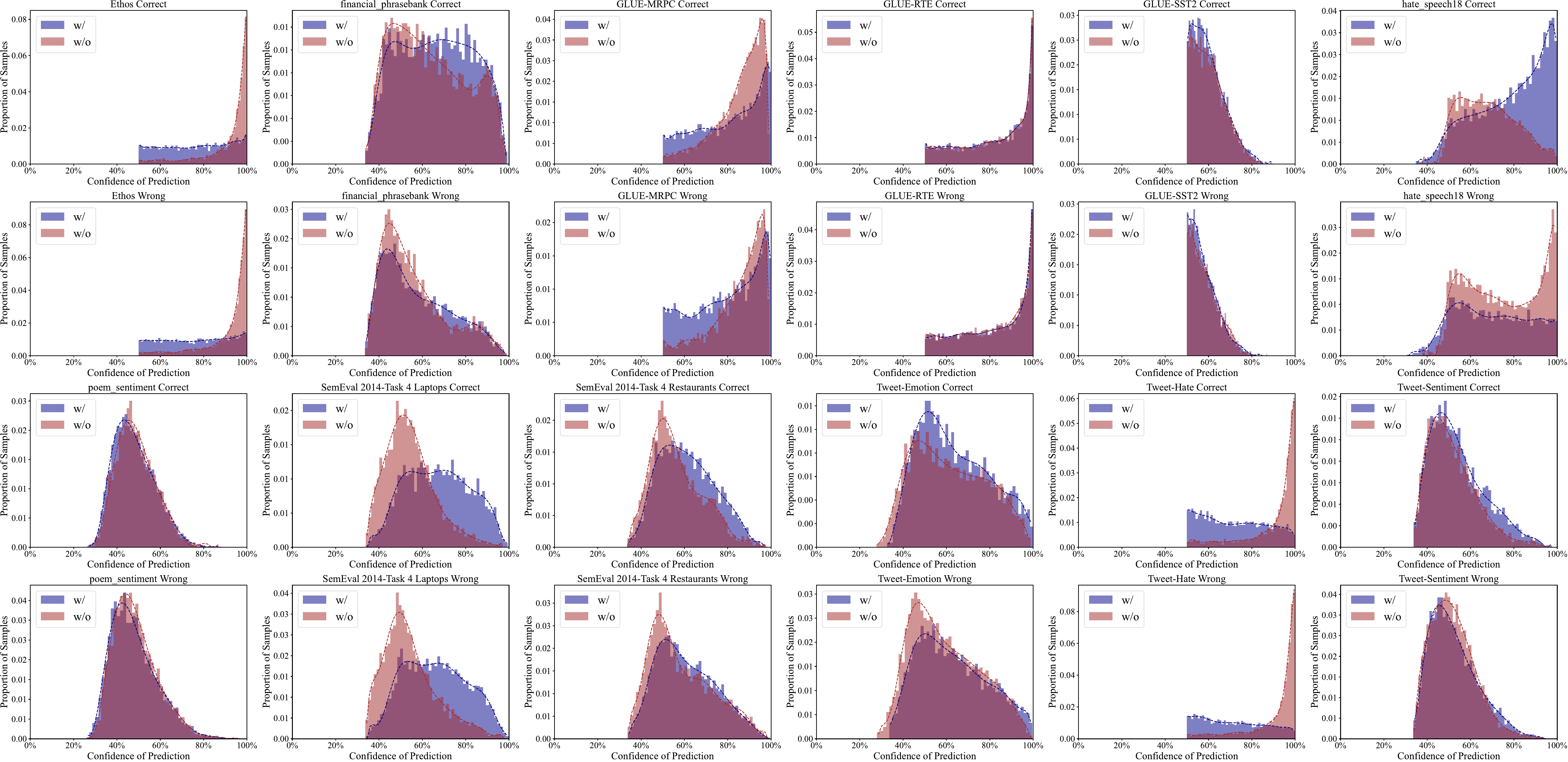}}
    \caption{The confidence distribution of GPT-2 ($k=4$).}
    \label{fig:dis24}
\end{figure*}

\noindent\textbf{Confidence distributions}: as shown in Fig. \ref{fig:disJ0} - \ref{fig:disJ16} for GPT-J, and Fig. \ref{fig:dis20} - \ref{fig:dis24} for GPT-2.

%%%%%%%%%%%%%%%%%%%%%%%%%%%%%%%%%%%%%%%%%%%%%%%%%%%%%
%%%%%%%%%%%%%%%%%%%%%%%%%%%%%%%%%%%%%%%%%%%%%%%%%%%%%%%
%%%%%%%%%%%%%%%%%%%%%%%%%%%%%%%%%%%%%%%%%%

\section{Case Analysis: \M\ Furtherance Correct ICL}

Moreover, we find that in some cases, unperturbed ICL can't benefit correctly from scaling the number of demos, while, \M\ can help the model correct this issue, as shown in Fig.~\ref{fig:demo}. These unperturbed models exhibit an overfitting-like phenomenon and also low accuracies, while \M\ can relieve it.

We speculate the reason is the mismatch between the pre-trained knowledge and ICL inputs. This leads to a decrease in the model's in-context task learning \cite{pan2023context} ability, while \M\ reduces such a gap between pre-trained data and ICL style data, which makes models extract information from ICL inputs better.

\section{License for Artifacts}

Here we discuss the license of the artifacts used in this paper.

\paragraph{Models.} GPT-2 is under the MIT license, and GPT-J is under the apache-2.0 license.

\paragraph{Datasets.} We list the open-source license for the datasets used in this paper as follows:

\begin{itemize}
    \item cc-by-4.0: PS, SE'14R, SE'14L, TEE, TES, TEH
    \item cc-by-sa-3.0: HS, SST2, RTE, MRPC, FP
    \item agpl-3.0: Ethos
\end{itemize}

\paragraph{Consistency of Usage.} Models are used with their original usage. Due to the different data splitting of these datasets, to ensure the consistency of the experiment methods, we use a re-splitting method as described in Appendix ~\ref{Appendix:Datasets}. However, the overall usage is consistent with their intended use.

\end{document}